\documentclass[openany]{now}
\usepackage{amsmath,cancel,multirow,subfigure,pifont,tikz,amssymb}
\usetikzlibrary{arrows,shapes,backgrounds,through,shadows,snakes}
\usepackage{rotating}

\newcommand{\figref}[1]{Figure \ref{#1}}
\newcommand{\chapref}[1]{Chapter \ref{#1}}
\newcommand{\secref}[1]{\S \ref{#1}}
\renewcommand{\eqref}[1]{Equation (\ref{#1})}
\def\ci{\perp\!\!\!\perp}
\newcommand{\av}[1]{\langle #1 \rangle}
\DeclareMathOperator*{\argmax}{arg\,max}

\newcommand{\trans}{^{\textsf{T}}}
\newcommand{\HMSM}{MSM}
\newcommand{\HMSMs}{MSMs}
\newcommand{\SARM}{SARM}
\newcommand{\GSARM}{GSARM}

\newcommand{\HMM}{HMM}

\newcommand{\SLGSSM}{SLGSSM}
\newcommand{\LGSSM}{LGSSM}
\newcommand{\LGSSMs}{LGSSMs}

\definecolor{myred}{cmyk}{0,0.9,0.9,0.1}
\definecolor{myredd}{cmyk}{0.1,1,1,0.5}

\newcommand{\myredd}{\color{myredd}}

\newcommand\Tstrut{\rule{0pt}{3.6ex}}         
\tikzstyle{disc}=[rectangle,draw=blue!50,thick,line width=1pt,minimum size=6mm]  
\tikzstyle{obs}=[fill=blue!20,thick]  
\tikzstyle{ocont}=[ellipse,draw=blue!50,thick,minimum size=6mm,>=stealth]  
\tikzstyle{dgraph}=[->, line width=1.5pt]
\tikzstyle{ugraph}=[line width=1.5pt]

\newcommand{\timeseries}{time series}
\newcommand{\Timeseries}{Time series}
\newcommand{\ie}{\emph{i.e.}}
\newcommand{\eg}{\emph{e.g.}}

\title{Explicit-Duration Markov Switching Models}

\author{Silvia Chiappa\footnote{The author is currently at Google DeepMind, UK.}\\
Statistical Laboratory, University of Cambridge, UK\\
and\\
Microsoft Research Cambridge, UK\\
chiappa.silvia@gmail.com}

\begin{document}

\copyrightowner{S.~Chiappa}
\volume{7}
\issue{6}
\pubyear{2014}
\copyrightyear{2014}
\isbn{978-1-60198-830-0}
\doi{10.1561/2200000054}
\firstpage{803}
\lastpage{886}

\frontmatter  

\maketitle

\tableofcontents

\mainmatter
\begin{abstract}
Markov switching models (\HMSMs) are probabilistic models that employ multiple sets of parameters
to describe different dynamic regimes that a \timeseries~may exhibit at different periods of time.
The switching mechanism between
regimes is controlled by unobserved random variables that form a first-order Markov chain.
Explicit-duration \HMSMs~contain additional variables that explicitly model the distribution of time spent in each regime.
This allows to define duration distributions of any form, but also to impose complex
dependence between the observations and to reset the dynamics to initial conditions.
Models that focus on the first two properties are most commonly known as hidden semi-Markov models or segment models,
whilst models that focus on the third property are most commonly known as changepoint models or reset models.
In this monograph, we provide a description of explicit-duration modelling by categorizing the
different approaches into three groups, which differ in encoding in the explicit-duration variables different information 
about regime change/reset boundaries.
The approaches are described using the formalism of graphical models, which allows to graphically represent and assess
statistical dependence and therefore to easily describe the structure of complex models and derive inference routines.
The presentation is intended to be pedagogical, focusing on providing a characterization of the three groups 
in terms of model structure constraints and inference properties.
The monograph is supplemented with a software package that contains most of the models and examples described\footnote{More information about the package is available at www.nowpublishers.com.}.
The material presented should be useful to both researchers wishing to learn about these models and
researchers wishing to develop them further.
\end{abstract}
\chapter{Introduction\label{chap:intro}}
Markov switching models (\HMSMs) are probabilistic models that
employ multiple sets of parameters to describe different
dynamic regimes that a \timeseries~may exhibit at different periods of time.
The switching mechanism between
regimes is controlled by unobserved variables that form a first-order Markov chain.

\HMSMs~are commonly used for segmenting \timeseries~or to retrieve the hidden dynamics underlying noisy observations.

\begin{figure}[t] 
\hskip-0.15cm
\subfigure[]{
\includegraphics[]{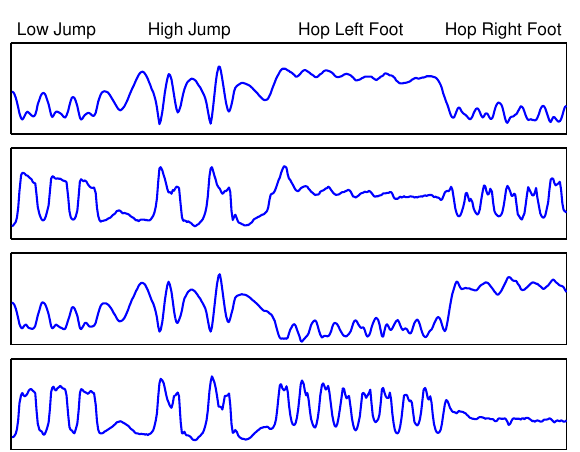}}
\hskip0.00cm
\subfigure[]{
\includegraphics[]{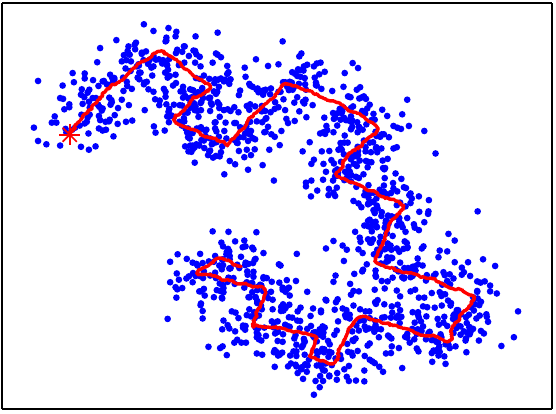}}
\caption{(a): Body-marker recording 
of an individual performing repetitions of the actions low jumping
up and down, high jumping up and down, hopping
on the left foot and hopping on the right foot (CMU Graphics Lab Motion Capture Database).
(b): Actual positions (continuous line) and measured positions (dots) of a two-wheeled robot moving in the two-dimensional space. The initial actual position is indicated with a star.}
\label{fig:intro}
\end{figure}

Consider, for example, the \timeseries~displayed in \figref{fig:intro}(a), which corresponds to the measured leg positions
of an individual performing repetitions of the actions low/high jumping and hopping on the left/right foot.
A segmentation of the \timeseries~into the underlying actions could be obtained
with a \HMSM~in which each action forms a separate regime,
\eg~by computing the regimes with highest posterior probabilities\footnote{This example is discussed in detail in \secref{sec:uslgssm}.}.

As another example, consider the \timeseries~displayed with dots in \figref{fig:intro}(b),
which corresponds to noisy observations of the positions of a two-wheeled robot moving in the two-dimensional space
according to straight movements, left-wheel rotations and right-wheel rotations (the actual positions are displayed with a continuous line).
Denoised estimates of the positions could be obtained with a \HMSM~in which the robot movements are described with continuous
unobserved variables and in which each type of movement forms a separate regime, \eg~by computing
the posterior means of the continuous variables\footnote{This example is discussed
in detail in \secref{sec:SLGSSM} and in Appendix~\ref{app:RL}.}.

In standard \HMSMs, the regime variables implicitly define a geometric distribution on the time spent in each regime.
In explicit-duration \HMSMs, this constraint is relaxed by using additional unobserved variables that allow to define duration distributions of any form.
Explicit-duration variables also allow to impose complex dependence between the observations and to reset the dynamics to initial conditions.

Explicit-duration \HMSMs~were first introduced in the speech community \citep{ferguson80variable} and are mostly used
to achieve more powerful modelling than standard \HMSMs~through the specification of more accurate duration distributions and dependencies between
the observations.
In this case, the models are most commonly known with the names of hidden semi-Markov models or segment models.
However, the possibility to reset the dynamics to initial conditions has recently led to the use of explicit-duration variables also for
Bayesian approaches to abrupt-change detection, for identifying repetitions of patterns (such as, \eg, the action repetitions underlying the \timeseries~in \figref{fig:intro}(a)),
and for performing/approximating inference\footnote{By inference we mean the computation of posterior distributions, namely distributions of unobserved variables conditioned on the observations.} \citep{fearnhead06exact,fearnhead09bayesian,chiappa10movement,bracegirdle11switch}.
In these cases, the models are most commonly known with the names of changepoint models or reset models.

Explicit-duration \HMSMs~have been used in many application areas including speech analysis
\citep{russell85explicit,levinson86continuously,rabiner89tutorial,gu91isolated,gales93theory,russell93segmental,ostendorf96from,moore04speech,liang11improvised},
handwriting recognition \citep{chen95variable},
activity recognition \citep{yu03hidden,huang06variable,oh08learning,chiappa10movement}, musical pattern
recognition \citep{pikrakis06classification}, financial \timeseries~analysis \citep{bulla06stylized}, rainfall \timeseries~analysis \citep{samsom01fitting},
protein structure segmentation \citep{schmidler10bayesian}, gene finding \citep{winters10hidden}, DNA analysis \citep{barbu08semi,fearnhead09bayesian}, plant analysis \citep{guedon01pattern}, MRI sequence analysis \citep{faisan02hidden}, ECG segmentation \citep{hughes04semi}, and waveform modelling \citep{kim06segmental}; see references in \cite{yu10hidden} for more examples.

Explicit-duration \HMSMs~originated from the idea of explicitly modelling the duration distribution by defining
a semi-Markov process on the regime variables, namely a process in which the trajectories are piecewise constant functions -- with interval durations drawn from an
explicitly defined duration distribution -- and in which the variables at jump times form a Markov chain. The first and
currently standard approach achieves that with variables indicating the
interval duration, and derives inference recursions using only jump times \citep{rabiner89tutorial,gales93theory,ostendorf96from,yu10hidden}.
To simplify the derivations of posterior distributions at times that are different from jump times, 
\cite{chiappa10movement} use count variables in addition to duration
variables, such that the combined regime and count-duration variables form a first-order Markov chain.
Other methods that explicitly model the duration distribution have been proposed with different goals and in different communities. These methods can
all be viewed as different ways to define a first-order Markov chain on the combined regime and explicit-duration variables that induces
a semi-Markov process on the regime variables.

In this monograph we provide a description of explicit-duration modelling that aims at elucidating the characteristics
of the different approaches and at clarifying and unifying the literature.
We identify three fundamentally different ways to define the
first-order Markov chain on the combined regime and explicit-duration variables, which differ in
encoding in the explicit-duration variables the location of (i) the preceding,
(ii) the following, or (iii) both the preceding and following regime change or reset.
We discuss each encoding in the context of \HMSMs~of simple unobserved structure and of \HMSMs~that contain extra
unobserved variables related by first-order Markovian dependence.
The models are described using the formalism of graphical models, which allows to graphically represent and assess
statistical dependence, and therefore to easily describe the structure of complex models and derive inference routines.

The remainder of the manuscript is organized as follows. \chapref{chap:Back} contains some background material.
We start with a general description of \HMSMs~and by showing that the regime variables implicitly define a geometric duration distribution.
In \secref{sec:HMM} we introduce the hidden Markov model, which represents the simplest \HMSM,
and explain how to obtain a negative binomial duration distribution with regime copies.
In \secref{sec:BN} we introduce the framework of graphical models, and explain how to graphically assess statistical independence in
a particular type of graphical models, called belief networks, that will be used for describing the models.
In \secref{sec:ILEX} we illustrate how belief networks can
be used to easily derive the standard inference recursions of \HMSMs. In \secref{sec:EM} we give a general explanation of
the expectation maximization algorithm, which represents the most popular algorithm for parameter learning in probabilistic models with unobserved variables.
In \chapref{chap:Dur} we describe the different approaches to explicit-duration modelling by categorizing them into three groups. The groups are introduced in \secref{sec:M1}, \secref{sec:M1alt} and \secref{sec:M2}.
In \secref{sec:VVM} we discuss in detail explicit-duration modelling in \HMSMs~containing only regime variables, explicit-duration variables, and observations.
In \secref{sec:SLGSSM} we discuss in detail explicit-duration modelling in a popular \HMSM~containing additional unobserved variables related by first-order Markovian dependence, namely the switching linear Gaussian state-space model,
and discuss how the findings generalize to similar models with unobserved variables related by first-order  Markovian dependence. The case of more complex unobserved structure is not considered.
In \secref{sec:Approx} we describe approximation schemes to reduce the computational cost of inference.
In \chapref{chap:Discuss} we summarize the most important points of our exposition and make some historical remarks.

\chapter{Background\label{chap:Back}}
Markov switching models (\HMSMs) describe a \timeseries~$v_1,\ldots,v_T=v_{1:T}$ using $S$ different sets of parameters, each defining a different dynamic regime.
This is achieved by using unobserved variables $s_{1:T}$, where $s_t\in\{1,\ldots,S\}$\footnote{For simplicity of exposition, we use the same symbol to indicate a random variable and its values.} indicates which of the $S$ regimes underlies observations $v_t$.
The regime variables form a first-order, time-homogeneous, Markov chain, \ie~the joint distribution $p(s_{1:T})$\footnote{We use the notation $p(\cdot)$ and $p(\cdot|\cdot)$ to indicate the probability density function and conditional probability density function with respect to a measure or product measures involving the Lebesgue and/or the counting
measures. We use term distribution to indicate the probability density function.} can be written as
\begin{align*}
p(s_{1:T})=p(s_1)\prod_{t=2}^T p(s_t|s_{t-1})=\tilde{\pi}_{s_1}\prod_{t=2}^T\pi_{s_ts_{t-1}},
\end{align*}
where $\tilde\pi$ is a vector of elements $\tilde{\pi}_{s_1}=p(s_1)$ and $\pi$ is a time-independent transition matrix of elements $\pi_{s_ts_{t-1}}=p(s_t|s_{t-1})$.

In standard \HMSMs, the regime variables implicitly define a geometric distribution on the time spent in each regime.
Indeed, given \eg~$s_t=i$, the probability of remaining in regime $i$ at time-steps $t+1,\ldots,t+d-1$ and switching to a different regime at time-step $t+d$ is
\begin{align*}
p(s_{t+1:t+d-1}\!=\!i,s_{t+d}\!\neq\! i|s_t\!=\!i)&=\pi_{ii}^{d-1}\sum_{j\neq i}\pi_{ji}=\pi_{ii}^{d-1}(1\!-\!\pi_{ii}),
\end{align*}
which corresponds to the geometric distribution with parameter $\pi_{ii}$. The geometric distribution with $\pi_{ii}\in\{0.1,0.5,0.9\}$ is shown in \figref{fig:GeoNBin}(a).

The remainder of the chapter is organized as follows. In \secref{sec:HMM} we describe the simplest \HMSM, namely the hidden Markov model, and show that a negative binomial duration distribution can be obtained with regime copies.
In \secref{sec:BN} we introduce the formalism of graphical models and show how this formalism can be used to derive the standard inference recursions of \HMSMs~-- a similar approach will be employed to derive inference recursions in the explicit-duration extensions.
In \secref{sec:EM} we describe the expectation maximization algorithm, which will be used for parameter learning throughout the manuscript.

\section{Hidden Markov Model\label{sec:HMM}}
\begin{figure}[t] 
\hskip-0.3cm
\subfigure[]{
\includegraphics[]{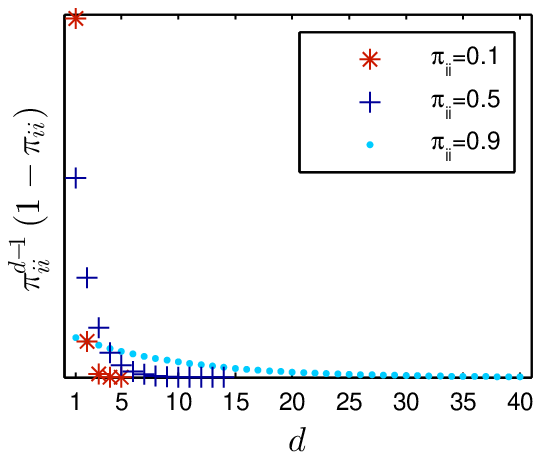}}
\hskip-0.02cm
\subfigure[]{
\includegraphics[]{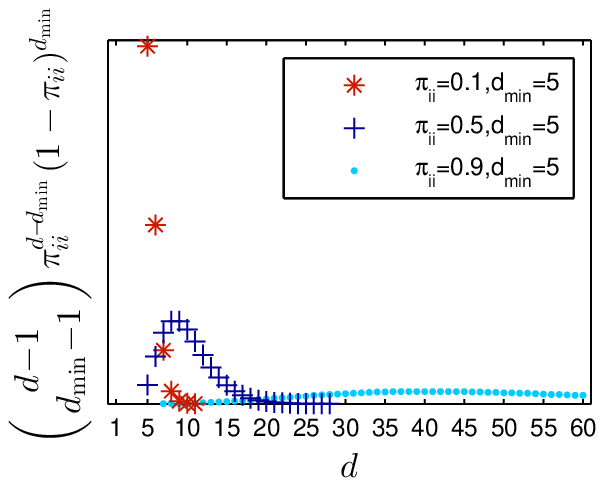}}
\caption{(a): Geometric duration distribution implicitly defined in a standard \HMSM~with $\pi_{ii}\in\{0.1,0.5,0.9\}$.
(b): Negative binomial duration distribution obtained by replacing regime $i$ in the standard \HMSM~in (a) with $d_{\min}=5$ copies.}
\label{fig:GeoNBin}
\end{figure}
The hidden Markov model (\HMM)~\citep{rabiner89tutorial} is defined by a joint distribution $p(s_{1:T},v_{1:T})$ that factorizes as
\begin{align}
p(s_{1:T},v_{1:T})=p(v_1|s_1)p(s_1)\prod_{t=2}^T p(v_t|s_t)p(s_t|s_{t-1}),
\label{eq:HMM}
\end{align}
where the emission distribution $p(v_t|s_t)$ is time-homogeneous and, for continuous $v_t$, commonly modelled
as a Gaussian mixture. As discussed above, $s_{1:T}$ implicitly define a geometric duration distribution. A more flexible negative binomial duration distribution can be obtained by imposing a minimum duration $d_{\min}$ on the time spent in a regime \citep{durbin01biological}.
This can be achieved, \eg, by replacing the original regimes with $S$ ordered sets of regimes $R_i=\{(i-1)d_{\min}+1,\ldots,id_{\min}\}$, $i=1,\ldots,S$, where the elements of $R_i$ have the same emission distribution as original regime $i$
and transition distribution
\begin{align*}
&p(s_{t+1}|s_t\!\in\!R_i\!\setminus\! id_{\min})=
\begin{cases}
\pi_{ii} & \hskip-0.15cm\textrm{if } s_{t+1}\!=\!s_t\\
1\!-\!\pi_{ii} &  \hskip-0.15cm\textrm{if } s_{t+1}\!=\!s_t\!+\!1,
\end{cases}\\
&p(s_{t+1}|s_t\!=\!id_{\min})=
\begin{cases}
    \pi_{ii} & \hskip-0.15cm\textrm{if } s_{t+1}\!=\!s_t\\
	\pi_{ji} & \hskip-0.15cm\textrm{if } s_{t+1}\!=\!(j-1)d_{\min}+1, \hskip0.1cm j\!\neq\!i.
\end{cases}
\end{align*}
Given $s_t=i$, any sequence $s_{t+1},\ldots,s_{t+d-1}$ in $R_i$ such that $s_{t+d}\notin R_i$
has probability $\pi^{d-1-(d_{\min}-1)}_{ii}(1-\pi_{ii})^{d_{\min}-1}(1-\pi_{ii})$, and there are ${d-1 \choose d_{\min}-1}$ such sequences. Therefore
\begin{align*}
p(s_{t+1:t+d-1}\!\in\! R_i,s_{t+d}\!\notin\! R_i|s_t\!=\!i)={d\!-\!1 \choose d_{\min}\!-\!1}\pi^{d-d_{\min}}_{ii}(1\!-\!\pi_{ii})^{d_{\min}},
\end{align*}
which corresponds to the negative binomial distribution with parameters $\pi_{ii}$ and $d_{\min}$.
The negative binomial distribution with $\pi_{ii}\in\{0.1,0.5,0.9\}$ and $d_{\min}=5$ is shown in \figref{fig:GeoNBin}(b).

\section{Graphical Models and Belief Networks\label{sec:BN}}
\begin{figure}[t]
\begin{center}
\subfigure[]{\scalebox{0.8}{
\begin{tikzpicture}[dgraph]
\node[ocont] (x2) at (4.5,0) {$x_2$};
\node[ocont] (x1) at (1.5,0) {$x_1$};
\node[ocont] (x3) at (3,0) {$x_3$};
\node[ocont] (x4) at (3,-1.5) {$x_4$};
\draw[line width=1.15pt](x1)--(x3);\draw[line width=1.15pt](x2)--(x3);\draw[line width=1.15pt](x3)--(x4);\draw[line width=1.15pt](x2)to [bend left=35](x4);
\end{tikzpicture}}}
\hskip0.2cm
\subfigure[]{\scalebox{0.8}{
\begin{tikzpicture}[dgraph]
\node[ocont] (x2) at (4.5,0) {$x_2$};
\node[ocont] (x1) at (1.5,0) {$x_1$};
\node[ocont] (x3) at (3,0) {$x_3$};
\node[ocont] (x4) at (3,-1.5) {$x_4$};
\draw[line width=1.15pt](x1)--(x3);\draw[line width=1.15pt](x2)--(x3);\draw[line width=1.15pt](x3)--(x4);\draw[line width=1.15pt](x2)to [bend left=35](x4);\draw[line width=1.15pt](x4)to [bend left=35](x1);
\end{tikzpicture}}}
\hskip0.2cm
\subfigure[]{\scalebox{0.8}{
\begin{tikzpicture}[dgraph]
\node[] at (1.1,0) {$\cdots$};
\node[disc] (sigmatm) at (2,0) {$s_{t-1}$};
\node[disc] (sigmat) at (3.5,0) {$s_t$};
\node[disc] (sigmatp) at (5,0) {$s_{t+1}$};
\node[] at (5.9,0) {$\cdots$};
\node[ocont,obs] (vtm) at (2,-1.5) {$v_{t-1}$};
\node[ocont,obs] (vt) at (3.5,-1.5) {$v_t$};
\node[ocont,obs] (vtp) at (5,-1.5) {$v_{t+1}$};
\draw[line width=1.15pt](sigmatm)--(sigmat);\draw[line width=1.15pt](sigmat)--(sigmatp);
\draw[line width=1.15pt](sigmatm)--(vtm);\draw[line width=1.15pt](sigmat)--(vt);\draw[line width=1.15pt](sigmatp)--(vtp);
\end{tikzpicture}}}
\caption{(a): Directed acyclic graph. The node $x_3$ is a collider on the path $x_2,x_3,x_1$ and a non-collider on the path $x_2,x_3,x_4$. (b): Cyclic graph obtained from (a) by adding a link from $x_4$ to $x_1$.
(c): Belief network representation of the \HMM. Rectangular nodes indicate discrete variables, whilst oval nodes indicate discrete or continuous variables.
Filled nodes indicate observed variables. This convention is used throughout the manuscript.}
\label{fig:indep_HMM}
\end{center}
\end{figure}
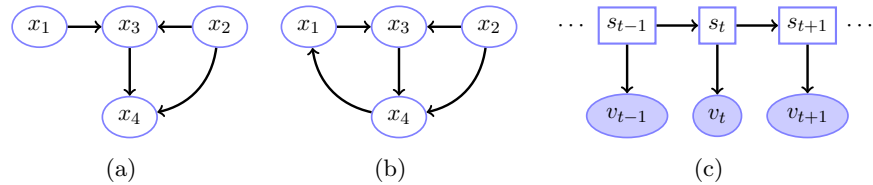
Graphical models \citep{pearl88probabilistic,bishop06pattern,kollerl09probabilistic,barber12bayesian,murphy12machine} are a marriage between graph and probability theory that allows
to graphically represent and assess statistical dependence, and therefore to easily describe the
structure of complex models and derive inference routines. \HMSMs~are most commonly described using a type of graphical models called belief networks.
In the following sections, we give some basic definitions and explain two equivalent methods for graphically assessing statistical independence in belief networks.

\subsubsection*{Basic definitions}
A {\bf graph} is a collection of nodes and links connecting pairs of nodes.
The links may be directed or undirected, giving rise to {\bf directed} or {\bf undirected graphs} respectively.\\[5pt]
A {\bf path} from node $x_i$ to node $x_j$ is a sequence of linked nodes starting
at $x_i$ and ending at $x_j$. A {\bf directed path} is a path whose links are directed and pointing from preceding towards following nodes in the sequence.\\[5pt]
A {\bf directed acyclic graph} is a directed graph with no
directed paths starting and ending at the same node. For example, the directed graph in \figref{fig:indep_HMM}(a) is acyclic. The addition of a link from $x_4$ to $x_1$ gives rise to a cyclic graph (\figref{fig:indep_HMM}(b)).\\[5pt]
A node $x_i$ with a directed link to $x_j$ is called {\bf parent}
of $x_j$. In this case, $x_j$ is called {\bf child} of $x_i$.\\[5pt]
A node is a {\bf collider} on a specified path if it has two parents on that path.
Notice that a node can be a collider on a path and a non-collider on another path. For example, in \figref{fig:indep_HMM}(a) $x_3$ is a collider on the path $x_2,x_3,x_1$ and a non-collider on the path $x_2,x_3,x_4$.\\[5pt]
A node $x_i$ is an {\bf ancestor} of a node $x_j$ if there exists a directed path from $x_i$ to $x_j$. In this case, $x_j$ is a {\bf descendant} of $x_i$.\\[5pt]
A {\bf graphical model} is a graph in which nodes represent random variables and links express statistical relationships between the variables.\\[5pt]
A {\bf belief network} is a directed acyclic graphical model in which each node $x_i$ is associated
with the conditional distribution $p(x_i|\text{par}(x_i))$, where $\text{par}(x_i)$ indicates the parents of $x_i$. The joint distribution of all nodes in
the graph, $p(x_{1:D})$, is given by the product of all conditional distributions, \ie
\begin{align*}
p(x_{1:D})=\prod_{i=1}^Dp(x_i|\text{par}(x_i)).
\end{align*}
The belief network corresponding to \eqref{eq:HMM}, and therefore representing the \HMM, is given in \figref{fig:indep_HMM}(c).
\subsubsection*{Assessing statistical independence in belief networks}
\begin{figure}[t]
\hskip-0.0cm
\subfigure[]{\scalebox{0.8}{
\begin{tikzpicture}[dgraph]
\node[] at (-0.8,0) {$\cdots$};
\node[disc] (sigmatm2) at (0,0) {$s_{t-2}$};
\node[disc] (sigmatm) at (2,0) {$s_{t-1}$};
\node[disc] (sigmat) at (4,0) {$s_t$};
\node[disc] (sigmatp) at (6,0) {$s_{t+1}$};
\node[] at (6.9,0) {$\cdots$};
\node[ocont,obs] (vtm2) at (0,-1.5) {$v_{t-2}$};
\node[ocont,obs] (vtm) at (2,-1.5) {$v_{t-1}$};
\node[ocont,obs] (vt) at (4,-1.5) {$v_t$};
\node[ocont,obs] (vtp) at (6,-1.5) {$v_{t+1}$};
\draw[line width=1.15pt](sigmatm2)--(sigmatm);
\draw[line width=1.15pt](sigmatm2)--(vtm2);
\draw[line width=1.15pt](vtm2)--(vtm);
\draw[line width=1.15pt](sigmatm)--(sigmat);\draw[line width=1.15pt](sigmat)--(sigmatp);
\draw[line width=1.15pt](sigmatm)--(vtm);\draw[line width=1.15pt](sigmat)--(vt);\draw[line width=1.15pt](sigmatp)--(vtp);
\draw[line width=1.15pt](vtm)--(vt);
\draw[line width=1.15pt](vt)--(vtp);
\end{tikzpicture}}}
\hskip0.0cm
\subfigure[]{\scalebox{0.8}{
\begin{tikzpicture}[ugraph]
\node[] at (1.2,0) {$\cdots$};
\node[disc] (sigmatm) at (2,0) {$s_{t-2}$};
\node[disc] (sigmat) at (4,0) {$s_{t-1}$};
\node[disc] (sigmatp) at (6,0) {$s_{t}$};
\node[ocont,obs] (vtm) at (2,-1.5) {$v_{t-2}$};
\node[ocont,obs] (vt) at (4,-1.5) {$v_{t-1}$};
\node[ocont,obs] (vtp) at (6,-1.5) {$v_{t}$};
\draw[line width=1.15pt](sigmatm)--(sigmat);\draw[line width=1.15pt](sigmat)--(sigmatp);
\draw[line width=1.15pt](sigmatm)--(vtm);\draw[line width=1.15pt](sigmat)--(vt);\draw[line width=1.15pt](sigmatp)--(vtp);
\draw[line width=1.15pt](sigmat)--(vtm);\draw[line width=1.15pt](sigmatp)--(vt);
\draw[line width=1.15pt](vtm)--(vt);
\draw[line width=1.15pt](vt)--(vtp);
\end{tikzpicture}}}
\caption{(a): Belief network representing the extension of the \HMM~in which the observations are related by first-order Markovian dependence, as indicated by the link from $v_{t-1}$ to $v_t$.
(b): Undirected graph obtained from the belief network in (a) after performing steps (IIa) and (IIb) with ${\cal X}=v_t, {\cal Y}=v_{1:t-2}$ and ${\cal Z}=\{s_t,v_{t-1}\}$.}
\label{fig:HMMMark}
\end{figure}
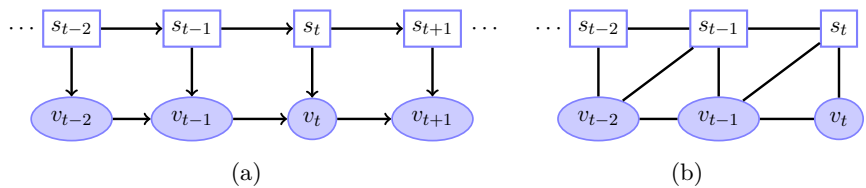

{\bf Method I.} Given the sets of random variables ${\cal X}, {\cal Y}$ and ${\cal Z}$,
${\cal X}$ and ${\cal Y}$ are statistically independent given ${\cal Z}$ (${\cal X} \ci {\cal Y} \,|\, {\cal Z}$) if all paths
from any element of ${\cal X}$ to any element of ${\cal Y}$ are blocked. A path is blocked if at least one of the following conditions is satisfied:
\begin{enumerate}
\item[(Ia)] There is a non-collider on the path which belongs to the conditioning set ${\cal Z}$.
\item[(Ib)] There is a collider on the path such that neither the collider nor any of its descendants belong to the conditioning set ${\cal Z}$.
\end{enumerate}
{\bf Method II.} This method converts the directed graph into an undirected one and then uses the rule of independence for undirected graphs. This is achieved with the following steps:
\begin{enumerate}
\item[(IIa)] Create the ancestral graph: Remove all nodes that are not in ${\cal X}\cup{\cal Y}\cup {\cal Z}$ and are not ancestors of a node in this set, together with all links in or out of such nodes.
\item[(IIb)] Moralize: Add a link between any two nodes that have a common child. Remove arrowheads.
\item[(IIc)] Use the independence rule for undirected graphs: ${\cal X}\ci{\cal Y} \,|\, {\cal Z}$ if all paths connecting a node in ${\cal X}$ with one in ${\cal Y}$
pass through a member of ${\cal Z}$.
\end{enumerate}
In \figref{fig:HMMMark}(b) we display the undirected graph obtained from the belief network shown in \figref{fig:HMMMark}(a) after performing steps (IIa) and (IIb) with ${\cal X}=v_t, {\cal Y}=v_{1:t-2}$ and ${\cal Z}=\{s_t,v_{t-1}\}$.

\subsection{Inference in \HMSMs\label{sec:ILEX}}
In this section we illustrate how the formalism of graphical models can be used to derive the standard inference recursions of \HMSMs.

We consider
the extension of the \HMM~in which the observations (given the regime variables) are related by $k$th-order Markovian dependence, \ie~the joint distribution factorizes as\footnote{We use the convention $x_{t}=\emptyset$ for ${t}\leq0$. The \HMM~can be obtained as a special case by setting $k=0$, with the convention $x_{t:t'}=\emptyset$ for $t>t'$.}
\begin{align*}
p(s_{1:T},v_{1:T})=\prod_{t=1}^T p(v_t|s_t,v_{t-k:t-1})p(s_t|s_{t-1}).
\end{align*}
The introduction of Markovian dependence corresponds to adding links from past to current observations to the belief network representing the \HMM,
as shown in \figref{fig:HMMMark}(a) for the case of first-order dependence.
The most popular model within this class is the switching autoregressive model \citep{hamilton89new,hamilton90analysis,hamilton93estimation},
also called autoregressive \HMM, defined as
\begin{align}
&p(v_t|s_t,v_{t-k:t-1})={\cal N}\Big(v_t;\sum_{i=1}^k a^{s_t}_i v_{t-i},(\sigma^{s_t})^2\Big),
\label{eq:sarm}
\end{align}
where ${\cal N}(x;\mu,\sigma^2)$ denotes a Gaussian distribution on variable $x$ with mean $\mu$ and variance $\sigma^2$, and $a^{s_t}_i$ is called autoregressive coefficient.

As discussed in \chapref{chap:intro}, these types of models are often used for \timeseries~segmentation. A segmentation can be obtained by computing $s^*_t=\argmax_{s_t}\alpha^{s_t}_t$ where $\alpha^{s_t}_t=p(s_t|v_{1:t})$ is called the filtered distribution, $s^*_t=\argmax_{s_t}\gamma^{s_t}_t$ where $\gamma^{s_t}_t=p(s_t|v_{1:T})$ is called the smoothed distribution, or the most likely sequence of regimes $s^*_{1:T}=\argmax_{s_{1:T}} p(s_{1:T}|v_{1:T})$. Unknown model parameters can be learned with similar quantities.
These quantities can be efficiently computed using time-recursive routines, namely routines which at each time-step make use of computations previously performed at the preceding or following time-step (\eg~$\alpha^{s_t}_t$ can be computed from $\alpha^{s_{t-1}}_{t-1}$ and $\gamma^{s_t}_t$ can be computed from $\gamma^{s_{t+1}}_{t+1}$).

In the following sections we describe the two most common approaches to compute the filtered and smoothed distributions,
namely parallel and sequential filtering-smoothing, and an extension of Viterbi decoding
for computing the most likely sequence of regimes. 

The approaches described can be applied to all models in which the unobserved variables form a first-order Markov chain -- including the case in which these variables are continuous, by replacing sums with integrations -- although computational tractability is not guaranteed.
As the explicit-duration \HMSMs~described in \chapref{chap:Dur} are extensions of the standard \HMSMs~in which the combined regime and explicit-duration variables, $\sigma_{1:T}$, form a first-order Markov chain,
we will be able to use similar approaches to derive inference recursions for $\sigma_{1:T}$, which will then be simplified using the deterministic constraints of the Markov chain.
Furthermore, as the continuous unobserved variables of the explicit-duration linear Gaussian state-space model described in \secref{sec:SLGSSM} are related by first-order Markovian dependence,
we will also be able to use similar approaches to derive inference recursions on these variables.
In the case considered in \secref{sec:M2Inf}, in which the \timeseries~is formed by segments whose observations are related by non-Markovian dependence, time-steps at the segment boundaries only will need to be considered, giving rise to segment-recursive routines.

\subsubsection*{Parallel filtering-smoothing}
The filtered distribution $\alpha^{s_t}_t=p(s_t|v_{1:t})$ can be obtained by normalizing $\bar\alpha^{s_t}_t=p(s_t,v_{1:t})$, where
$\bar\alpha^{s_t}_t$ can be recursively computed as\footnote{The initialization is given by $\bar\alpha^{s_1}_1=p(v_1|s_1)\tilde\pi_{s_1}$.}
\begin{align}
\bar\alpha^{s_t}_t&=p(v_t|s_t,\cancel{v_{1:t-k-1}},v_{t-k:t-1})\sum_{s_{t-1}}p(s_t|s_{t-1},\cancel{v_{1:t-1}})p(s_{t-1},v_{1:t-1})\nonumber\\[4pt]
&=p(v_t|s_t,v_{t-k:t-1})\sum_{s_{t-1}}\pi_{s_ts_{t-1}}\bar\alpha^{s_{t-1}}_{t-1}.\label{eq:alpha}
\end{align}
The independence relation $v_t\ci v_{1:t-k-1}\,|\,\{s_t,v_{t-k:t-1}\}$ can be graphically assessed by observing that
(considering, for simplicity, $k=1$) all paths from $v_{1:t-2}$ to $v_t$ are blocked, as $v_t$ is reached by passing from: (i) both $s_t$ and $v_{t-1}$, (ii) $s_t$ only, (iii) $v_{t-1}$ only (\figref{fig:HMMMark}(a)).
In cases (i) and (ii), $s_t$ is a non-collider on the path that belongs to the conditioning set.
In case (iii), $v_t$ is a non-collider on the path that belongs to the conditioning set. Therefore, in all cases condition (Ia) is satisfied\footnote{
Alternatively, we can apply steps (IIa) and (IIb) to the belief network shown in \figref{fig:HMMMark}(a), obtaining the undirected graph shown in \figref{fig:HMMMark}(b), and
observe that all paths from $v_{1:t-2}$ to $v_t$ pass through $s_t$ or $v_{t-1}$, which belong to the conditioning set.}.

\noindent The independence relation $s_t\ci v_{1:t-1}\,|\,s_{t-1}$
holds since all paths from $v_{1:t-1}$ to $s_t$ reach $s_t$ from: (i) the non-collider
$s_{t-1}$ that belongs to the conditioning set, (ii) the collider $v_t$ that (as well as all its
descendants) does not belong to the conditioning set, (iii) $s_{t+1}$ that imposes passing through
a collider (\eg~$v_{t+1}$) that (together with all its descendants) does not belong to the conditioning set.

\noindent The smoothed distribution $\gamma^{s_t}_t=p(s_t|v_{1:T})$ can be obtained as\footnote{The normalization term $p(v_{1:T})$ can be estimated as $p(v_{1:T})=\sum_{s_t} \bar\alpha^{s_t}_T$.}
\begin{align*}
\gamma^{s_t}_t&\propto p(s_t,v_{1:T})
=p(v_{t+1:T}|s_t,\cancel{v_{1:t-k}},v_{t-k+1:t})p(s_t,v_{1:t})=\beta^{s_t}_t\bar\alpha^{s_t}_t,
\end{align*}
where $\beta^{s_t}_t=p(v_{t+1:T}|s_t,v_{t-k+1:t})$ can be recursively computed as\footnote{The initialization is given by $\beta^{s_T}_T=1$.}
\begin{align}
\beta^{s_t}_t&=\sum_{s_{t+1}}p(v_{t+1:T}|\cancel{s_t},s_{t+1},v_{t-k+1:t})p(s_{t+1}|s_t,\cancel{v_{t-k+1:t}})\nonumber\\
&=\sum_{s_{t+1}}p(v_{t+2:T}|s_{t+1},\cancel{v_{t-k+1}},v_{t-k+2:t+1})p(v_{t+1}|s_{t+1},v_{t-k+1:t})\pi_{s_{t+1}s_t}\nonumber\\
&=\sum_{s_{t+1}}\beta^{s_{t+1}}_{t+1}p(v_{t+1}|s_{t+1},v_{t-k+1:t})\pi_{s_{t+1}s_t}.
\label{eq:beta}
\end{align}
Notice that recursions (\ref{eq:alpha}) and (\ref{eq:beta}) can be performed in parallel.
Neglecting the cost of estimating $p(v_t|s_t,v_{t-k:t-1})$, the recursions have computational cost ${\cal O}(TS^2)$.
In order to avoid numerical underflow or overflow, the computations are commonly performed in the log domain.

\subsubsection*{Sequential filtering-smoothing}
An alternative way of performing filtering-smoothing is to first compute the filtered distribution $\alpha^{s_t}_t=p(s_t|v_{1:t})$ as
\begin{align*}
\alpha^{s_t}_t&=\frac{p(s_t,v_t|v_{1:t-1})}{p(v_t|v_{1:t-1})}
=\frac{p(v_t|s_t,v_{t-k:t-1})\sum_{s_{t-1}}\pi_{s_ts_{t-1}}\alpha^{s_{t-1}}_{t-1}}{\sum_{\tilde{s}_t}p(v_t|\tilde{s}_t,v_{t-k:t-1})\sum_{s_{t-1}}\pi_{\tilde{s}_ts_{t-1}}\alpha^{s_{t-1}}_{t-1}},
\end{align*}
and then compute the smoothed distribution $\gamma^{s_t}_t=p(s_t|v_{1:T})$ as
\begin{align}
\gamma^{s_t}_t&=\sum_{s_{t+1}}p(s_t|s_{t+1},v_{1:t},\cancel{v_{t+1:T}})p(s_{t+1}|v_{1:T})\nonumber\\
&=\sum_{s_{t+1}}\frac{p(s_{t+1}|s_t,\cancel{v_{1:t}})p(s_t|v_{1:t})}{\sum_{\tilde s_t}p(s_{t+1}|\tilde s_t,\cancel{v_{1:t}})p(\tilde s_t|v_{1:t})}\gamma^{s_{t+1}}_{t+1}\nonumber\\
&=\sum_{s_{t+1}}\frac{\pi_{s_{t+1}s_t}\alpha^{s_t}_t}{\sum_{\tilde s_t}\pi_{s_{t+1}\tilde s_t}\alpha^{\tilde s_t}_t}\gamma^{s_{t+1}}_{t+1}.
\label{eq:gamma}
\end{align}
These routines do not require working in the log domain.

\subsubsection*{Extended Viterbi}
With the definition $\xi^{s_t}_t=\max_{s_{1:t-1}} p(s_{1:t},v_{1:t})$, the most likely sequence of regimes $s^*_{1:T}=\argmax_{s_{1:T}} p(s_{1:T}|v_{1:T})$ can be obtained with the following extension of the Viterbi algorithm \citep{rabiner89tutorial}:
\begin{align*}
&\xi^{s_1}_1=p(s_1,v_1)=\bar\alpha^{s_1}_1\\
&\textrm{for } t=2,\ldots,T\\
& \hspace{0.55cm}\xi^{s_t}_t=p(v_t|s_t,v_{t-k:t-1})\max_{s_{t-1}} \pi_{s_ts_{t-1}}\xi^{s_{t-1}}_{t-1},
 \hspace{0.2cm}\psi^{s_t}_t=\argmax_{s_{t-1}} \pi_{s_ts_{t-1}}\xi^{s_{t-1}}_{t-1}\\[-1pt]
&s^*_T=\argmax_{s_T} \xi^{s_T}_T\\
&\textrm{for } t=T\!-\!1,\ldots,1\\
&\hspace{0.55cm} s^*_t=\psi^{s^*_{t+1}}_{t+1},
\end{align*}
where the recursion for $\xi^{s_t}_t$ is obtained as the recursion for $\bar\alpha^{s_t}_t$ with the sum replaced by the max operator. 

\section{Expectation Maximization\label{sec:EM}}
The expectation maximization (EM) algorithm \citep{dempester77maximum,mclachlan08em} is a popular iterative approach for parameter estimation in probabilistic models with unobserved variables.
From a modern variational viewpoint \citep{bishop06pattern,barber12bayesian}, EM replaces the maximization of the log-likelihood $\log p({\cal V}|\theta)$ of observations ${\cal V}$,
in which summation/integration over the unobserved variables ${\cal H}$ couples parameters $\theta$, with the maximization of a lower bound that has a decoupled form in the parameters
$\theta_{\cal V}$ and $\theta_{\cal H}$ corresponding to observed and unobserved variables respectively. More specifically, consider the distribution $q$ and the Kullback-Leibler (KL) divergence
\begin{align*}
KL(q(H|{\cal V})||p({\cal H}|{\cal V},\theta))&=\av{\log q({\cal H}|{\cal V})-\log \frac{p({\cal H},{\cal V}|\theta)}{p({\cal V}|\theta)}}_{q({\cal H}|{\cal V})}\,,
\end{align*}
where $\av{\cdot}_{q}$ indicates averaging with respect to $q$. As the KL divergence is always nonnegative, $\log p({\cal V}|\theta)$ can be lower-bounded as
\begin{align*}
&\log p({\cal V}|\theta)\geq \underbrace{-\av{\log q({\cal H}|{\cal V})}_{q({\cal H}|{\cal V})}}_{\textrm{Entropy}}+\underbrace{\av{\log p({\cal H},{\cal V}|\theta)}_{q({\cal H}|{\cal V})}}_{\text{Energy}}.
\end{align*}
For $q({\cal H}|{\cal V})=p({\cal H}|{\cal V},\bar\theta)$, where $\bar\theta$ is a fixed set of parameters, the entropy does not depend on $\theta$ and the energy, also called expectation of the complete data log-likelihood, has a decoupled form, \ie
\begin{align*}
\!\av{\log p({\cal H},\!{\cal V}|\theta)}_{p({\cal H}|{\cal V},\theta^{k})}\!=\!\av{\log p({\cal V}|{\cal H},\!\theta_{\cal V})}_{p({\cal H}|{\cal V},\theta^{k})}\!\!+\!\av{\log p({\cal H}|\theta_{\cal H})}_{p({\cal H}|{\cal V},\theta^{k})}.
\end{align*}
At iteration $k$ of EM, the following two steps are performed:
\begin{itemize}
\item E-step: Compute the marginal distributions of $p({\cal H}|{\cal V},\theta^{k-1})$ required to carry out the M-step, where $\theta^{k-1}$ is the set of parameters estimated at iteration $k-1$.
\item M-step: Compute $\theta^{k}=\argmax_{\theta}\av{\log p({\cal H},{\cal V}|\theta)}_{p({\cal H}|{\cal V},\theta^{k-1})}$.
\end{itemize}
At each iteration, the log-likelihood is guaranteed not to decrease. Indeed
\begin{align*}
\log p({\cal V}|\theta^{k})-\log p({\cal V}|\theta^{k-1})&=\text{KL}(p({\cal H}|{\cal V},\theta^{k-1})||p({\cal H}|{\cal V},\theta^{k}))\\
&+\av{\log p({\cal H},{\cal V}|\theta^{k})}_{p({\cal H}|{\cal V},\theta^{k-1})}\\
&-\av{\log p({\cal H},{\cal V}|\theta^{k-1})}_{p({\cal H}|{\cal V},\theta^{k-1})}\\
&\geq 0,
\end{align*}
as the KL divergence is always nonnegative and, by construction, $\av{\log p({\cal H},{\cal V}|\theta^{k})}_{p({\cal H}|{\cal V},\theta^{k-1})}\geq \av{\log p({\cal H},{\cal V}|\theta)}_{p({\cal H}|{\cal V},\theta^{k-1})}$ for all $\theta$, and therefore also for $\theta^{k-1}$.
Under general conditions, this iterative approach is guaranteed to converge to a local maximum of $\log p({\cal V}|\theta)$.

In Appendix~\ref{app:SARM} we show how to apply the EM algorithm to learn the parameters of the switching autoregressive model (\ref{eq:sarm}).

\chapter{Explicit-Duration Modelling\label{chap:Dur}}
In \chapref{chap:Back} we have shown that, in standard \HMSMs, the regime variables implicitly define a geometric duration distribution,
and that a negative binomial duration distribution can be obtained with regime copies.
Explicit-duration \HMSMs~use extra unobserved variables to explicitly model the duration distribution, such that duration distributions of any form
can be defined. Additionally, explicit-duration variables give the possibility to impose complex dependence between the observations and to reset the dynamics to initial conditions.
In this chapter we describe the different ways in which explicit-duration modelling can be achieved,
and analyse their characteristics in models of simple unobserved structure and in models with extra unobserved variables related by first-order Markovian dependence
(the case of more complex unobserved structure is not considered).

Explicit-duration variables influence the time spent in a regime by allowing $s_t$ to differ from $s_{t-1}$ (through sampling from the transition distribution $\pi_{s_ts_{t-1}}$)
only if the variables take certain values, and by forcing $s_t$ to be equal to $s_{t-1}$ otherwise.
This is achieved by defining a first-order Markov chain on the combined regime and explicit-duration variables $\sigma_{1:T}$. Realizations of the chain
partition the \timeseries~into segments, with boundaries at those time-steps in which sampling occurs and with durations
distributed according to specified segment-duration distributions.

If $\pi_{s_ts_t}=0$, as it is most commonly assumed, a segment begins when a change of regime occurs;
whilst if $\pi_{s_ts_t}\neq 0$, as it may be desirable or required for certain tasks (\eg~for detecting changepoints or for identifying repetitions of patterns),
segment beginnings do not coincide with regime changes.

The first-order Markov chain on $\sigma_{1:T}$ can be defined using three fundamentally different encodings for the explicit-duration variables.
More specifically, we can encode distance to current-segment end using count variables $c_{1:T}$ that decrease within a segment; distance to current-segment beginning
using count variables $c_{1:T}$ that increase within a segment; or distance to both current-segment beginning and current-segment end using decreasing or increasing count variables
and duration variables $d_{1:T}$ indicating current-segment duration.

Different encoding leads to different possible structures for the distribution $p(v_{1:T}|\sigma_{1:T})$. 
More specifically, increasing count variables and count-duration variables always enable the factorization of $p(v_{1:T}|\sigma_{1:T})$ across segments (across-segment independence).
Furthermore, count-duration variables allow any structure within a segment, as segment-recursive inference can be performed;
whist count variables only allow a distribution that can be efficiently computed as (omitting conditioning on $\sigma_{1:T}$)
$\prod_{t} p(v_t|v_{1:t-1})$, as only time-recursive inference can be performed.
Examples of models with distributions that can be efficiently computed as $\prod_{t} p(v_t|v_{1:t-1})$ are the explicit-duration extensions of the \HMSMs~analysed in \secref{sec:ILEX},
the explicit-duration extension of the switching linear Gaussian state-space model described in \secref{sec:SLGSSM} -- in which the Markovian structure of the hidden dynamics $h_{1:T}$ enables time-recursive computation of
$p(v_t|v_{1:t-1})$,
and the model in \cite{fearnhead09bayesian} -- in which observations $v_{t-d+1:t}$ forming a segment generated by regime $j$ are linked through integration over parameters $\theta^j$,
\ie~$p(v_{t-d+1:t})=\int p(\theta^j)\prod_{\tau=t-d+1}^t p(v_{\tau}|\theta^j)d\theta^j$,
so that $p(v_{\tau}|v_{t-d+1:{\tau}-1})=p(v_{t-d+1:{\tau}})/p(v_{t-d+1:{\tau}-1})$.

In models with extra unobserved variables related by first-order Markovian dependence in addition to $\sigma_{1:T}$, for which inference is more complex, different encoding leads to different computational cost and, potentially, to different approximation requirements.

Taking the viewpoint in \citep{murphy2002hsm}, the original and currently standard approach to explicit-duration modelling \citep{ferguson80variable,rabiner89tutorial,ostendorf96from,yu10hidden} considers duration variables $d_{1:T}$
and variables $c_{1:T}$ such that, \eg, $c_t=1$ at the end of the segment and $c_t=2$ otherwise.
These variables can be seen as collapsed count variables that encode information about \emph{whether} (rather than \emph{where}) the segment is ending,
such that information about segment beginning and segment end is available only at the end of the segment.
Therefore this approach is a special case of the count-duration-variable approach. The possible structures for $p(v_{1:T}|\sigma_{1:T})$
are the same as with count-duration variables. However, as $\sigma_{1:T}$ do not form a first-order Markov chain, deriving
posterior distributions of interest is less immediate than with count-duration variables.
All other approaches to explicit-duration modelling in the literature use explicit-duration variables that encode the same information about segment boundaries as
decreasing count variables, increasing count variables or count-duration variables, although the parametrizations can be different.

The remainder of the chapter is organized as follows. In \secref{sec:M1}, \secref{sec:M1alt} and \secref{sec:M2}, we describe the three approaches to explicit-duration modelling in generality. In \secref{sec:VVM}
we analyse in detail explicit-duration modelling for \HMSMs~with simple unobserved structure. In \secref{sec:SLGSSM}, we analyse in detail explicit-duration modelling for the more complex switching linear Gaussian state-space model,
using an approach to inference that allows to understand how the results generalize to
similar models with extra unobserved variables related by first-order Markovian dependence. In \secref{sec:Approx} we discuss approximation schemes for reducing the computational cost of inference.
We focus our exposition on parametric segment-duration distributions defined on the set $\{d_{\min},\ldots,d_{\max}\}$ (for simplicity, we assume $d_{\min}$ and $d_{\max}$ to be regime-independent).
The computational cost of inference is computed assuming $d_{\min}=1$.

\section{Decreasing Count Variables \label{sec:M1}}
This approach uses variables $c_{1:T}$ taking decreasing values within a segment, starting from the segment duration and ending with 1.
Therefore, $c_t$ indicates that the current segment ends at time-step $t+c_t-1$.
More specifically, the joint distribution $p(\sigma_{1:T})$, where $\sigma_t=(s_t,c_t)$, has
the following first-order Markovian structure:
\begin{align*}
p(\sigma_{1:T})\!=\!p(\sigma_1)\prod_{t=2}^T p(\sigma_t|\sigma_{t-1})=p(c_1|s_1)p(s_1)\prod_{t=1}^T p(c_t|s_t,c_{t-1})p(s_t|\sigma_{t-1}),
\end{align*}
with\footnote{The term $\delta_{x=y}$ has value 1 if $x=y$ and 0 otherwise.}
\begin{align*}
p(s_t|\sigma_{t-1})\!=\!\begin{cases}
	\pi_{s_ts_{t-1}}      & \hskip-0.3cm \textrm{if } c_{t-1}\!=\!1\\
	\delta_{s_t=s_{t-1}}  & \hskip-0.3cm \textrm{if } c_{t-1}\!>\!1,
\end{cases}
\hskip0.07cm
p(c_t|s_t,c_{t-1})\!=\!\begin{cases}
	\rho_{\sigma_t} 		& \hskip-0.3cm \textrm{if } c_{t-1}\!=\!1\\
	\delta_{c_t=c_{t-1}-1} 	& \hskip-0.3cm \textrm{if } c_{t-1}\!>\!1,
\end{cases}
\end{align*}
$p(s_1)=\tilde\pi_{s_1}$, and $p(c_1|s_1)=\tilde\rho_{\sigma_1}$, and where $\tilde\rho$ and $\rho$ are a vector and a matrix that specify the segment-duration distribution
on the set $\{d_{\min},\ldots,d_{\max}\}$. We assume $d_{\max}<T$, unless otherwise specified. 
In this encoding, we can impose that the last segment ends at the last time-step ($c_T=1$) with a time-dependent $\rho$, or condition inference on this event.
In this case, only $c_t\leq \min(T-t+1,d_{\max})$ needs to be considered.
This constraint is necessary if $d_{\max}=\infty$.
We cannot impose that the first segment starts at the first time-step, nor condition inference on this event.

Notice that $c_{t-1}>1$ implies $\sigma_{t}=(s_{t-1},c_{t-1}-1)$, \ie~$p(\sigma_{t}=(s_{t-1},c_{t-1}-1)|s_{t-1},c_{t-1}>1)=1$ (also when conditioning on the observations).
We will make extensive use of this result in \secref{sec:SLGSSMDec}.%

\section{Increasing Count Variables \label{sec:M1alt}}
This approach uses variables $c_{1:T}$ taking increasing values within a segment, starting from 1 and ending with the segment duration.
Therefore, $c_t$ indicates that the current segment begins at time-step $t-c_t+1$.
More specifically, the joint distribution $p(\sigma_{1:T})$, where $\sigma_t=(s_t,c_t)$, has
the following first-order Markovian structure:
\begin{align*}
p(\sigma_{1:T})\!=\!p(\sigma_1)\prod_{t=2}^T p(\sigma_t|\sigma_{t-1})=p(c_1|s_1)p(s_1)\prod_{t=2}^T p(s_t|s_{t-1},c_t)p(c_t|\sigma_{t-1}),
\end{align*}
with
\begin{align*}
p(s_t|s_{t-1},c_t)\!=\!\begin{cases}
	\pi_{s_ts_{t-1}} 	& \hskip-0.3cm \textrm{if } c_t\!=\!1\\
	\delta_{s_t=s_{t-1}} & \hskip-0.3cm \textrm{if } c_t\!>\!1,
\end{cases}
\hskip0.17cm
p(c_t|\sigma_{t-1})\!=\!\begin{cases}
	\lambda_{\sigma_{t-1}} & \hskip-0.3cm\textrm{if } c_t\!=\!c_{t-1}\!+\!1\\	
	1\!-\!\lambda_{\sigma_{t-1}} & \hskip-0.3cm  \textrm{if } c_t\!=\!1,
\end{cases}
\end{align*}
$p(s_1)=\tilde \pi_{s_1}$, and $p(c_1|s_1)=\tilde\lambda_{\sigma_1}$, and where $\lambda_{\sigma_t}=0$ for $c_t\geq d_{\max}$, and $\lambda_{\sigma_t}=1$ for $c_t<d_{\min}$.
For simplicity, consider the case in which $\lambda_{\sigma_t}$ depends on the count variable only ($\lambda_{\sigma_t}=\lambda_{c_t}$).
The probability that a segment, starting at time-step $t$, ends
at time-step $t+d-1$ (\ie~the probability of segment duration $d$) is
\begin{align*}
p(c_{t+1:t+d-1}=2,\ldots,d,c_{t+d}\!=\!1|c_t\!=\!1)\!=\!
\begin{cases}
(1\!-\!\lambda_{d})\prod_{k=1}^{d-1}\lambda_{k} & \hskip-0.25cm\textrm{if } d<d_{\max}\\
\prod_{k=1}^{d-1}\lambda_{k} & \hskip-0.25cm\textrm{if } d=d_{\max}.
\end{cases}
\end{align*}
The term $\lambda_d$ represents the probability of segment duration $> d$, given segment duration $\geq d$.
Indeed
\begin{align*}
\frac{p(c_{t+1:t+d}=2,\ldots,d+1|c_t=1)}{p(c_{t+1:t+d-1}=2,\ldots,d|c_t=1)}=\frac{\prod_{k=1}^{d}\lambda_k}{\prod_{k=1}^{d-1}\lambda_k}=\lambda_d\,.
\end{align*}
Therefore, the term $1-\lambda_d$ represents the probability of segment duration $d$, given segment duration $\geq d$.
The relation between $\lambda_d$ and the segment-duration distribution in \secref{sec:M1} is given by
\begin{align*}
\lambda_d=\frac{1-\sum_{k=1}^{d}\rho_k}{1-\sum_{k=1}^{d-1}\rho_k}=
\frac{\sum_{k=d+1}^{d_{\max}}\rho_k}{\sum_{k=d}^{d_{\max}}\rho_k}=1-\frac{\rho_d}{\sum_{k=d}^{d_{\max}}\rho_k}.
\end{align*}
The term $\tilde\lambda_{c_1}$ represents the probability that the first segment starts at time-step $2-c_1$.
Therefore, we can impose that the first segment starts at the first time-step ($c_1=1$) by setting $\tilde\lambda_1=1$.
In this case, $p(c_t>t)=0$ and thus only $c_t\leq \min(t,d_{\max})$ needs to be considered.
This constraint is necessary if $d_{\max}=\infty$ (\eg~if $\lambda_{\sigma_{t-1}}$ does not depend on $c_{t-1}$, which corresponds to a geometric segment-duration distribution).
In this encoding we cannot impose that the last segment ends at the last time-step, nor condition inference on this event.

Notice that $c_t>1$ implies $\sigma_{t-1}=(s_t,c_t-1)$, \ie~$p(\sigma_{t-1}=(s_t,c_t-1)|s_t,c_t>1)=1$. We will make extensive use of this result in \secref{sec:SLGSSMInc}.

\section{Count-Duration Variables \label{sec:M2}}
This approach uses either decreasing or increasing
count variables $c_{1:T}$, and duration variables $d_{1:T}$ indicating the duration of the current segment.
With decreasing count variables, $(c_t,d_t)$ indicates that the current segment starts at time-step $t-d_t+c_t$ and ends at time-step $t+c_t-1$.
More specifically, the joint distribution $p(\sigma_{1:T})$, where $\sigma_t=(s_t,d_t,c_t)$, has
the following first-order Markovian structure:
\begin{align*}
&p(\sigma_{1:T})=p(\sigma_1)\prod_{t=2}^T p(\sigma_t|\sigma_{t-1})\\
&=p(c_1|d_1)p(d_1|s_1)p(s_1)\prod_{t=2}^T p(c_t|d_t,c_{t-1})p(d_t|d_{t-1},c_{t-1})p(s_t|s_{t-1},c_{t-1}),
\end{align*}
with
\begin{align*}
p(s_t|s_{t-1},c_{t-1})&=\begin{cases}
	\pi_{s_ts_{t-1}} 	 & \hskip0.07cm\textrm{if } c_{t-1}\!=\!1\\
	\delta_{s_t=s_{t-1}} & \hskip0.07cm\textrm{if } c_{t-1}\!>\!1,
\end{cases}\\[2pt]
p(d_t|d_{t-1},c_{t-1},s_t)&=\begin{cases}
	\rho_{s_td_t} 	      &\hskip0.1cm\textrm{if } c_{t-1}\!=\!1\\
	\delta_{d_t=d_{t-1}}  & \hskip0.1cm\textrm{if } c_{t-1}\!>\!1,
\end{cases}\\[2pt]
p(c_t|c_{t-1},d_t)&=\begin{cases}
	\delta_{c_t=d_t} 		& \hskip-0.2cm\textrm{if } c_{t-1}\!=\!1\\
	\delta_{c_t=c_{t-1}-1} 	& \hskip-0.2cm\textrm{if } c_{t-1}\!>\!1,
\end{cases}
\end{align*}
$p(s_1)=\tilde{\pi}_{s_1}, p(d_1|s_1)=\tilde\rho_{s_1d_1}$, and $p(c_1|d_1)=\tilde{\tilde{\rho}}_{d_1c_1}$.

The term $\tilde{\tilde{\rho}}_{d_1c_1}$ represents the probability that the first segment of duration $d_1$ ends at time-step $c_1$.
Therefore, we can impose that the first segment starts at the first time-step by setting $\tilde{\tilde{\rho}}_{d_1d_1}=1$. In this case,
$p(d_t>t,c_t=1)=0$ and thus only $d_t\leq\min(t,d_{\max})$ needs to be considered. This constraint is necessary if $d_{\max}=\infty$.
We can impose that the last segment ends at the last time-step ($c_T=1$) with a time-dependent $\rho$, or condition inference on this event.

Notice that $c_t<d_t$ implies $\sigma_{t-1}=(s_t,d_t,c_t+1)$,
\ie~$p(\sigma_{t-1}=(s_t,d_t,c_t+1)|s_t,d_t,c_t<d_t)=1$.
In addition, $c_{t-1}>1$ implies $\sigma_{t}=(s_{t-1},d_{t-1},c_{t-1}-1)$, \ie~$p(\sigma_{t}=(s_{t-1},d_{t-1},c_{t-1}-1)|s_{t-1},d_{t-1},c_{t-1}>1)=1$.
We will make extensive use of this result in \secref{sec:SLGSSMCD} and Appendix~\ref{app:SS}.

\section{Explicit-Duration \HMSMs~$p(\sigma_{1:T},v_{1:T})$\label{sec:VVM}}
In this section, we describe explicit-duration modelling for \HMSMs~that contain only regime and explicit-duration variables $\sigma_{1:T}$ and observations $v_{1:T}$\footnote{In Appendix~\ref{app:HMM} we show that, if a geometric duration distribution is used,
a model which is similar to the standard \HMM~is retrieved.}. Models with additional unobserved variables that are independent can be treated similarly.

\subsection{Decreasing Count Variables \label{sec:M1Inf}}
\begin{figure}[t]
\begin{center}
\scalebox{0.85}{
\begin{tikzpicture}[dgraph]
\node[] at (1,0) {$\cdots$};
\node[disc] (sigmatm) at (2,0) {$c_{t-1}$};
\node[disc] (sigmat) at (4,0) {$c_t$};
\node[disc] (sigmatp) at (6,0) {$c_{t+1}$};
\node[] at (7,0) {$\cdots$};
\node[disc] (stm) at (2,-1.25) {$s_{t-1}$};
\node[disc] (st) at (4,-1.25) {$s_t$};
\node[disc] (stp) at (6,-1.25) {$s_{t+1}$};
\node[ocont,obs] (vtm) at (2,-2.5) {$v_{t-1}$};
\node[ocont,obs] (vt) at (4,-2.5) {$v_t$};
\node[ocont,obs] (vtp) at (6,-2.5) {$v_{t+1}$};
\draw[line width=1.15pt](sigmatm)--(sigmat);\draw[line width=1.15pt](sigmat)--(sigmatp);
\draw[line width=1.15pt](stm)--(sigmatm);\draw[line width=1.15pt](st)--(sigmat);\draw[line width=1.15pt](stp)--(sigmatp);
\draw[line width=1.15pt](stm)--(st);\draw[line width=1.15pt](st)--(stp);
\draw[line width=1.15pt](sigmatm)--(st);\draw[line width=1.15pt](sigmat)--(stp);
\draw[line width=1.15pt](stm)--(vtm);\draw[line width=1.15pt](st)--(vt);\draw[line width=1.15pt](stp)--(vtp);
\draw[line width=1.15pt](vtm)--(vt);
\draw[line width=1.15pt](vt)--(vtp);
\end{tikzpicture}}
\end{center}
\caption{\HMSM~in which the segment-duration distribution is explicitly modelled using decreasing count variables $c_{1:T}$.}
\label{fig:M1st}
\end{figure}
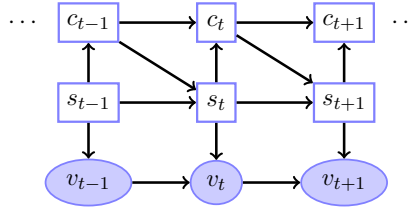

As explained above, decreasing count variables allow a distribution $p(v_{1:T}|\sigma_{1:T})$ that can be efficiently computed as $\prod_{t} p(v_t|\sigma_t,v_{1:t-1})$.
For models that contain only $\sigma_{1:T}$ and $v_{1:T}$, this translates into Markovian dependence between the observations.
This type of models is represented by the belief network shown in \figref{fig:M1st} (for Markovian order $k=1$). Dependence across segments can be cut only for $k=1$ with a link from $c_{t-1}$ to $v_t$.
In the following sections we derive inference recursions by using the approach described in \secref{sec:ILEX} and by exploiting the deterministic part of the first-order Markov chain formed by $\sigma_{1:T}$
to obtain simplifications.

\subsubsection*{Parallel filtering-smoothing}
The filtered distribution $\alpha^{\sigma_t}_t=p(\sigma_t|v_{1:t})$ can be obtained by normalizing $\bar\alpha^{\sigma_t}_t=p(\sigma_t,v_{1:t})$, where $\bar\alpha^{\sigma_t}_t$ can be computed
as\footnote{The initialization is given by $\bar\alpha^{\sigma_1}_1=p(v_1|s_1)\tilde\pi_{s_1}\tilde\rho_{\sigma_1}$.}
\begin{align}
\hskip-0.15cm\bar\alpha^{\sigma_t}_t&=p(v_t|s_t,\cancel{c_t},\cancel{v_{1:t-k-1}},v_{t-k:t-1})\sum_{\sigma_{t-1}}p(\sigma_t|\sigma_{t-1},\cancel{v_{1:t-1}})p(\sigma_{t-1},v_{1:t-1})\nonumber\\[3pt]
&=p(v_t|s_t,v_{t-k:t-1})\bigg\{\delta_{\begin{subarray}{l} c_t<d_{\max} \\ s_{t-1}=s_t \\ c_{t-1}=c_t+1 \end{subarray}}
\hskip-0.15cm+\delta_{\begin{subarray}{l} c_t\geq d_{\min} \\ c_{t-1}=1 \end{subarray}}
\rho_{\sigma_t}\hskip-0.0cm\sum_{s_{t-1}}\hskip-0.0cm\pi_{s_ts_{t-1}}\bigg\}\bar\alpha^{\sigma_{t-1}}_{t-1}.\hskip-0.05cm
\label{eq:alphaM1st}
\end{align}
With pre-computation of $\sum_{s_{t-1}}\pi_{s_ts_{t-1}}\bar\alpha^{s_{t-1},1}_{t-1}$, which does not depend on $c_t$,
this recursion has computational cost ${\cal O}(TS(S+Ed_{\max}))$,
where $E$ is the cost of computing $e^{s_t}_t=p(v_t|s_t,v_{t-k:t-1})$. 

\noindent The smoothed distribution $\gamma^{\sigma_t}_t=p(\sigma_t|v_{1:T})$ can be obtained as $\gamma^{\sigma_t}_t\propto p(v_{t+1:T}|\sigma_t,\cancel{v_{1:t-k}},v_{t-k+1:t})p(\sigma_t,v_{1:t})
=\beta^{\sigma_t}_t\bar\alpha^{\sigma_t}_t$, where $\beta^{\sigma_t}_t=p(v_{t+1:T}|\sigma_t,v_{t-k+1:t})$ can be computed as\footnote{The initialization is given by $\beta^{\sigma_T}_T=1$.
Setting $\beta^{\sigma_T}_T=0$ for $c_T>1$ corresponds to conditioning inference on the event $c_T=1$. In this case, only
$c_t\leq\min(T-t+1,d_{\max})$ needs to be considered.}
\begin{align*}
\beta^{\sigma_t}_t&=\sum_{\sigma_{t+1}}p(v_{t+1:T}|\cancel{\sigma_t},\sigma_{t+1},v_{t-k+1:t})p(\sigma_{t+1}|\sigma_t,\cancel{v_{t-k+1:t}})\\
&=\sum_{\sigma_{t+1}}p(v_{t+2:T}|\sigma_{t+1},\cancel{v_{t-k+1}},v_{t-k+2:t+1})p(v_{t+1}|s_{t+1},\cancel{c_{t+1}},v_{t-k+1:t})\\
&\times p(\sigma_{t+1}|\sigma_t)\\
&=\delta_{\begin{subarray}{l} c_t>1  \end{subarray}} e^{s_t}_{t+1}\beta^{s_t,c_t-1}_{t+1}
+\delta_{\begin{subarray}{l} c_t=1 \end{subarray}}\hskip-0.0cm\sum_{s_{t+1}}\hskip-0.0cm e^{s_{t+1}}_{t+1}\pi_{s_{t+1}s_t}\sum_{c_{t+1}}
\rho_{\sigma_{t+1}}\beta^{\sigma_{t+1}}_{t+1}.
\end{align*}
With pre-computation of $\sum_{c_{t+1}}\rho_{\sigma_{t+1}}\beta^{\sigma_{t+1}}_{t+1}$, which does not depend on $s_t$, this recursion has cost ${\cal O}(TS(S+d_{\max}))$.

\subsubsection*{Sequential filtering-smoothing}
The filtered distribution $\alpha^{\sigma_t}_t=p(\sigma_t|v_{1:t})$ can be computed as
\begin{align*}
\alpha^{\sigma_t}_t&=\frac{p(\sigma_t,v_t|v_{1:t-1})}{p(v_t|v_{1:t-1})}\\
&\propto p(v_t|s_t,\cancel{c_t},\cancel{v_{1:t-k-1}},v_{t-k:t-1})\sum_{\sigma_{t-1}}p(\sigma_t|\sigma_{t-1},\cancel{v_{1:t-1}})p(\sigma_{t-1}|v_{1:t-1})\\
&=p(v_t|s_t,v_{t-k:t-1})\bigg\{\delta_{\begin{subarray}{l} c_t<d_{\max} \\ s_{t-1}=s_t \\ c_{t-1}=c_t+1 \end{subarray}}
\hskip-0.15cm+\delta_{\begin{subarray}{l} c_t\geq d_{\min} \\ c_{t-1}=1 \end{subarray}}
\rho_{\sigma_t}\hskip-0.0cm\sum_{s_{t-1}}\hskip-0.0cm\pi_{s_ts_{t-1}}\bigg\}\alpha^{\sigma_{t-1}}_{t-1}.
\end{align*}
With pre-summation over $s_{t-1}$ as in recursion (\ref{eq:alphaM1st}), this recursion has cost ${\cal O}(TS(S+Ed_{\max}))$. 

The smoothed distribution $\gamma^{\sigma_t}_t=p(\sigma_t|v_{1:T})$ can be computed as
\begin{align*}
\hskip-0.1cm&\gamma^{\sigma_t}_t=\sum_{\sigma_{t+1}}p(\sigma_t|\sigma_{t+1},v_{1:t},\cancel{v_{t+1:T}})p(\sigma_{t+1}|v_{1:T})\nonumber\\
&=\sum_{\sigma_{t+1}}\frac{p(\sigma_{t+1}|\sigma_t,\cancel{v_{1:t}})p(\sigma_t|v_{1:t})}{\sum_{\tilde\sigma_t}p(\sigma_{t+1}|\tilde\sigma_t,\cancel{v_{1:t}})p(\tilde\sigma_t|v_{1:t})}\gamma^{\sigma_{t+1}}_{t+1}\nonumber\\
&=\delta_{\begin{subarray}{l} c_t>1 \end{subarray}}\frac{\alpha^{\sigma_t}_t\gamma^{s_t,c_t-1}_{t+1}}{\alpha^{\sigma_t}_t\!+\!\delta_{\begin{subarray}{l} c_t>d_{\min} \end{subarray}}\rho_{s_tc_t-1}\sum_{\tilde s_t}\pi_{s_t\tilde s_t}\alpha^{\tilde s_t,1}_t}\\
&\!+\!\delta_{\begin{subarray}{l} c_t=1 \end{subarray}}\alpha^{s_t,1}_t
\sum_{s_{t+1}}\hskip-0.0cm\!\pi_{s_{t+1}s_t}\!\sum_{c_{t+1}}\!\frac{\rho_{\sigma_{t+1}}\gamma^{\sigma_{t+1}}_{t+1}}
{\delta_{\begin{subarray}{l} c_{t+1}<d_{\max} \end{subarray}}\alpha^{s_{t+1},c_{t+1}+1}_t\!+\!\rho_{\sigma_{t+1}}\sum_{\tilde s_t}\hskip-0.0cm\pi_{s_{t+1}\tilde s_t}\alpha^{\tilde s_t,1}_t}.\hskip-0.08cm\nonumber
\end{align*}
which, with pre-summation over $c_{t+1}$, has cost ${\cal O}(TS(S+d_{\max}))$.

\subsubsection*{Extended Viterbi} With the definition $\xi^{\sigma_t}_t=\max_{\sigma_{1:t-1}} p(\sigma_{1:t},v_{1:t})$,
the most likely sequence $\sigma^*_{1:T}=\argmax_{\sigma_{1:T}} p(\sigma_{1:T}|v_{1:T})$ can be obtained as follows:
\begin{align*}
&\xi^{\sigma_1}_1=p(\sigma_1,v_1)=\bar\alpha^{\sigma_1}_1\\[3pt]
&\textrm{for }t=2,\ldots,T\\[1pt]
&\hskip0.55cm\xi^{\sigma_t}_t=
\begin{cases}
e^{s_t}_t\xi^{s_t,c_t+1}_{t-1} & \hskip-0.2cm  \textrm{if } c_t\!<\!d_{\min}\\
e^{s_t}_t\max [\xi^{s_t,c_t+1}_{t-1}\!,\rho_{\sigma_t}\max\limits_{s_{t-1}}\pi_{s_ts_{t-1}}\xi^{s_{t-1},1}_{t-1}] & \hskip-0.2cm \textrm{if } d_{\min}\!\leq\! c_t\!<\!d_{\max}\\
e^{s_t}_t\rho_{\sigma_t}\max\limits_{s_{t-1}}\pi_{s_ts_{t-1}}\xi^{s_{t-1},1}_{t-1} & \hskip-0.2cm \textrm{if } c_t\!=\!d_{\max}
\end{cases}\\
&\hskip0.55cm\psi^{\sigma_t}_t=\begin{cases}
(\argmax\limits_{s_{t-1}}\pi_{s_t,s_{t-1}}\xi^{s_{t-1},1}_{t-1},1) & \hskip-0.2cm\textrm{if } c_t\!=\!d_{\max}, \textrm{ or } d_{\min}\!\leq\! c_t\!<\!d_{\max}\\[-3pt]
& \hskip-0.2cm\textrm{\& } \rho_{\sigma_t}\max\limits_{s_{t-1}}\pi_{s_ts_{t-1}}\xi^{s_{t-1},1}_{t-1}\!>\!\xi^{s_t,c_t+1}_{t-1} \\
(s_t,c_t\!+\!1) & \hskip-0.2cm\textrm{otherwise}
\end{cases}\\
&\sigma^*_T=\argmax_{\sigma_T} \xi^{\sigma_T}_T\\
&\textrm{for } t=T\!-\!1,\ldots,1\\
&\hskip0.55cm \sigma^*_t=\psi^{\sigma^*_{t+1}}_{t+1}.
\end{align*}

\subsubsection*{Segment-duration distribution learning}
The part of the expectation of the complete data log-likelihood that depends on $\rho_{\sigma_t}$ is
$\sum_{t=2}^T\sum_{\sigma_t}p(c_{t-1}=1,\sigma_t|v_{1:T})\log\rho_{\sigma_t}$,
giving update
\begin{align*}
\rho_{\sigma_t}&\!=\!\frac{\sum_tp(c_{t-1}\!=\!1,\sigma_t|v_{1:T})}{\sum_{t,\tilde c_t}p(c_{t-1}\!=\!1,s_t,\tilde c_t|v_{1:T})}\\
&\!\propto\!\sum_t\frac{\rho_{\sigma_t}\gamma^{\sigma_t}_t}{\delta_{\begin{subarray}{l} c_t<d_{\max} \end{subarray}}\alpha^{s_t,c_t+1}_{t-1}
\hskip-0.0cm+\rho_{\sigma_t}\hskip-0.0cm\sum_{\tilde s_{t-1}}\hskip-0.0cm\pi_{s_t\tilde s_{t-1}}\alpha^{\tilde{s}_{t-1},1}_{t-1}}.
\end{align*}
When $\rho$ is high dimensional, the number of parameters to be estimated can be reduced by constraining $\rho_{\sigma_t}$ to be the same for count variables in a neighbourhood.

\subsubsection*{Artificial data example}
\begin{figure}[t] 
\hskip0.08cm
\includegraphics[]{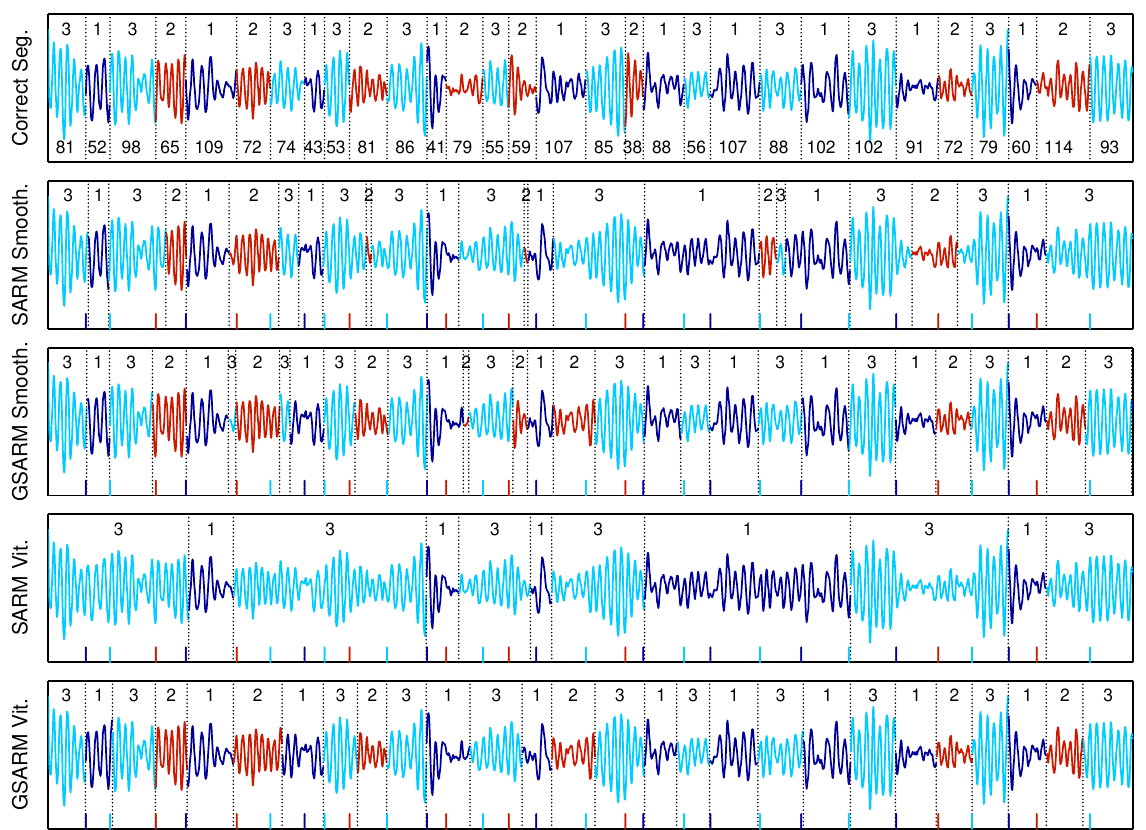}
\caption{Top: Segmentation up to the first 30 regime changes of the \timeseries~generated from the switching autoregressive process (\ref{eq:ts_sar_sardur}).
The numbers at the top and bottom indicate the regimes and the durations respectively. Bottom:
Segmentations obtained with \SARM~and \GSARM~using smoothing and extended Viterbi. The correct segmentation is indicated with bars.}
\label{fig:sar_sardur_obs}
\end{figure}

In this section, we illustrate the benefit of explicit-duration modelling on an artificial \timeseries~generated from the following switching autoregressive process:
\begin{align}
\label{eq:ts_sar_sardur}
&a^1_1=1.8, a^1_2-0.92;\hspace{0.14cm}a^2_1=1.75,a^2_2=-0.95;\hspace{0.14cm}a^3_1=1.8,a^3_2=-0.98\nonumber\\
& t\!=\!0\nonumber\\
&\text{for }k\!=\!1,\ldots,100\nonumber\\
&\hskip0.55cm\text{Sample a regime } s\!\in\!\{1,2,3\}\text{ from }\tilde{\pi}\text{ with } \tilde{\pi}_j\!=\!1/3 \text{ for } t\!=\!0,\nonumber\\
&\hskip0.55cm \text{and from }\pi \text{ with } \pi_{ii}\!=\!0 \text{ and } \pi_{ji}\!=\!1/2 \text{ for } t\!>\!0.\nonumber\\
&\hskip0.55cm\text{Sample a duration } d\!\in\!\{30,\ldots,120\} \text{ from the distribution}\nonumber\\
&\hskip0.55cm\text{obtained by discretizing and truncating a Gaussian distribution} \nonumber\\
&\hskip0.55cm \text{with mean 75 and variance 500.}\nonumber\\ 
&\hskip0.55cm \text{for } \tau\!=\!1,\ldots,d \nonumber\\
&\hskip1cm \text{Generate }v_{t+\tau}=\sum_{i=1}^2 a^s_i v_{t+\tau-i}+\eta_t\,,\hskip0.2cm \eta_t\sim{\cal N}(\eta_t;0,\sigma^2)\\[-2pt]
&\hskip0.55cm t\!=\!t\!+\!d\nonumber 
\end{align}
The \timeseries~up to the first 30 regime changes is shown at the top of \figref{fig:sar_sardur_obs}.
Notice that the underlying regimes are difficult to identify, as the autoregressive coefficients are
very similar and the transition matrix $\pi$ is uninformative; and that
it is not clear whether knowledge of the segment-duration distribution can aid the identification, as this is shared across regimes and has high variance.

We compared the segmentations obtained with a standard switching autoregressive model (\SARM) and
its explicit-duration extension employing the discretized truncated Gaussian distribution used to generate the \timeseries~(\GSARM), assuming that the autoregressive coefficients
and noise variance were known.
\SARM~used the maximum likelihood values of $\tilde\pi$ and $\pi$ estimated using the correct segmentation.

\begin{figure}[t] 
\hskip0.1cm
\subfigure[]{
\includegraphics[]{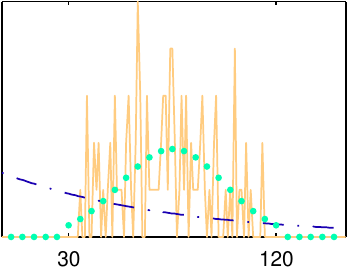}}
\hskip0.15cm
\subfigure[]{
\includegraphics[]{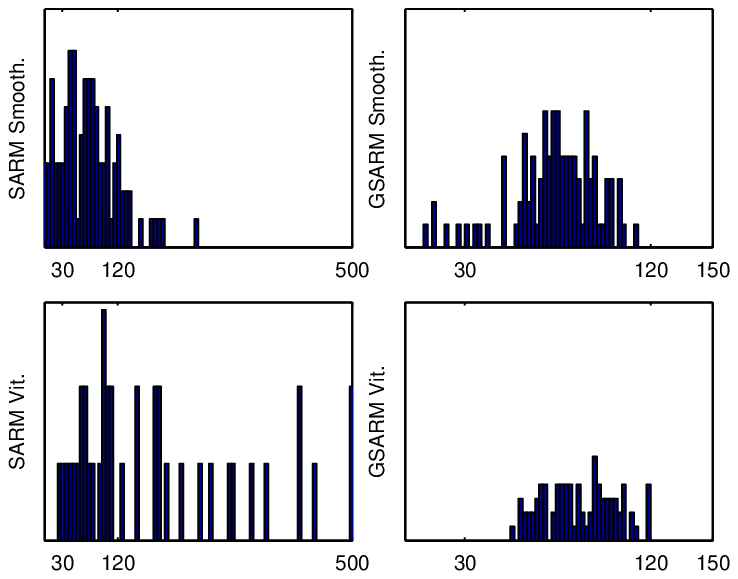}
}
\caption{(a): Empirical segment-duration distribution (continuous line), geometric segment-duration distribution
implicitly defined in \SARM~(dashed line), and segment-duration distribution used in \GSARM~(dotted line).
(b): Empirical segment-duration distribution of the estimated segmentation for \SARM~(left) and \GSARM~(right) using smoothing (top) and extended Viterbi (bottom).}
\label{fig:sar_sardur_dist}
\end{figure}

In \figref{fig:sar_sardur_dist}(a) we plot the empirical segment-duration distribution (continuous line), the geometric segment-duration distribution
implicitly defined in \SARM~(dashed line), and the segment-duration distribution used in \GSARM~(dotted line).

\noindent The segmentations, obtained by estimating $s^*_t=\argmax_{s_t}\sum_{c_t}\gamma^{\sigma_t}_t$ (smoothing)
and $\sigma^*_{1:T}=\argmax_{\sigma_{1:T}} p(\sigma_{1:T}|v_{1:T})$ (extended Viterbi),
are displayed at the bottom of \figref{fig:sar_sardur_obs}.
As a measure of segmentation error, we used the discrepancy between the correct and the estimated regimes.
\SARM~gave $30\%$ error with smoothing and $43\%$ error with extended Viterbi,
whilst \GSARM~gave $18\%$ error with smoothing and $25\%$ with extended Viterbi.

In \figref{fig:sar_sardur_dist}(b) we plot the empirical segment-duration distributions estimated from the segmentations.

\subsection{Increasing Count Variables \label{sec:M1altInf}}
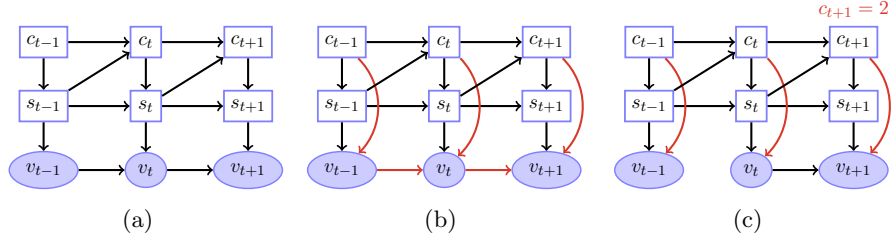
\begin{figure}[t]
\begin{center}
\hskip-0.2cm
\subfigure[]{\scalebox{0.68}{
\begin{tikzpicture}
\node[disc] (sigmatm) at (2,0) {$c_{t-1}$};
\node[disc] (sigmat) at (4,0) {$c_t$};
\node[disc] (sigmatp) at (6,0) {$c_{t+1}$};
\node[disc] (stm) at (2,-1.25) {$s_{t-1}$};
\node[disc] (st) at (4,-1.25) {$s_t$};
\node[disc] (stp) at (6,-1.25) {$s_{t+1}$};
\node[ocont,obs] (vtm) at (2,-2.5) {$v_{t-1}$};
\node[ocont,obs] (vt) at (4,-2.5) {$v_t$};
\node[ocont,obs] (vtp) at (6,-2.5) {$v_{t+1}$};
\draw[->, line width=1.15pt](stm)--(st);\draw[->, line width=1.15pt](st)--(stp);
\draw[->, line width=1.15pt](stm)--(vtm);\draw[->, line width=1.15pt](st)--(vt);\draw[->, line width=1.15pt](stp)--(vtp);
\draw[->, line width=1.15pt](sigmatm)--(stm);\draw[->, line width=1.15pt](sigmat)--(st);\draw[->, line width=1.15pt](sigmatp)--(stp);
\draw[->, line width=1.15pt](sigmatm)--(sigmat);\draw[->, line width=1.15pt](sigmat)--(sigmatp);
\draw[->,line width=1.15pt](vtm)--(vt);\draw[->, line width=1.15pt](vt)--(vtp);
\draw[->, line width=1.15pt](stm)--(sigmat);\draw[->, line width=1.15pt](st)--(sigmatp);
\end{tikzpicture}}}
\hskip0.1cm
\subfigure[]{\scalebox{0.68}{
\begin{tikzpicture}
\node[disc] (sigmatm) at (2,0) {$c_{t-1}$};
\node[disc] (sigmat) at (4,0) {$c_t$};
\node[disc] (sigmatp) at (6,0) {$c_{t+1}$};
\node[disc] (stm) at (2,-1.25) {$s_{t-1}$};
\node[disc] (st) at (4,-1.25) {$s_t$};
\node[disc] (stp) at (6,-1.25) {$s_{t+1}$};
\node[ocont,obs] (vtm) at (2,-2.5) {$v_{t-1}$};
\node[ocont,obs] (vt) at (4,-2.5) {$v_t$};
\node[ocont,obs] (vtp) at (6,-2.5) {$v_{t+1}$};
\draw[->, line width=1.15pt](stm)--(st);\draw[->, line width=1.15pt](st)--(stp);
\draw[->, line width=1.15pt](stm)--(vtm);\draw[->, line width=1.15pt](st)--(vt);\draw[->, line width=1.15pt](stp)--(vtp);
\draw[->, line width=1.15pt](sigmatm)--(stm);\draw[->, line width=1.15pt](sigmat)--(st);\draw[->, line width=1.15pt](sigmatp)--(stp);
\draw[->, line width=1.15pt](sigmatm)--(sigmat);\draw[->, line width=1.15pt](sigmat)--(sigmatp);
\draw[->, color=myred,line width=1.15pt](sigmatm)to [bend left=45](vtm);\draw[->, color=myred,line width=1.15pt](sigmat)to [bend left=45](vt);\draw[->, color=myred,line width=1.15pt](sigmatp)to [bend left=45](vtp);
\draw[->,color=myred,line width=1.15pt](vtm)--(vt);\draw[->,color=myred,line width=1.15pt](vt)--(vtp);
\draw[->, line width=1.15pt](stm)--(sigmat);\draw[->, line width=1.15pt](st)--(sigmatp);
\end{tikzpicture}}}
\hskip0.1cm
\subfigure[]{\scalebox{0.68}{
\begin{tikzpicture}
\node[disc] (sigmatm) at (2,0) {$c_{t-1}$};
\node[disc] (sigmat) at (4,0) {$c_t$};
\node[disc] (sigmatp) at (6,0) {$c_{t+1}$};
\node[disc] (stm) at (2,-1.25) {$s_{t-1}$};
\node[disc] (st) at (4,-1.25) {$s_t$};
\node[disc] (stp) at (6,-1.25) {$s_{t+1}$};
\node[ocont,obs] (vtm) at (2,-2.5) {$v_{t-1}$};
\node[ocont,obs] (vt) at (4,-2.5) {$v_t$};
\node[ocont,obs] (vtp) at (6,-2.5) {$v_{t+1}$};
\node [color=myred,above] at (sigmatp.north) {$c_{t+1}=2$};
\draw[->, line width=1.15pt](stm)--(st);\draw[->, line width=1.15pt](st)--(stp);
\draw[->, line width=1.15pt](stm)--(vtm);\draw[->, line width=1.15pt](st)--(vt);\draw[->, line width=1.15pt](stp)--(vtp);
\draw[->, line width=1.15pt](sigmatm)--(stm);\draw[->, line width=1.15pt](sigmat)--(st);\draw[->, line width=1.15pt](sigmatp)--(stp);
\draw[->, line width=1.15pt](sigmatm)--(sigmat);\draw[->, line width=1.15pt](sigmat)--(sigmatp);
\draw[->, color=myred,line width=1.15pt](sigmatm)to [bend left=45](vtm);\draw[->, color=myred,line width=1.15pt](sigmat)to [bend left=45](vt);\draw[->, color=myred,line width=1.15pt](sigmatp)to [bend left=45](vtp);
\draw[->,line width=1.15pt](vt)--(vtp);
\draw[->, line width=1.15pt](stm)--(sigmat);\draw[->, line width=1.15pt](st)--(sigmatp);
\end{tikzpicture}}}
\end{center}
\caption{(a): \HMSM~in which the segment-duration distribution is explicitly modelled using increasing count variables $c_{1:T}$.
(b): Across-segment independence is enforced with a link from $c_t$ to $v_t$. (c): Explicit representation of across-segment independence.
The values $c_{t+1}=2$ indicates that the segment passing through time-step $t+1$ starts at time-step $t+1-c_{t+1}+1=t$.}
\label{fig:M1alt}
\end{figure}

Like decreasing count variables, increasing count variables allow Markovian dependence among the observations.
For Markovian order $k=1$, this type of models is represented by the belief network shown in \figref{fig:M1alt}(a).
Across-segment independence can be enforced by adding a link from $c_t$ to $v_t$ (as shown in \figref{fig:M1alt}(b)
and explicitly represented in \figref{fig:M1alt}(c)), such that
$p(v_t|\sigma_t,v_{t-k:t-1})=p(v_t|\sigma_t,v_{t-\min(c_t,k)+1:t-1})$.

In the following sections we describe inference assuming across-segment independence.

\subsubsection*{Parallel filtering-smoothing}
The filtered distribution $\alpha^{\sigma_t}_t=p(\sigma_t|v_{1:t})$ can be obtained by normalizing $\bar\alpha^{\sigma_t}_t=p(\sigma_t,v_{1:t})$, where $\bar\alpha^{\sigma_t}_t$ can be computed as\footnote{The initialization is given by $\bar\alpha^{\sigma_1}_1=p(v_1|s_1)\tilde\pi_{s_1}\tilde\lambda_{\sigma_1}$.}
\begin{align}
\hskip-0.2cm\bar\alpha^{\sigma_t}_t&=p(v_t|\sigma_t,\cancel{v_{1:t-k-1}},v_{t-k:t-1})\sum_{\sigma_{t-1}}p(\sigma_t|\sigma_{t-1},\cancel{v_{1:t-1}})p(\sigma_{t-1},v_{1:t-1})\nonumber\\[3pt]
&=e^{\sigma_t}_t\bigg\{\delta_{\begin{subarray}{l} c_t>1 \\ s_{t-1}=s_t \\ c_{t-1}=c_t-1 \end{subarray}}
\hskip-0.3cm\lambda_{\sigma_{t-1}}\hskip-0.1cm
+\delta_{c_t=1}\sum_{s_{t-1}}\pi_{s_ts_{t-1}}\hskip-0.0cm\sum_{c_{t-1}}\hskip-0.0cm(1\!-\!\lambda_{\sigma_{t-1}})\bigg\}\bar\alpha^{\sigma_{t-1}}_{t-1},\hskip-0.1cm
\label{eq:alphaM1alt}
\end{align}
with $e^{\sigma_t}_t=p(v_t|\sigma_t,v_{t-\min(c_t,k)+1:t-1})$. With pre-computation of $\sum_{c_{t-1}}(1-\lambda_{\sigma_{t-1}})\bar\alpha^{\sigma_{t-1}}_{t-1}$, which does not depend on $s_t$, this recursion has cost
${\cal O}(TS(S+Ed_{\max}))$, where $E$ is the cost of computing $e^{\sigma_t}_t$.

Notice that $\bar\alpha^{\sigma_t}_t=0$ implies $\bar\alpha^{s_t,c_t+1}_{t+1}=\ldots=\bar\alpha^{s_t,d_{\max}}_{t+d_{\max}-c_t}=0$,
\ie~if according to $v_{1:t}$ a segment starting at time-step $t-c_t+1$ and generated by $s_t$ cannot have duration $\geq c_t$, that segment cannot have duration $\geq c_t+1$ after incorporating observations $v_{t+1}$, etc.
This result can be used to design approximation schemes for reducing the computational cost by pruning some $\bar\alpha^{\sigma_t}_t$, see \secref{sec:Approx}.

The smoothed distribution $\gamma^{\sigma_t}_t=p(\sigma_t|v_{1:T})$ can be obtained as $\gamma^{\sigma_t}_t\propto p(v_{t+1:T}|\sigma_t,\cancel{v_{1:t-k}},v_{t-k+1:t})p(\sigma_t,v_{1:t})=\beta^{\sigma_t}_t\bar\alpha^{\sigma_t}_t$, where $\beta^{\sigma_t}_t=p(v_{t+1:T}|\sigma_t,v_{t-k+1:t})$ can be computed as\footnote{The initialization is given by $\beta^{\sigma_T}_T=1$.}
\begin{align*}
&\beta^{\sigma_t}_t\!=\!\!\sum_{\sigma_{t+1}}\!p(v_{t+1:T}|\cancel{\sigma_t},\sigma_{t+1},v_{t-k+1:t})p(\sigma_{t+1}|\sigma_t,\cancel{v_{t-k+1:t}})\\
&\!=\!\!\sum_{\sigma_{t+1}}\!p(v_{t+2:T}|\sigma_{t+1},\cancel{v_{t-k+1}},v_{t-k+2:t+1})p(v_{t+1}|\sigma_{t+1},v_{t-k+1:t})p(\sigma_{t+1}|\sigma_t)\\
&\!=\!\bigg\{\delta_{\begin{subarray}{l} c_t<d_{\max} \\ s_{t+1}=s_t \\ c_{t+1}=c_t+1 \end{subarray}}
\lambda_{\sigma_t}
+\delta_{\begin{subarray}{l} c_t\geq d_{\min} \\[2pt] c_{t+1}=1 \end{subarray}}
(1\!-\!\lambda_{\sigma_t})\sum_{s_{t+1}}\pi_{s_{t+1}s_t}
\bigg\}e^{\sigma_{t+1}}_{t+1}\beta^{\sigma_{t+1}}_{t+1}.
\end{align*}
With pre-computation of $\sum_{s_{t+1}}\pi_{s_{t+1}s_t}e^{s_{t+1},1}_{t+1}\beta^{s_{t+1},1}_{t+1}$, which does not depend on $c_t$, this recursion has cost ${\cal O}(TS(S+d_{\max}))$.

\subsubsection*{Sequential filtering-smoothing}
The filtered distribution $\alpha^{\sigma_t}_t=p(\sigma_t|v_{1:t})$ can be obtained as $\alpha^{\sigma_t}_t=\frac{p(\sigma_t,v_t|v_{1:t-1})}{p(v_t|v_{1:t-1})}$, where the numerator can be computed as in recursion (\ref{eq:alphaM1alt}).

The smoothed distribution $\gamma^{\sigma_t}_t=p(\sigma_t|v_{1:T})$ can be computed as
\begin{align}
\label{eq:gammaM1alt}
&\gamma^{\sigma_t}_t\!=\!\sum_{\sigma_{t+1}}p(\sigma_t|\sigma_{t+1},v_{1:t},\cancel{v_{t+1:T}})p(\sigma_{t+1}|v_{1:T})\\
&\!=\!\delta_{c_t<d_{\max}}\gamma^{s_t,c_t+1}_{t+1}\!\!+\delta_{\begin{subarray}{l} c_t\geq d_{\min} \\[2pt] c_{t+1}=1 \end{subarray}}\sum_{s_{t+1}}\frac{p(\sigma_{t+1}|\sigma_t,\cancel{v_{1:t}})p(\sigma_t|v_{1:t})}
{\sum_{\tilde\sigma_t}p(\sigma_{t+1}|\tilde\sigma_t,\cancel{v_{1:t}})p(\tilde\sigma_t|v_{1:t})}\gamma^{\sigma_{t+1}}_{t+1}\nonumber\\
&\!=\!\delta_{c_t<d_{\max}}\gamma^{s_t,c_t+1}_{t+1}\!\!+\delta_{\begin{subarray}{l} c_t\geq d_{\min} \\[2pt] c_{t+1}=1 \end{subarray}}(1\!-\!\lambda_{\sigma_t})\alpha^{\sigma_t}_t\!\sum_{s_{t+1}}\frac{\pi_{s_{t+1}s_t}\gamma^{\sigma_{t+1}}_{t+1}}
{\sum_{\tilde s_t}\!\pi_{s_{t+1}\tilde s_t}\sum_{\tilde c_t}(1\!-\!\lambda_{\tilde\sigma_t})\alpha^{\tilde\sigma_t}_t},\nonumber
\end{align}
where we have used $p(\sigma_t|\sigma_{t+1}=(s_t,c_t+1),v_{1:t})=1$.
With pre-summation over $s_{t+1}$, this recursion has cost ${\cal O}(TS(S+d_{\max}))$.

Notice that $\alpha^{\sigma_t}_t=0$ implies $\gamma^{\sigma_t}_t=\gamma^{s_t,c_t+1}_{t+1}=\ldots=\gamma^{s_t,d_{\max}}_{t+d_{\max}-c_t}=0$.

\subsubsection*{Extended Viterbi} With the definition $\xi^{\sigma_t}_t=\max_{\sigma_{1:t-1}} p(\sigma_{1:t},v_{1:t})$,
the most likely sequence $\sigma^*_{1:T}=\argmax_{\sigma_{1:T}} p(\sigma_{1:T}|v_{1:T})$ can be obtained as follows:
\begin{align*}
&\xi^{\sigma_1}_1=p(\sigma_1,v_1)=\bar\alpha^{\sigma_1}_1\\[3pt]
&\textrm{for }t=2,\ldots,T\\
&\hskip0.55cm\textrm{for }c_t=1,\ldots,d_{\max}\\[1pt]
&\hskip1.1cm\xi^{\sigma_t}_t=
\begin{cases}
e^{\sigma_t}_t\max\limits_{s_{t-1}} \pi_{s_ts_{t-1}}\max\limits_{c_{t-1}}(1\!-\!\lambda_{\sigma_{t-1}})\xi^{\sigma_{t-1}}_{t-1} & \hskip0.55cm\textrm{if } c_t\!=\!1\\
e^{\sigma_t}_t\lambda_{s_{t-1}c_t-1}\xi^{s_t,c_t-1}_{t-1}  & \hskip0.55cm\textrm{if } c_t\!>\!1
\end{cases}\\
&\hskip1.1cm\psi^{\sigma_t}_t=
\begin{cases}
\argmax\limits_{s_{t-1}}\pi_{s_ts_{t-1}}\argmax\limits_{c_{t-1}}(1\!-\!\lambda_{\sigma_{t-1}})\xi^{\sigma_{t-1}}_{t-1} & \hskip-0.2cm\textrm{if } c_t\!=\!1\\
(s_t,c_t\!-\!1)&\hskip-0.2cm\textrm{if } c_t\!>\!1
\end{cases}\\
&\sigma^*_T=\argmax_{\sigma_T} \xi^{\sigma_T}_T\\
&\textrm{for } t=T\!-\!1,\ldots,1\\
&\hskip0.55cm\sigma^*_t=\psi^{\sigma^*_{t+1}}_{t+1}.
\end{align*}

\subsection{Count-Duration Variables \label{sec:M2Inf}}
\begin{figure}[t]
\hskip0.2cm
\subfigure[]{
\scalebox{0.75}{
\begin{tikzpicture}
\node[] at (1,1.25) {$\cdots$};
\node[disc] (sigmatm) at (2,1.25) {$c_{t-1}$};
\node[disc] (sigmat) at (4,1.25) {$c_t$};
\node[disc] (sigmatp) at (6,1.25) {$c_{t+1}$};
\node[disc] (dtm) at (2,0) {$d_{t-1}$};
\node[disc] (dt) at (4,0) {$d_t$};
\node[disc] (dtp) at (6,0) {$d_{t+1}$};
\node[disc] (stm) at (2,-1.25) {$s_{t-1}$};
\node[disc] (st) at (4,-1.25) {$s_t$};
\node[disc] (stp) at (6,-1.25) {$s_{t+1}$};
\node[] at (7,1.25) {$\cdots$};
\node[ocont,obs] (vtm) at (2,-2.5) {$v_{t-1}$};
\node[ocont,obs] (vt) at (4,-2.5) {$v_t$};
\node[ocont,obs] (vtp) at (6,-2.5) {$v_{t+1}$};
\draw[->, line width=1.15pt](stm)--(st);\draw[->, line width=1.15pt](st)--(stp);
\draw[->, line width=1.15pt](stm)--(vtm);\draw[->, line width=1.15pt](st)--(vt);\draw[->, line width=1.15pt](stp)--(vtp);
\draw[->, line width=1.15pt](dtm)--(dt);\draw[->, line width=1.15pt](dt)--(dtp);
\draw[->, line width=1.15pt](sigmatm)--(dt);\draw[->, line width=1.15pt](sigmat)--(dtp);
\draw[->, line width=1.15pt](sigmatm)--(st);\draw[->, line width=1.15pt](sigmat)--(stp);
\draw[->, line width=1.15pt](dtm)--(sigmatm);\draw[->, line width=1.15pt](dt)--(sigmat);\draw[->, line width=1.15pt](dtp)--(sigmatp);
\draw[->,color=myred,line width=1.15pt](sigmatm)to [bend left=35](vtm);\draw[->,color=myred,line width=1.15pt](sigmat)to [bend left=35](vt);\draw[->,color=myred,line width=1.15pt](sigmatp)to [bend left=35](vtp);
\draw[->,color=myred,line width=1.15pt](dtm)to [bend left=35](vtm);\draw[->,color=myred,line width=1.15pt](dt)to [bend left=35](vt);\draw[->,color=myred,line width=1.15pt](dtp)to [bend left=35](vtp);
\draw[->, line width=1.15pt](sigmatm)--(sigmat);\draw[->, line width=1.15pt](sigmat)--(sigmatp);
\draw[color=myred,line width=1.15pt](vtm)--(vt);\draw[color=myred,line width=1.15pt](vtm)to [bend left=30](vtp);\draw[color=myred,line width=1.15pt](vt)--(vtp);
\draw[->, line width=1.15pt](stm)--(dtm);\draw[->, line width=1.15pt](st)--(dt);\draw[->, line width=1.15pt](stp)--(dtp);
\end{tikzpicture}}}
\hskip0.3cm
\subfigure[]{
\scalebox{0.75}{
\begin{tikzpicture}
\node[] at (1,1.25) {$\cdots$};
\node[disc] (sigmatm) at (2,1.25) {$c_{t-1}$};
\node[disc] (sigmat) at (4,1.25) {$c_t$};
\node[disc] (sigmatp) at (6,1.25) {$c_{t+1}$};
\node[disc] (dtm) at (2,0) {$d_{t-1}$};
\node[disc] (dt) at (4,0) {$d_t$};
\node[disc] (dtp) at (6,0) {$d_{t+1}$};
\node[disc] (stm) at (2,-1.25) {$s_{t-1}$};
\node[disc] (st) at (4,-1.25) {$s_t$};
\node[disc] (stp) at (6,-1.25) {$s_{t+1}$};
\node[] at (7,1.25) {$\cdots$};
\node[ocont,obs] (vtm) at (2,-2.5) {$v_{t-1}$};
\node[ocont,obs] (vt) at (4,-2.5) {$v_t$};
\node[ocont,obs] (vtp) at (6,-2.5) {$v_{t+1}$};
\node [color=myred,above] at (sigmatp.north) {$d_{t+1}=3, c_{t+1}=2$};
\draw[->, line width=1.15pt](stm)--(st);\draw[->, line width=1.15pt](st)--(stp);
\draw[->, line width=1.15pt](stm)--(vtm);\draw[->, line width=1.15pt](st)--(vt);\draw[->, line width=1.15pt](stp)--(vtp);
\draw[->, line width=1.15pt](dtm)--(dt);\draw[->, line width=1.15pt](dt)--(dtp);
\draw[->, line width=1.15pt](sigmatm)--(dt);\draw[->, line width=1.15pt](sigmat)--(dtp);
\draw[->, line width=1.15pt](sigmatm)--(st);\draw[->, line width=1.15pt](sigmat)--(stp);
\draw[->, line width=1.15pt](dtm)--(sigmatm);\draw[->, line width=1.15pt](dt)--(sigmat);\draw[->, line width=1.15pt](dtp)--(sigmatp);
\draw[->,color=myred,line width=1.15pt](sigmatm)to [bend left=35](vtm);\draw[->,color=myred,line width=1.15pt](sigmat)to [bend left=35](vt);\draw[->,color=myred,line width=1.15pt](sigmatp)to [bend left=35](vtp);
\draw[->,color=myred,line width=1.15pt](dtm)to [bend left=35](vtm);\draw[->,color=myred,line width=1.15pt](dt)to [bend left=35](vt);\draw[->,color=myred,line width=1.15pt](dtp)to [bend left=35](vtp);
\draw[->, line width=1.15pt](sigmatm)--(sigmat);\draw[->, line width=1.15pt](sigmat)--(sigmatp);
\draw[line width=1.15pt](vt)--(vtp);
\draw[->, line width=1.15pt](stm)--(dtm);\draw[->, line width=1.15pt](st)--(dt);\draw[->, line width=1.15pt](stp)--(dtp);
\end{tikzpicture}}}
\caption{(a): \HMSM~in which the segment-duration distribution is explicitly modelled using decreasing count variables $c_{1:T}$ and duration variables $d_{1:T}$.
The undirected links between the observations indicate non-Markovian dependence. Across-segment independence is enforced with a link from $c_t$ and $d_t$ to $v_t$.
(b): Explicit representation of across-segment independence. The values $d_{t+1}=3, c_{t+1}=2$ indicate that the segment passing through time-step $t+1$ starts at time-step $t+1-d_{t+1}+c_{t+1}=t$.}
\label{fig:M2}
\end{figure}
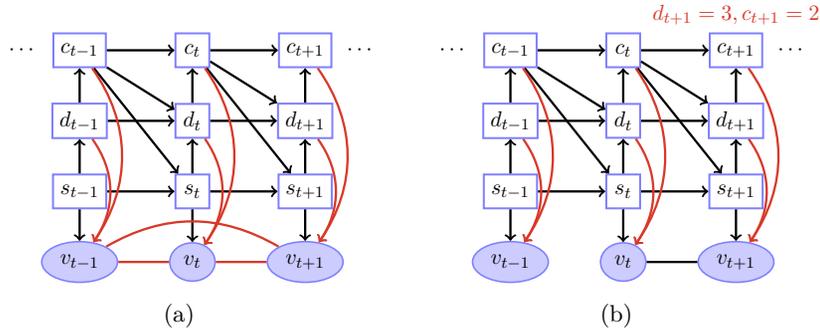

Count-duration variables allow any structure for $p(v_{1:T}|\sigma_{1:T})$ within a segment and therefore, unlike count variables,
also a distribution $p(v_{1:T}|\sigma_{1:T})$ that cannot be efficiently computed as $\prod_{t} p(v_t|\sigma_t,v_{1:t-1})$.
For models that contain only $\sigma_{1:T}$ and $v_{1:T}$ and with across-segment independence, this translates into non-Markovian dependence between the observations.
This type of models is represented by the belief network shown in \figref{fig:M2}(a),
where across-segment independence is enforced with a link from $c_{t}$ and $d_t$ to $v_t$ (explicitly represented in \figref{fig:M2}(b))
and non-Markovian dependence is indicated by undirected links.

Non-Markovian dependence between the observations within a segment is possible as, whilst time-recursive inference cannot be performed in this complex scenario,
knowledge about segment beginning and segment end enables segment-recursive inference,
namely in terms of count variables that take value 1 and involving the whole segment-emission distribution
$e^{s_t,d_t}_t=p(v_{t-d_t+1:t}|s_t,d_t,c_t=1)
=p(v_{t-d_t+1:t}|\sigma_{t-d_t+1}=(s_t,d_t,d_t),\ldots,\sigma_{t-1}=(s_t,d_t,2),s_t,d_t,c_t=1)$.

The segmental recursions that we describe coincide with the standard recursions of hidden semi-Markov/segment models \citep{ferguson80variable,rabiner89tutorial,ostendorf96from,murphy2002hsm,yu10hidden},
which are obtained by defining only duration variables and by performing the computations at the occurrence of the events \emph{segment end} and \emph{segment beginning}.
As discussed above and in \cite{murphy2002hsm}, this approach can be more easily explained by defining also variables $c_{1:T}$ such that, \eg, $c_t=1$ at the end of the segment
and $c_t=2$ otherwise, and by performing inference in terms of time-steps for which $c_{1:T}$ take value 1.
These variables can be seen as collapsed count variables that encode information about \emph{whether} (rather than \emph{where}) the segment is ending,
such that information about segment beginning and segment end is available only at the end of the segment.
In this encoding $\sigma_{1:T}$ do not form a first-order Markov chain. Indeed, \eg, $c_t$ depends on $d_{t},c_{t-1}$ if $c_{t-1}=1$,
whilst it depends on $c_{t-d_t+1:t-1}$ if $c_{t-1}=2$.

Encoding information about segment beginning and segment end anywhere within the segment, whilst not having any computational disadvantage, has the advantage
of making the derivation of posterior distributions more immediate. This is particularly useful in models with additional unobserved variables related by first-order Markovian dependence,
as we will see in \secref{sec:SLGSSM}.

In the following sections we describe segmental inference and learning assuming across-segment independence.

\subsubsection*{Segmental parallel filtering-smoothing}
Using the notation $\sigma^{1}_t=(s_t,d_t,c_t=1)$, the filtered distribution $\alpha^{\sigma^{1}_t}_t=p(\sigma^{1}_t|v_{1:t})$ can be obtained by normalizing $\bar\alpha^{\sigma^{1}_t}_t=p(\sigma^{1}_t,v_{1:t})$,
where $\bar\alpha^{\sigma^{1}_t}_t$ can be computed as\footnote{For $t=1,\ldots,d_{\max}$, $\bar\alpha^{\sigma^1_t}_t=p(v_{1:t}|\sigma^1_t)\tilde{\pi}_{s_t}\tilde{\rho}_{s_td_t}\tilde{\tilde{\rho}}_{d_tt}$ if $d_t\geq t$.
}
\begin{align}
\bar\alpha^{\sigma^{1}_t}_t&=\sum_{\sigma_{t-d_t:t-1}}\hskip-0.2cm p(v_{t-d_t+1:t}|\cancel{\sigma_{t-d_t}},\sigma_{t-d_t+1:t-1},\sigma^1_t,\cancel{v_{1:t-d_t}})\nonumber\\
&\hskip0.6cm\times p(\sigma_{t-d_t+1:t-1},\sigma^{1}_t|\sigma_{t-d_t},\cancel{v_{1:t-d_t}})p(\sigma_{t-d_t},v_{1:t-d_t})\nonumber\\[5pt]
&=e^{s_t,d_t}_t\hspace{-0.4cm}\sum_{s_{t-d_t},d_{t-d_t}}\hskip-0.2cm p(s_{t-d_t+1}\!=\!s_t|s_{t-d_t},c_{t-d_t}\!=\!1)\nonumber\\
&\hskip0.6cm\times p(d_{t-d_t+1}\!=\!d_t|d_{t-d_t},c_{t-d_t}\!=\!1)p(\sigma^1_{t-d_t},v_{1:t-d_t})\nonumber\\[5pt]
&=e^{s_t,d_t}_t\rho_{s_td_t}\sum_{s_{t-d_t}}\pi_{s_ts_{t-d_t}}\sum_{d_{t-d_t}}\bar\alpha^{\sigma^{1}_{t-d_t}}_{t-d_t}.
\label{eq:alphacountdur}
\end{align}
Naive computation of this recursion has cost ${\cal O}(TS^2Ed_{\max}^2)$, where $E$ is the cost of computing $e^{s_t,d_t}_t$. However, with pre-computation of
$\sum_{d_{t-d_t}}\bar\alpha^{\sigma^{1}_{t-d_t}}_{t-d_t}$, which
does not depend on $s_t$ and $d_t$, and with pre-computation of $\sum_{s_{t-d_t}}\pi_{s_ts_{t-d_t}}\sum_{d_{t-d_t}}\bar\alpha^{\sigma^{1}_{t-d_t}}_{t-d_t}$, which
does not depend on $d_t$, the cost reduces to ${\cal O}(TS(S+Ed_{\max}))$\footnote{In the case of Markovian dependence between the observations, $E$ is the cost of computing $p(v_t|\sigma^{1}_t,v_{t-d_t+1:t-1})$,
as $e^{s_t,d_t}_t$ can be computed recursively, \ie~$e^{s_t,d_t}_t=p(v_t|\sigma^{1}_t,v_{t-d_t+1:t-1})e^{s_t,d_t-1}_{t-1}$.}.

In the case of Markovian dependence between the observations, if $\bar\alpha^{\sigma_t}_t$ for $c_t>1$ is of interest,
a time-recursive routine on the line of the one described in Appendix~\ref{app:SS} for $\alpha^{\sigma_t}_t$ can be used.

\noindent The smoothed distribution $\gamma^{\sigma^{1}_t}_t=p(\sigma^{1}_t|v_{1:T})$ can be obtained
as\footnote{The normalization term $p(v_{1:T})$ can be computed by summing the rhs of \eqref{eq:alphacountdur} over $s_t$ for a time-step $t$, or as $\sum_{s_T,d_T}\bar\alpha^{\sigma^{1}_T}_T$ if the constraint $c_T=1$ is imposed or if inference is conditioned on this event.}
$\gamma^{\sigma^{1}_t}_t\propto p(v_{t+1:T}|s_t,\cancel{d_t},c_t\!=\!1,\cancel{v_{1:t}})p(\sigma^{1}_t,v_{1:t})=\beta^{s_t,1}_t\bar\alpha^{\sigma^{1}_t}_t$,
where, with the notation $\sigma^{j,k,1}_{t+k}=(s_{t+k}=j,d_{t+k}=k,c_{t+k}=1)$, $\beta^{s_t,1}_t=p(v_{t+1:T}|s_t,c_t=1)$ can be computed as\footnote{For $t\geq T$, $\beta^{s_t,1}_t=1$.
Setting $\beta^{s_t,1}_t=0$ for $t>T$ corresponds to conditioning inference on the event $c_T=1$.}
\begin{align}
\beta^{s_t,1}_t&=\!\sum_{j,k}\!p(v_{t+1:T}|\sigma^{j,k,1}_{t+k},\cancel{s_t,c_t\!=\!1})p(\sigma^{j,k,1}_{t+k}|s_t,c_t\!=\!1)\nonumber\\[2pt]
&=\!\sum_{j,k}\!p(v_{t+1:t+k}|\sigma^{j,k,1}_{t+k},\cancel{v_{t+k+1:T}})p(v_{t+k+1:T}|\sigma^{j,k,1}_{t+k})\pi_{js_t}\rho_{jk}\nonumber\\[5pt]
&=\!\sum_{j}\!\pi_{js_t}\sum_{k}p(v_{t+1:t+k}|\sigma^{j,k,1}_{t+k})\beta^{j,1}_{t+k}\rho_{jk}.
\label{eq:betacountdur}
\end{align}
With pre-computation of $\sum_{k}p(v_{t+1:t+k}|\sigma^{j,k,1}_{t+k})\beta^{j,1}_{t+k}\rho_{jk}$, this recursion has cost ${\cal O}(TS(S+d_{\max}))$.

Notice that recursions (\ref{eq:alphacountdur}) and (\ref{eq:betacountdur}) correspond to the standard recursions of hidden semi-Markov/segment
models using collapsed count variables \citep{ferguson80variable,rabiner89tutorial,ostendorf96from,murphy2002hsm,yu10hidden}. 

The smoothed distribution $\gamma^{\sigma_t}_t$ for $c_t>1$ can be obtained as $\gamma^{\sigma_t}_t=\gamma^{s_t,d_t,1}_{t+c_t-1}$. Indeed,
in such a case,
\begin{align}
\gamma^{\sigma_t}_t&=\sum_{\sigma_{t+1}}p(\sigma_t|\sigma_{t+1},v_{1:t},\cancel{v_{t+1:T}})p(\sigma_{t+1}|v_{1:T})\nonumber\\
&=\gamma^{s_t,d_t,c_t-1}_{t+1}=\gamma^{s_t,d_t,c_t-2}_{t+2}=\cdots=\gamma^{s_t,d_t,1}_{t+c_t-1},
\label{eq:gammacountdurg1}
\end{align}
where we have used $p(\sigma_t|\sigma_{t+1}=(s_t,d_t,c_t-1),v_{1:t})=1$.

From \eqref{eq:gammacountdurg1} we can immediately derive $p(s_t,c_t|v_{1:T})$ and $p(s_t|v_{1:T})$ as
\begin{align*}
&p(s_t,c_t|v_{1:T})=\sum_{d_t=\max(d_{\min},c_t)}^{d_{\max}}\gamma^{\sigma_t}_t=\sum_{d_t}\gamma^{s_t,d_t,1}_{t+c_t-1}
\propto\beta^{s_t,1}_{t+c_t-1}\sum_{d_t}\bar\alpha^{s_t,d_t,1}_{t+c_t-1},
\end{align*}
and
\begin{align}
\hskip-0.2cm p(s_t|v_{1:T})=\sum_{c_t=1}^{d_{\max}}p(s_t,c_t|v_{1:T})\propto\sum_{\tau=t}^{t+d_{\max}-1}\beta^{s_t,1}_{\tau}\hskip-0.3cm\sum_{d_t=\max(d_{\min},\tau-t+1)}^{d_{\max}}\hskip-0.7cm\bar\alpha^{s_t,d_t,1}_{\tau}.\hskip-0.1cm
\label{eq:gpost}
\end{align}
In the standard approach that uses collapsed count variables, $p(s_t|v_{1:T})$ is derived by observing that
the set of all segments passing through time-step $t$ needs to be considered, and that this set can be obtained by subtracting all segments ending before time-step $t$
from all segments starting at time-step $t$ or before (see Appendix~\ref{app:SS}).
In \eqref{eq:gpost}, $p(s_t|v_{1:T})$ is computed by summing over all segments passing through time-step $t$,
which are obtained as all segments that start at time-step $t$ or before and end at time-step $t$ or after.
However, the equation was derived by use of equivalence (\ref{eq:gammacountdurg1}) rather than by use of this observation.
Therefore, uncollapsed count variables enable more automatic derivations of posterior distributions of interest.

\subsubsection*{Segmental sequential filtering-smoothing}
The filtered distribution $\alpha^{\sigma^{1}_t}_t=p(\sigma^{1}_t|v_{1:t})$ can be obtained as  $\alpha^{\sigma^{1}_t}_t=\frac{p(\sigma^{1}_t,v_{t-d_t+1:t}|v_{1:t-d_t})}{p(v_{t-d_t+1:t}|v_{1:t-d_t})}$, where
the numerator can be computed as in recursion (\ref{eq:alphacountdur}).

The smoothed distribution $\gamma^{\sigma^{1}_t}_t=p(\sigma^{1}_t|v_{1:T})$ can be computed as
\begin{align}
\gamma^{\sigma^{1}_t}_t
&=\sum_{s_{t+1},d_{t+1}}\frac{\pi_{s_{t+1}s_t}\cancel{\rho_{s_{t+1}d_{t+1}}}\alpha^{\sigma^{1}_t}_t}
{\sum_{\tilde s_t,\tilde d_t}\pi_{s_{t+1}\tilde s_t}\cancel{\rho_{s_{t+1}d_{t+1}}}\alpha^{\tilde\sigma^{1}_t}_t}\gamma^{s_{t+1},d_{t+1},d_{t+1}}_{t+1}\nonumber\\
&=\alpha^{\sigma^{1}_t}_t\sum_{s_{t+1}}\frac{\pi_{s_{t+1}s_t}}{\sum_{\tilde s_t}\pi_{s_{t+1}\tilde s_t}\sum_{\tilde d_t}\alpha^{\tilde\sigma^{1}_t}_t}\sum_{d_{t+1}}\gamma^{s_{t+1},d_{t+1},1}_{t+d_{t+1}}.
\label{eq:gammacountdur}
\end{align}
With pre-summation over $k$ and $j$, the cost of this recursion is ${\cal O}(TS(S+d_{\max}))$.

\subsubsection*{Segmental extended Viterbi}
With the definition $\xi^{\sigma^{1}_t}_t=\max_{s_{1:t-1},d_{1:t-1}}p(s_{1:t-1},d_{1:t-1},\sigma^{1}_t,v_{1:t})$,
the most likely sequence $\sigma^*_{1:T}=\argmax_{\sigma_{1:T}} p(\sigma_{1:T}|v_{1:T})$ can be computed as follows (assuming $c^*_T=1$):\footnote{For $t=1,\ldots,d_{\max}$, $\xi^{\sigma^{1}_t}_t=\bar\alpha^{\sigma^{1}_t}_t$ and $\psi^{s_t,d_t}_t=\emptyset$ if $d_t\geq t$ .}
\begin{align*}
&\textrm{for } t=1,\ldots,T\\
&\hskip0.3cm\left.\begin{array}{l}
\xi^{\sigma^{1}_t}_t=p(v_{t-d_t+1:t}|\sigma^{1}_t)\rho_{s_td_t}\max\limits_{s_{t-d_t}}\pi_{s_ts_{t-d_t}}\max\limits_{d_{t-d_t}}\xi^{\sigma^{1}_{t-d_t}}_{t-d_t}\\
\psi^{s_t,d_t}_t=\argmax\limits_{s_{t-d_t},d_{t-d_t}}\pi_{s_ts_{t-d_t}}\xi^{\sigma^{1}_{t-d_t}}_{t-d_t}
\end{array}\right.\\[5pt]
&\sigma^*_T=
(\argmax\limits_{s_T,d_T} \xi^{\sigma^1_T}_T, 1)\\
&s^*_{T-d^*_T+1:T-1}=s^*_T, \hspace{0.2cm} d^*_{T-d^*_T+1:T-1}=d^*_T, \hspace{0.2cm} c^*_{T-d^*_T+1:T-1}=d^*_T,\ldots,2\\
&t=T\!-\!d^*_T\\
& \textrm{while } t\!\geq\!1\\
&\hskip0.55cm\left.\begin{array}{l}
\sigma^*_t=(\psi^{s^*_{t+1},d^*_{t+1}}_{t+1},1)\\
s^*_{t-d^*_t+1:t-1}=s^*_t,\hspace{0.2cm} d^*_{t-d^*_t+1:t-1}=d^*_t,\hspace{0.2cm}c^*_{t-d^*_t+1:t-1}=d^*_t,\ldots,2\\
t=t-d^*_t\,.
\end{array}\right.
\end{align*}
%
\subsubsection{Segmental learning\label{sec:M2L}}
In this section we show how count-duration variables enable to derive EM updates in a straightforward way.
The relation with the standard approach that uses collapsed count variables is given in Appendix~\ref{app:SS}.

The expectation of the complete data log-likelihood can be written as
\begin{align*}
{\cal L}&=\sum_{t=1}^T \sum_{d_t} \gamma^{\sigma^1_t}_t\log p(v_{t-d_t+1:t}|\sigma^1_t)\\
&+\sum_{s_1} p(s_1|v_{1:T})\log \tilde \pi_{s_1}+\sum_{t=2}^T \sum_{s_{t-1},s_t} p(s_{t-1},c_{t-1}\!=\!1,s_t|v_{1:T})\log \pi_{s_ts_{t-1}}\\
&+\sum_{s_1,d_1} p(s_1,d_1|v_{1:T})\log \tilde \rho_{s_1d_1}+\sum_{t=2}^T \sum_{s_t,d_t} p(c_{t-1}\!=\!1,s_t,d_t|v_{1:T})\log \rho_{s_td_t},
\end{align*}
giving update for $\rho_{s_td_t}$
\begin{align}
\rho_{s_td_t}=\frac{\sum_tp(c_{t-1}\!=\!1,s_t,d_t|v_{1:T})}{\sum_t\sum_{\tilde d_t}p(c_{t-1}\!=\!1,s_t,\tilde d_t|v_{1:T})}
=\frac{\sum_t\gamma^{s_t,d_t,1}_{t+d_t-1}}{\sum_t\sum_{\tilde d_t}\gamma^{s_t,\tilde d_t,1}_{t+\tilde d_t-1}},
\label{eq:updatepho}
\end{align}
as
\begin{align*}
p(c_{t-1}\!=\!1,s_t,d_t|v_{1:T})
&=\cancel{\frac{\sum_{s_{t-1},d_{t-1}}\pi_{s_ts_{t-1}}\rho_{s_td_t}\alpha^{\sigma^1_{t-1}}_{t-1}}
{\sum_{\tilde s_{t-1},\tilde d_{t-1}}\pi_{s_t\tilde s_{t-1}}\rho_{s_td_t}\alpha^{\tilde\sigma^1_{t-1}}_{t-1}}}\gamma^{s_t,d_t,d_t}_t
=\gamma^{s_t,d_t,1}_{t+d_t-1},
\end{align*}
and update for $\pi_{s_ts_{t-1}}$
\begin{align}
\pi_{s_ts_{t-1}}=\frac{\sum_tp(s_{t-1},c_{t-1}\!=\!1,s_t|v_{1:T})}{\sum_t\sum_{\tilde s_t}p(s_{t-1},c_{t-1}\!=\!1,\tilde s_t|v_{1:T})},
\label{eq:updatepi}
\end{align}
where
\begin{align}
\hskip-0.2cm p(s_{t-1},c_{t-1}\!=\!1,s_t|v_{1:T})
&=\frac{\pi_{s_ts_{t-1}}\sum_{d_{t-1}}\alpha^{\sigma^1_{t-1}}_{t-1}}
{\sum_{\tilde s_{t-1}}\pi_{s_t\tilde s_{t-1}}\sum_{\tilde d_{t-1}}\alpha^{\tilde\sigma^1_{t-1}}_{t-1}}\sum_{d_t}\gamma^{s_t,d_t,1}_{t+d_t-1}.
\label{eq:updatepho1}
\end{align}

\section{Explicit-Duration SLGSSM\label{sec:SLGSSM}}
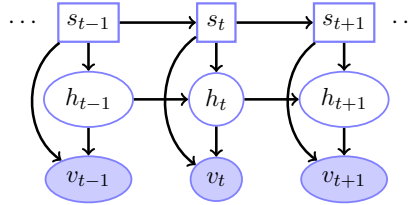
\begin{figure}[t]
\begin{center}
\scalebox{0.85}{
\begin{tikzpicture}[dgraph]
\node[] at (1,-1.05) {$\cdots$};
\node[] at (7,-1.05) {$\cdots$};
\node[disc] (stm) at (2,-1.05) {$s_{t-1}$};
\node[disc] (st) at (4,-1.05) {$s_t$};
\node[disc] (stp) at (6,-1.05) {$s_{t+1}$};
\node[ocont] (htm) at (2,-2.25) {$h_{t-1}$};
\node[ocont] (ht) at (4,-2.25) {$h_t$};
\node[ocont] (htp) at (6,-2.25) {$h_{t+1}$};
\node[ocont,obs] (vtm) at (2,-3.5) {$v_{t-1}$};
\node[ocont,obs] (vt) at (4,-3.5) {$v_t$};
\node[ocont,obs] (vtp) at (6,-3.5) {$v_{t+1}$};
\draw[line width=1.15pt](stm)--(st);\draw[line width=1.15pt](st)--(stp);
\draw[line width=1.15pt](stm)--(htm);\draw[line width=1.15pt](st)--(ht);\draw[line width=1.15pt](stp)--(htp);
\draw[line width=1.15pt](htm)--(vtm);\draw[line width=1.15pt](ht)--(vt);\draw[line width=1.15pt](htp)--(vtp);
\draw[line width=1.15pt](htm)--(ht);\draw[line width=1.15pt](ht)--(htp);
\draw[line width=1.15pt](stm)to [bend right=55](vtm);\draw[line width=1.15pt](st)to [bend right=55](vt);\draw[line width=1.15pt](stp)to [bend right=55](vtp);
\end{tikzpicture}}
\end{center}
\caption{Belief network representation of the switching linear Gaussian state-space model.}
\label{fig:SLGSSM}
\end{figure}
In \secref{sec:VVM} we have seen that the cost of inference in explicit-duration \HMSMs~of type $p(\sigma_{1:T},v_{1:T})$ does not depend on the type of explicit-duration variables used.
This is not the case in models that contain additional unobserved variables $h_{1:T}$ related by Markovian dependence, for which inference is more complex.

\noindent In this section we consider the most popular of such models, namely
the switching linear Gaussian state-space model
(\SLGSSM), also called switching linear dynamical system \citep{barber06ec}.

In the \SLGSSM, $v_t\in\mathbb{R}^V$, $h_t\in\mathbb{R}^H$, and the joint distribution of all variables $p(s_{1:T},h_{1:T},v_{1:T})$ factorizes as
\begin{align*}
p(v_1|h_1,s_1)p(h_1|s_1)p(s_1)\prod_{t=2}^Tp(v_t|h_t,s_t)p(h_t|h_{t-1},s_t)p(s_t|s_{t-1}),
\end{align*}
giving the belief network representation shown in \figref{fig:SLGSSM}. The factors are defined as
\begin{align*}
&p(s_1)=\tilde \pi_{s_1}, \hspace{0.3cm} p(s_t|s_{t-1})=\pi_{s_ts_{t-1}},\\
&p(h_1|s_1)={\cal N}(h_1;\mu^{s_1},\Sigma^{s_1}), \hspace{0.3cm} p(h_t|h_{t-1},s_t)={\cal N}(h_t;A^{s_t}h_{t-1},\Sigma^{s_t}_H),\\
&p(v_t|h_t,s_t)={\cal N}(v_t;B^{s_t}h_t,\Sigma^{s_t}_V),
\end{align*}
where $\mu^{s_1}$ is a $H$-dimensional vector, $\Sigma^{s_1}$, $A^{s_t}$ and $\Sigma^{s_t}_H$ are $H\times H$-dimensional matrices,
$B^{s_t}$ is a $V\times H$-dimensional matrix, and $\Sigma^{s_t}_V$ is a $V\times V$-dimensional matrix.
The model can be equivalently defined by the following linear equations:
\begin{align}
\label{eq:lgssmh}
&h_t=A^{s_t}h_{t-1}+\eta^h_t,\hspace{0.2cm}\eta^h_t\sim{\cal N}(\eta^h_t;0,\Sigma^{s_t}_H),\hspace{0.05cm}h_1\sim{\cal N}(h_1;\mu^{s_t},\Sigma^{s_t}),\\
\label{eq:lgssmv}
&v_t=B^{s_t}h_t+\eta^v_t,\hspace{0.2cm}\eta^v_t\sim{\cal N}(\eta^v_t;0,\Sigma^{s_t}_V).
\end{align}
Performing inference in the \SLGSSM~requires approximations since, \eg, $p(h_t|v_{1:t})$ is a Gaussian mixture with $S^t$ components\footnote{This explosion of mixture components with time can be understood by noticing that $p(h_t|v_{1:t})$ is given by $\sum_{s_{1:T}}p(h_t|s_{1:t},\cancel{s_{t+1:T}},v_{1:t})p(s_{1:T}|v_{1:t})=\sum_{s_{1:t}}p(h_t|s_{1:t},v_{1:t})p(s_{1:t}|v_{1:t})$ and that $p(h_t|s_{1:t},v_{1:t})$ is Gaussian.}.
In the expectation-correction (EC) approach of \cite{barber06ec},
the filtered distribution $p(h_t,s_t|v_{1:t})$ is first computed by forming separate recursions for $p(h_t|s_t,v_{1:t})$ and $p(s_t|v_{1:t})$, and then used
to compute the smoothed distribution $p(h_t,s_t|v_{1:T})$ by forming separate recursions for $p(h_t|s_t,v_{1:T})$ and $p(s_t|v_{1:T})$.
The recursions are similar to the sequential filtering-smoothing recursions used in \secref{sec:ILEX}. The explosion of mixture components
with time is addressed by collapsing, at each time-step, the obtained Gaussian mixture to a Gaussian mixture with a lower number of components
\citep{alspach72nonlinear}. 
In addition to Gaussian collapsing, EC introduces one approximation in the recursion for $p(h_t|s_t,v_{1:T})$, due to lack of knowledge about the regime at the previous time-step,
and one approximation in the recursion for $p(s_t|v_{1:T})$. The resulting routines for $p(h_t|s_t,v_{1:t})$
and $p(h_t|s_t,v_{1:T})$ resemble the standard predictor-corrector filtering
routines and Rauch-Tung-Striebel smoothing routines of the
linear Gaussian state-space model (\LGSSM) \citep{rauch65maximum,grewal93kalman,chiappa06phd}.
\begin{figure}[t] 
\hskip-0.03cm
\subfigure[]{
\includegraphics[]{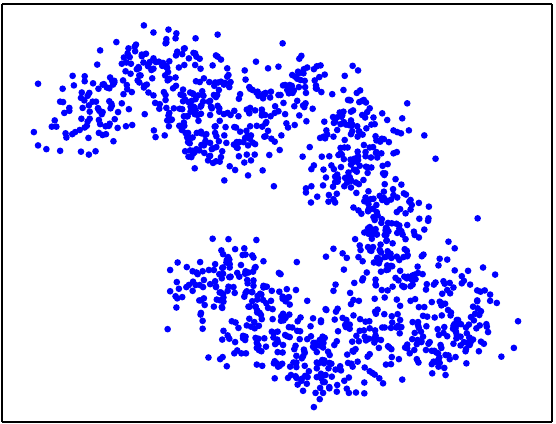}}
\subfigure[]{
\includegraphics[]{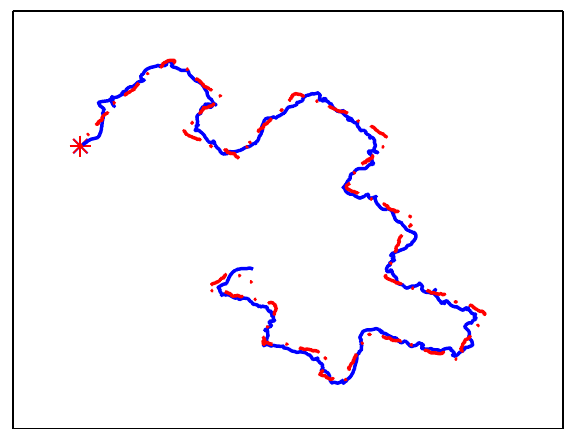}}
\caption{(a): Noisy
measurements of the positions of a two-wheeled robot moving in the two-dimensional
space generated from model (\ref{eq:roblgssm}). (b): Actual positions (dashed line) and estimated positions (continuous line) by a \SLGSSM~(means of $p(h_t|v_{1:T})$). The initial position is indicated with a star.}
\label{fig:robpos}
\end{figure}

The \SLGSSM~enables sophisticated modelling and estimation of hidden dynamics underlying noisy observations \citep{pavlovic01learning,zoeter05ep,mesot-barber-07a,chiappa08bayesian,quinn08factorial}. It can
can be used, \eg, to solve the robot localization problem discussed in \chapref{chap:intro},
namely to infer the positions of a two-wheeled robot moving in the two-dimensional
space plotted in \figref{fig:robpos}(b) with a dashed line
from the noisy measurements plotted in \figref{fig:robpos}(a).
As explained in detail in Appendix~\ref{app:RL}, the hidden dynamics and observation process can be formulated as a \SLGSSM~with nonlinear hidden dynamics.
The means of $p(h_t|v_{1:T})$, computed by combining EC with an unscented approximation \citep{sarkka8unscented}, give reasonably accurate estimates
of the positions, as shown in \figref{fig:robpos}(b) with a continuous line\footnote{The hidden dynamics $f^{s_t}$ (see Appendix~\ref{app:RL}),
$\mu^{s_1}$, $\Sigma^{s_1}$, $\Sigma^{s_t}_H$, and $\Sigma^{s_t}_V$ were assumed to be known, and maximum likelihood values of $\tilde \pi$ and $\pi$ were computed using the correct regimes.}.

In the \SLGSSM, all three approaches to explicit-duration modelling can be used (as the Markovian structure of $h_{1:T}$ enables recursive computation of $p(v_t|\sigma_t,v_{1:t-1})$)
and allow $p(v_{1:T}|\sigma_{1:T})$ to factorize across segments. 
Following closely EC, we describe a sequential filtering-smoothing approach that allows to generalize the results to similar models with unobserved variables related by first-order Markovian dependence.
In this approach, the filtered distribution $p(h_t,\sigma_t|v_{1:t})$ is first computed by
forming separate recursions for $\hat\alpha^{\sigma_t}_t=p(h_t|\sigma_t,v_{1:t})$ and $\alpha^{\sigma_t}_t=p(\sigma_t|v_{1:t})$,
and then used to compute the smoothed distribution $p(h_t,\sigma_t|v_{1:T})$ by
forming separate recursions $\hat\gamma^{\sigma_t}_t=p(h_t|\sigma_t,v_{1:T})$ and $\gamma^{\sigma_t}_t=p(\sigma_t|v_{1:T})$.

To gain some intuition about the differences between the three approaches, we can observe that inference on $h_{1:T}$ needs to consider all possible segmentations,
\ie~all possible partitioning of the \timeseries~into segments and, for each partitioning, all possible combinations of regimes.

In the across-segment-independence case, inference on $h_{1:T}$ given a segmentation reduces to inference in a separate \LGSSM~for each segment and regime.
Since the set of unique segments generated by all possible segmentations is $\{v_{t:t+d_t-1}, \forall t, \forall d_t\}$,
only \LGSSM~filtering-smoothing on segment $v_{t:t+d_t-1}$ for each $t$, $s_t$ and $d_t$ is required. Furthermore, as filtering can be shared between all segments that start at the same time-step and
are generated from the same regime, only \LGSSM~filtering on segment $v_{t:t+d_{\max}-1}$ for each $t$ and $s_t$ is required. Therefore, the computational cost of inference on $h_{1:T}$
for all possible segmentations is ${\cal O}(TSd_{\max})$ for filtering and ${\cal O}(TSd^2_{\max})$ for smoothing.
Computing $p(h_t|s_t,v_{1:t})$ requires to sum over all possible starts of the segment passing through time-step $t$, giving rise to a Gaussian mixture with $d_{\max}$ components.
This means that, regardless of the explicit-duration encoding used, $p(h_t|s_t,v_{1:t})$ cannot be simpler than a Gaussian mixture with $d_{\max}$ components.
Similarly, computing $p(h_t|s_t,v_{1:T})$ requires to sum over all possible starts and ends of the segment passing through time-step $t$, giving rise to a Gaussian mixture with number of components (of order) $d^2_{\max}$.
Therefore, $p(h_t|s_t,v_{1:T})$ cannot be simpler than a Gaussian mixture with number of components (of order) $d^2_{\max}$.
If knowledge about segment beginning is explicitly encoded in the explicit-duration variables, $\hat\alpha^{\sigma_t}_t$ is a Gaussian distribution,
and the mixture in $p(h_t|s_t,v_{1:t})$ arises from summing over the explicit-duration variables. If knowledge about segment beginning is not explicitly encoded in the explicit-duration variables,
$\hat\alpha^{\sigma_t}_t$ is a Gaussian mixture with (maximally, as segment end is encoded in this case) $d_{\max}$ components.
Similarly, $\hat\gamma^{\sigma_t}_t$ is a Gaussian distribution if knowledge about both segment beginning and segment end is explicitly encoded in the explicit-duration variables,
and a Gaussian mixture with maximally $d_{\max}$ components otherwise.

Decreasing count variables encode information about segment end.
The recursion for $\hat\alpha^{\sigma_t}_t$ (recursion (\ref{eq:filtcat})) produces a Gaussian mixture with maximally $d_{\max}$ components accounting for all possible segment starts, 
and therefore has computational cost ${\cal O}(TSd^2_{\max})$. The cost can be reduced to ${\cal O}(TSd_{\max})$
with Gaussian collapsing. The recursion for $\hat\gamma^{\sigma_t}_t$ (recursion (\ref{eq:smoothM1cat})) does not increase the number of components, as segment end is known.
Without Gaussian collapsing of $\hat\alpha^{\sigma_t}_t$, the cost of the recursion is therefore ${\cal O}(TSd^2_{\max})$. With Gaussian collapsing of $\hat\alpha^{\sigma_t}_t$,
the cost is reduced to ${\cal O}(TSd_{\max})$; however, as knowledge about segment beginning is lost, the EC approximation $p(h_{t+1}|s_t,c_t>1,\sigma_{t+1},v_{1:T})\approx \hat\gamma^{\sigma_{t+1}}_{t+1}$ in
\eqref{eq:approx1} is required.

Increasing count variables encode information about segment beginning. The recursion for $\hat\alpha^{\sigma_t}_t$ (recursion (\ref{eq:filtM1altcat})) produces a Gaussian distribution, and therefore has cost ${\cal O}(TSd_{\max})$.
This recursion essentially performs \LGSSM~filtering on segment $v_{t:t+d_{\max}-1}$ for each $t$ and $s_t$. The recursion for $\hat\gamma^{\sigma_t}_t$ (recursion (\ref{eq:smoothM1altcat})) produces a Gaussian mixture
with maximally $d_{\max}$ components, which accounts for all possible segment ends, and therefore has cost ${\cal O}(TSd^2_{\max})$. The cost can be reduced to ${\cal O}(TSd_{\max})$
with Gaussian collapsing.

Count-duration variables encode information about both segment beginning and segment end.
The estimation of $\hat\alpha^{\sigma_t}_t$ and $\hat\gamma^{\sigma_t}_t$ can be recast into filtering and smoothing in a \LGSSM, and therefore has cost ${\cal O}(TSd_{\max})$ and ${\cal O}(TSd^2_{\max})$ respectively.
Alternatively, the estimation can be achieved with time-recursive routines that produce Gaussian distributions and
have the same cost (recursions (\ref{eq:alphahatM2t}) and (\ref{eq:gammahatM2t})). The cost ${\cal O}(TSd_{\max})$ rather than ${\cal O}(TSd^2_{\max})$ in the recursion for $\hat\alpha^{\sigma_t}_t$  is achieved by taking care of redundancies.
\begin{table}[t]
\vspace{0.2cm}
\hspace{1.2cm}
\scalebox{0.75}{
\begin{tabular}{|c|c|c|}
\hline
\Tstrut \multirow{11}{*}{\rotatebox{90}{ Decreasing Count Variables}} & $\hat\alpha^{\sigma_t}_t$ & $\alpha^{\sigma_t}_t$\\[6pt]
\cline{2-3}
& \Tstrut GM with maximally $d_{\max}$ components: ${\cal O}(TSd_{\max}^2)$ & \multirow{2}{*}{${\cal O}(TS^2d_{\max})$}\\[6pt]
& Gaussian collapsing: ${\cal O}(TSd_{\max})$ &  \\[6pt]
\cline{2-3}
& \Tstrut $\hat\gamma^{\sigma_t}_t$ & $\gamma^{\sigma_t}_t$\\[6pt]
\cline{2-3}
& \Tstrut GM with maximally $d_{\max}$ components: ${\cal O}(TSd_{\max}^2)$ &\\[6pt]
& Gaussian collapsing of $\alpha^{\sigma_t}_t$: ${\cal O}(TSd_{\max})$ & ${\cal O}(TS^2d_{\max})$\\[5pt]
& $p(h_{t+1}|s_t,c_t > 1,\sigma_{t+1},v_{1:T}) \approx  \hat\gamma^{\sigma_{t+1}}_{t+1}$ & \\[5pt]
\hline
\hline
\Tstrut \multirow{8}{*}{\vspace{0cm}\rotatebox{90}{\hskip0.0cm Increasing Count Var.}}  & $\hat\alpha^{\sigma_t}_t$ & $\alpha^{\sigma_t}_t$\\[6pt]
\cline{2-3}
& \Tstrut Gaussian: ${\cal O}(TSd_{\max})$ & ${\cal O}(TS^2d_{\max})$\\[6pt]
\cline{2-3}
& \Tstrut $\hat\gamma^{\sigma_t}_t$ & $\gamma^{\sigma_t}_t$\\[6pt]
\cline{2-3}
& \Tstrut GM with maximally $d_{\max}$ components: ${\cal O}(TSd_{\max}^2)$ &  \multirow{2}{*}{${\cal O}(TS^2d_{\max})$}\\[6pt]
& Gaussian collapsing: ${\cal O}(TSd_{\max})$ & \\[5pt]
\hline
\hline
\Tstrut \multirow{1}{*}{\rotatebox{90}{\hskip0.0cm Count-Duration Var.}}  & $\hat\alpha^{\sigma_t}_t$ & $\alpha^{\sigma_t}_t$\\[6pt]
\cline{2-3}
& \Tstrut Gaussian: ${\cal O}(TSd_{\max})$ & ${\cal O}(TS^2d_{\max})$\\[6pt]
\cline{2-3}
& \Tstrut $\hat\gamma^{\sigma_t}_t$ & $\gamma^{\sigma_t}_t$\\[6pt]
\cline{2-3}
& \Tstrut Gaussian: ${\cal O}(TSd_{\max}^2)$ & ${\cal O}(TS^2d_{\max})$\\[6pt]
\hline
\end{tabular}
}
\caption{Characteristics of the different encodings for the explicit-duration \SLGSSM~with across-segment independence. GM indicates Gaussian mixture.}
\label{table:HVMC}
\end{table}

In all three approaches, the estimation of $\alpha^{\sigma_t}_t$ and $\gamma^{\sigma_t}_t$ has cost ${\cal O}(TS^2d_{\max})$\footnote{For simplicity of exposition, we do not consider
the possibility to reduce the cost to ${\cal O}(TS(S+d_{\max}))$ in this model.}.
The computation of $\alpha^{\sigma_t}_t$ requires filtering on $h_{1:T}$
and, if decreasing count variables are used, the computation of $\gamma^{\sigma_t}_t$ requires smoothing on $h_{1:T}$.

In summary, increasing count variables and count-duration variables have the advantage over decreasing count variables of requiring only filtering on $h_{1:T}$ to perform segmentation.
Without Gaussian collapsing, increasing count variables and count-duration variables give the same computational cost.
They are advantageous over decreasing count variables as filtering on $h_{1:T}$ has lower cost and as smoothing on $h_{1:T}$ is simpler.
The count-duration-variable approach is more intuitive than the increasing-count-variable approach.
However, increasing count variables do not require taking care of redundancies.
With Gaussian collapsing, which can be performed in filtering with decreasing count variables and in smoothing with increasing count variables,
count variables give that same computational cost, which is lower in smoothing on $h_{1:T}$ than with count-duration variables. However,
decreasing count variables require the EC approximation $p(h_{t+1}|s_t,c_t>1,\sigma_{t+1},v_{1:T})\approx \hat\gamma^{\sigma_{t+1}}_{t+1}$ in
\eqref{eq:approx1}. In similar models in which $h_{1:T}$ are discrete, similar conclusions to the Gaussian collapsing case can be made. The characteristics are summarized in Table \ref{table:HVMC}.

In the across-segment-dependence case, explicit-duration modelling increases the computational complexity with respect to the standard \SLGSSM, and therefore Gaussian collapsing is required.
If time-step $t$ corresponds to the beginning of a segment, $c_{t-1}$ must have value 1 in the decreasing-count-variable approach and can take any value in the increasing-count-variable approach.
This means that the recursion for $\hat\alpha^{\sigma_t}_t$ using decreasing count variables (recursion (\ref{eq:filt})) produces a Gaussian mixture with less components
than the recursion for $\hat\alpha^{\sigma_t}_t$ using increasing count variables (recursion (\ref{eq:filtM1alt})).
The reverse happens in the recursion for $\hat\gamma^{\sigma_t}_t$ (recursions (\ref{eq:smoothM1}) and (\ref{eq:smoothM1Alt})).
Count-duration variables (requiring time-recursive inference) give rise to more complex Gaussian mixtures than count variables.
Gaussian collapsing reduces the cost of the recursions for $\hat\alpha^{\sigma_t}_t$, and $\hat\gamma^{\sigma_t}_t$ to ${\cal O}(TS^2d_{\max})$ in the count-variable approaches
and to ${\cal O}(TS^2d^2_{\max})$ in the count-duration-variable approach. The computation of $\alpha^{\sigma_t}_t$ and $\gamma^{\sigma_t}_t$ has cost ${\cal O}(TS^2d_{\max})$ in all approaches.
Unlike decreasing count variables, increasing count variables and count-duration variables require the EC approximations
only for $c_t=1$.
In similar models with discrete unobserved variables related by first-order Markovian dependence, similar conclusions to the Gaussian collapsing case can be made.

Therefore, the increasing count variable approach is overall preferable in both the across-segment-independence and across-segment-dependence cases.

In the following sections we describe the three approaches in more detail.

The explicit-duration \SLGSSM~is also discussed in \cite{oh08learning} using increasing count variables and in \cite{bracegirdle11switch} and \cite{bracegirdle13inference} in the context of reset models (see \secref{sec:Approx}).
\cite{bracegirdle11switch} and \cite{bracegirdle13inference} present a recursion for $\hat\alpha^{\sigma_t}_t$ using increasing count variables that is equivalent to recursion (\ref{eq:filtM1altcat}), and a recursion for $\hat\gamma^{\sigma_t}_t$ using
increasing-decreasing count variables with cost ${\cal O}(TSd^2_{\max})$. Increasing-decreasing count variables provide the same information as count-duration variables but give rise to more convoluted recursions.
The computation of the smoothed distributions in the increasing-decreasing-count-variable representation using filtered distributions computed in the increasing-count-variable representation is possible as across-segment-dependence is cut.

\subsection{Decreasing Count Variables \label{sec:SLGSSMDec}}
The explicit-duration \SLGSSM~using decreasing count variables has belief network representation given in \figref{fig:SLGSSMDec}(a).
Across-segment independence can be enforced by adding a link from $c_t$ to $h_{t+1}$, as in \figref{fig:SLGSSMDec}(b),
which has the effect of removing the link from $h_t$ to $h_{t+1}$ if $c_t=1$, as explicitly represented in \figref{fig:SLGSSMDec}(c).
More specifically, dependence cut is defined as
\begin{align*}
p(h_t|h_{t-1},c_{t-1},s_t)&=\begin{cases}
    p(h_t|c_{t-1},s_t)={\cal N}(h_t;\mu^{s_t},\Sigma^{s_t}) & \hspace{-0.0cm} \textrm{if } c_{t-1}\!=\!1\\
	{\cal N}(h_t;A^{s_t}h_{t-1},\Sigma^{s_t}_H) 	& \hspace{-0.0cm} \textrm{if } c_{t-1}\!>\!1.
\end{cases}
\end{align*}

\subsubsection*{Filtering}
\begin{figure}[t]
\hskip-0.1cm
\subfigure[]{\scalebox{0.65}{
\begin{tikzpicture}[dgraph]
\node[disc] (sigmatm) at (2,0.2) {$c_{t-1}$};
\node[disc] (sigmat) at (4,0.2) {$c_t$};
\node[disc] (sigmatp) at (6,0.2) {$c_{t+1}$};
\node[disc] (stm) at (2,-1.05) {$s_{t-1}$};
\node[disc] (st) at (4,-1.05) {$s_t$};
\node[disc] (stp) at (6,-1.05) {$s_{t+1}$};
\node[ocont] (htm) at (2,-2.25) {$h_{t-1}$};
\node[ocont] (ht) at (4,-2.25) {$h_t$};
\node[ocont] (htp) at (6,-2.25) {$h_{t+1}$};
\node[ocont,obs] (vtm) at (2,-3.5) {$v_{t-1}$};
\node[ocont,obs] (vt) at (4,-3.5) {$v_t$};
\node[ocont,obs] (vtp) at (6,-3.5) {$v_{t+1}$};
\draw[line width=1.15pt](sigmatm)--(sigmat);\draw[line width=1.15pt](sigmat)--(sigmatp);
\draw[line width=1.15pt](stm)--(st);\draw[line width=1.15pt](st)--(stp);
\draw[line width=1.15pt](stm)--(htm);\draw[line width=1.15pt](st)--(ht);\draw[line width=1.15pt](stp)--(htp);
\draw[line width=1.15pt](sigmatm)--(st);\draw[line width=1.15pt](sigmat)--(stp);
\draw[line width=1.15pt](htm)--(vtm);\draw[line width=1.15pt](ht)--(vt);\draw[line width=1.15pt](htp)--(vtp);
\draw[line width=1.15pt](htm)--(ht);\draw[line width=1.15pt](ht)--(htp);
\draw[line width=1.15pt](stm)to [bend right=55](vtm);\draw[line width=1.15pt](st)to [bend right=55](vt);\draw[line width=1.15pt](stp)to [bend right=55](vtp);
\draw[->, line width=1.15pt](stm)--(sigmatm);\draw[->, line width=1.15pt](st)--(sigmat);\draw[->, line width=1.15pt](stp)--(sigmatp);
\end{tikzpicture}}}
\hskip0.1cm
\subfigure[]{\scalebox{0.65}{
\begin{tikzpicture}[dgraph]
\node[disc] (sigmatm) at (2,0.2) {$c_{t-1}$};
\node[disc] (sigmat) at (4,0.2) {$c_t$};
\node[disc] (sigmatp) at (6,0.2) {$c_{t+1}$};
\node[disc] (stm) at (2,-1.05) {$s_{t-1}$};
\node[disc] (st) at (4,-1.05) {$s_t$};
\node[disc] (stp) at (6,-1.05) {$s_{t+1}$};
\node[ocont] (htm) at (2,-2.25) {$h_{t-1}$};
\node[ocont] (ht) at (4,-2.25) {$h_t$};
\node[ocont] (htp) at (6,-2.25) {$h_{t+1}$};
\node[ocont,obs] (vtm) at (2,-3.5) {$v_{t-1}$};
\node[ocont,obs] (vt) at (4,-3.5) {$v_t$};
\node[ocont,obs] (vtp) at (6,-3.5) {$v_{t+1}$};
\draw[color=myred,line width=1.15pt](sigmatm)--(ht);\draw[color=myred,line width=1.15pt](sigmat)--(htp);
\draw[line width=1.15pt](sigmatm)--(sigmat);\draw[line width=1.15pt](sigmat)--(sigmatp);
\draw[line width=1.15pt](stm)--(st);\draw[line width=1.15pt](st)--(stp);
\draw[line width=1.15pt](stm)--(htm);\draw[line width=1.15pt](st)--(ht);\draw[line width=1.15pt](stp)--(htp);
\draw[line width=1.15pt](sigmatm)--(st);\draw[line width=1.15pt](sigmat)--(stp);
\draw[line width=1.15pt](htm)--(vtm);\draw[line width=1.15pt](ht)--(vt);\draw[line width=1.15pt](htp)--(vtp);
\draw[color=myred,line width=1.15pt](htm)--(ht);\draw[color=myred,line width=1.15pt](ht)--(htp);
\draw[line width=1.15pt](stm)to [bend right=55](vtm);\draw[line width=1.15pt](st)to [bend right=55](vt);\draw[line width=1.15pt](stp)to [bend right=55](vtp);
\draw[->, line width=1.15pt](stm)--(sigmatm);\draw[->, line width=1.15pt](st)--(sigmat);\draw[->, line width=1.15pt](stp)--(sigmatp);
\end{tikzpicture}}}
\hskip0.1cm
\subfigure[]{\scalebox{0.65}{
\begin{tikzpicture}[dgraph]
\node[disc] (sigmatm) at (2,0.2) {$c_{t-1}$};
\node[disc] (sigmat) at (4,0.2) {$c_t$};
\node[disc] (sigmatp) at (6,0.2) {$c_{t+1}$};
\node[disc] (stm) at (2,-1.05) {$s_{t-1}$};
\node[disc] (st) at (4,-1.05) {$s_t$};
\node[disc] (stp) at (6,-1.05) {$s_{t+1}$};
\node[ocont] (htm) at (2,-2.25) {$h_{t-1}$};
\node[ocont] (ht) at (4,-2.25) {$h_t$};
\node[ocont] (htp) at (6,-2.25) {$h_{t+1}$};
\node[ocont,obs] (vtm) at (2,-3.5) {$v_{t-1}$};
\node[ocont,obs] (vt) at (4,-3.5) {$v_t$};
\node[ocont,obs] (vtp) at (6,-3.5) {$v_{t+1}$};
\node [color=myred,above] at (sigmat.north) {$c_t=1$};
\draw[line width=1.15pt](sigmatm)--(ht);\draw[line width=1.15pt](sigmat)--(htp);
\draw[line width=1.15pt](sigmatm)--(sigmat);\draw[line width=1.15pt](sigmat)--(sigmatp);
\draw[line width=1.15pt](stm)--(st);\draw[line width=1.15pt](st)--(stp);
\draw[line width=1.15pt](stm)--(htm);\draw[line width=1.15pt](st)--(ht);\draw[line width=1.15pt](stp)--(htp);
\draw[line width=1.15pt](sigmatm)--(st);\draw[line width=1.15pt](sigmat)--(stp);
\draw[line width=1.15pt](htm)--(vtm);\draw[line width=1.15pt](ht)--(vt);\draw[line width=1.15pt](htp)--(vtp);
\draw[line width=1.15pt](htm)--(ht);
\draw[line width=1.15pt](stm)to [bend right=55](vtm);\draw[line width=1.15pt](st)to [bend right=55](vt);\draw[line width=1.15pt](stp)to [bend right=55](vtp);
\draw[->, line width=1.15pt](stm)--(sigmatm);\draw[->, line width=1.15pt](st)--(sigmat);\draw[->, line width=1.15pt](stp)--(sigmatp);
\end{tikzpicture}}}
\caption{(a): Explicit-duration \SLGSSM~using decreasing count variables.
(b): Across-segment independence is enforced with a link from $c_t$ to $h_{t+1}$,
as explicitly represented in (c).}
\label{fig:SLGSSMDec}
\end{figure}
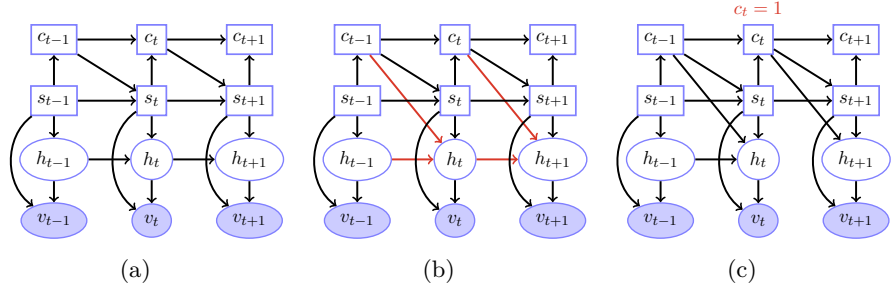
To compute the filtered distribution $p(h_t,\sigma_t|v_{1:t})$, we form separate recursions for $\alpha^{\sigma_t}_t=p(\sigma_t|v_{1:t})$ and $\hat\alpha^{\sigma_t}_t=p(h_t|\sigma_t,v_{1:t})$.

The recursion for $\alpha^{\sigma_t}_t$ is given by\footnote{Notice the similarity with recursion (\ref{eq:alphaM1st}).}
\begin{align*}
\alpha^{\sigma_t}_t&=\frac{\sum_{\sigma_{t-1}}p(\sigma_{t-1:t},v_t|v_{1:t-1})}{\sum_{\tilde\sigma_{t-1:t}}p(\tilde\sigma_{t-1:t},v_t|v_{1:t-1})}\\
&\propto\sum_{\sigma_{t-1}} p(v_t|\sigma_{t-1},s_t,\cancel{c_t},v_{1:t-1})p(\sigma_t|\sigma_{t-1},\cancel{v_{1:t-1}})p(\sigma_{t-1}|v_{1:t-1})\\
&=\bigg\{\delta_{\begin{subarray}{l} c_t<d_{\max} \\ s_{t-1}=s_t \\ c_{t-1}=c_t+1 \end{subarray}}
+\delta_{\begin{subarray}{l} c_t\geq d_{\min}\\c_{t-1}=1\end{subarray}}
\rho_{\sigma_t}\sum_{s_{t-1}}\pi_{s_ts_{t-1}}\bigg\}e^{\sigma_{t-1},s_t}_t\alpha^{\sigma_{t-1}}_{t-1},
\end{align*}
where (see \eqref{eq:predv} below) $e^{\sigma_{t-1},s_t}_t=p(v_t|\sigma_{t-1},s_t,v_{1:t-1})={\cal N}(v_t;B^{s_t}\hat h_t^{t-1,\sigma_{t-1},s_t},B^{s_t}P_{t}^{t-1,\sigma_{t-1},s_t}(B^{s_t})\trans+\Sigma^{s_t}_V)$,
with the symbol $\trans$ denoting the transpose operator.
Notice that $v_t \cancel{\ci} c_{t-1}|\{s_{t-1:t},v_{1:t-1}\}$ as the path $c_{t-1},c_{t-2},s_{t-2},h_{t-2:t},v_t$ in \figref{fig:SLGSSMDec}(a) is not blocked.
This recursion has computational cost ${\cal O}(TS^2d_{\max})$.

\noindent The recursion for $\hat\alpha^{\sigma_t}_t$ is given by
\begin{align}
\label{eq:filt}
&\hat\alpha^{\sigma_t}_t=\sum_{\sigma_{t-1}}p(h_t|\sigma_{t-1},s_t,\cancel{c_t},v_{1:t})p(\sigma_{t-1}|\sigma_t,v_{1:t})\\
&\!=\!\sum_{\sigma_{t-1}}p(h_t|\sigma_{t-1},s_t,v_{1:t})\frac{p(\sigma_{t-1:t},v_t|v_{1:t-1})}{\sum_{\tilde\sigma_{t-1}} p(\tilde\sigma_{t-1},\sigma_t,v_t|v_{1:t-1})}\nonumber\\[5pt]
&\!=\!\frac{1}{n^{\sigma_t}_t}\bigg\{\delta_{\begin{subarray}{l} c_t<d_{\max} \\ s_{t-1}=s_t \\ c_{t-1}=c_t+1  \end{subarray}}
\hskip-0.4cm +\delta_{\begin{subarray}{l} c_t\geq d_{\min} \\ c_{t-1}=1 \end{subarray}}
\rho_{\sigma_t}\hskip-0.0cm\sum_{s_{t-1}}\hskip-0.0cm\pi_{s_ts_{t-1}}\bigg\}e^{\sigma_{t-1},s_t}_t\alpha^{\sigma_{t-1}}_{t-1}p(h_t|\sigma_{t-1},s_t,v_{1:t}),\nonumber
\end{align}
with $n^{\sigma_t}_t=\sum_{\tilde\sigma_{t-1}} p(v_t,\tilde\sigma_{t-1},\sigma_t|v_{1:t-1})$ and\footnote{The notation $\int_{x}$ indicates integration over the entire range of $x$.}
\begin{align*}
p(h_t&|\sigma_{t-1},s_t,v_{1:t})
=\frac{p(v_t|h_t,\cancel{\sigma_{t-1}},s_t,\cancel{v_{1:t-1}})p(h_t|\sigma_{t-1},s_t,v_{1:t-1})}{p(v_t|\sigma_{t-1},s_t,v_{1:t-1})}\\
&=\frac{p(v_t|h_t,s_t)\int_{h_{t-1}}p(h_t|h_{t-1},\cancel{\sigma_{t-1}},s_t,\cancel{v_{1:t-1}})p(h_{t-1}|\sigma_{t-1},\cancel{s_t},v_{1:t-1})}{p(v_t|\sigma_{t-1},s_t,v_{1:t-1})}\\
&=\frac{p(v_t|h_t,s_t)\int_{h_{t-1}}p(h_t|h_{t-1},s_t)\hat\alpha^{\sigma_{t-1}}_{t-1}}{p(v_t|\sigma_{t-1},s_t,v_{1:t-1})}.
\end{align*}
If we assume $\hat\alpha^{\sigma_{t-1}}_{t-1}$ to be Gaussian with mean $\hat h_{t-1}^{t-1,\sigma_{t-1}}$\footnote{In this notation the lower index $t-1$ refers to $h_{t-1}$, whilst the upper index $t-1$ refers to conditioning on $v_{1:t-1}$.} and covariance $P_{t-1}^{t-1,\sigma_{t-1}}$,
rather than using the equation above, we can obtain $p(h_t|\sigma_{t-1},s_t,v_{1:t})$ more directly from the rules of linear transformations of Gaussian variables.
More specifically, from \eqref{eq:lgssmh} we deduce that $p(h_t|\sigma_{t-1},s_t,v_{1:t-1})$ is Gaussian with mean and covariance given by
\begin{align}
\hat h_t^{t-1,\sigma_{t-1},s_t}&=\av{h_t}_{p(h_t|\sigma_{t-1},s_t,v_{1:t-1})}=A^{s_t}\hat h_{t-1}^{t-1,\sigma_{t-1}},\nonumber\\[7pt]
P_t^{t-1,\sigma_{t-1},s_t}&=\av{(h_t-\hat h_t^{t-1,\sigma_{t-1},s_t})(h_t-\hat h_t^{t-1,\sigma_{t-1},s_t})\trans}_{p(h_t|\sigma_{t-1},s_t,v_{1:t-1})}\nonumber\\
&=A^{s_t}P_{t-1}^{t-1,\sigma_{t-1}}(A^{s_t})\trans+\Sigma^{s_t}_H.
\label{eq:pred}
\end{align}
Furthermore, from \eqref{eq:lgssmv} we deduce
\begin{align}
\label{eq:predv}
&\av{v_t}_{p(v_t|\sigma_{t-1},s_t,v_{1:t-1})}=B^{s_t}\hat h_t^{t-1,\sigma_{t-1},s_t}\,,\\[5pt]
&\av{(v_t-\av{v_t})(v_t-\av{v_t})\trans}_{p(v_t|\sigma_{t-1},s_t,v_{1:t-1})}=B^{s_t}P_t^{t-1,\sigma_{t-1},s_t}(B^{s_t})\trans+\Sigma^{s_t}_V\,,\nonumber\\[5pt]
&\av{(v_t-\av{v_t})(h_t-\hat h_t^{t-1,\sigma_{t-1},s_t})\trans}_{p(v_t,h_t|\sigma_{t-1},s_t,v_{1:t-1})}=B^{s_t}P_{t}^{t-1,\sigma_{t-1},s_t}\,.\nonumber
\end{align}
Finally, by using the formula of Gaussian conditioning, we deduce that $p(h_t|\sigma_{t-1},s_t,v_{1:t})$ is Gaussian with mean and covariance given by
\begin{align}
&\hat h_t^{t,\sigma_{t-1},s_t}=\hat h_t^{t-1,\sigma_{t-1},s_t}\!+\!K(v_t\!-\!B^{s_t}\hat h_t^{t-1,\sigma_{t-1},s_t}),\nonumber\\[5pt]
&P_t^{t,\sigma_{t-1},s_t}=(I\!-\!KB^{s_t})P_t^{t-1,\sigma_{t-1},s_t},
\label{eq:corr}
\end{align}
where $K=P_t^{t-1,\sigma_{t-1},s_t}(B^{s_t})\trans(B^{s_t}P_t^{t-1,\sigma_{t-1},s_t}(B^{s_t})\trans+\Sigma^{s_t}_V)^{-1}$ and $I$ is the identity matrix.
More generally, if $\hat\alpha^{\sigma_{t-1}}_{t-1}$ is a Gaussian mixture, $p(h_t|\sigma_{t-1},s_t,v_{1:t})$ is also a Gaussian mixture with the same number of components.

At time-step $t=1$, $\hat\alpha^{\sigma_t}_t$ is Gaussian with mean and covariance
\begin{align}
&\hat h_t^{t,s_t}=\mu^{s_t}\!+\!K_1(v_t\!-\!B^{s_t}\mu^{s_t}), \hskip0.7cm P_t^{t,s_t}=(I\!-\!K_1B^{s_t})\Sigma^{s_t},
\label{eq:predcorri}
\end{align}
where $K_1=\Sigma^{s_t}(B^{s_t})\trans(B^{s_t}\Sigma^{s_t}(B^{s_t})\trans+\Sigma^{s_t}_V)^{-1}$.

Notice that, if we remove dependence on $\sigma_{t-1},s_t$, Equations (\ref{eq:pred}), (\ref{eq:corr}) and (\ref{eq:predcorri}) become the standard predictor-corrector routines of the \LGSSM~(\cite{grewal93kalman,chiappa06phd}).

As $\hat\alpha^{\sigma_1}_1$ is Gaussian, from the reasoning above and recursion (\ref{eq:filt}) we deduce that $\hat\alpha^{\sigma_2}_2$ is a Gaussian mixture with $S$ components and, more generally, that at each time-step the number of components is multiplied by $S$,
so that $\hat\alpha^{\sigma_t}_t$ is a Gaussian mixture with $S^{t-1}$ components.
Therefore, the recursion for $\hat\alpha^{\sigma_t}_t$ has cost ${\cal O}(TS^td_{\max})$.
The collapsing of $\hat\alpha^{\sigma_t}_t$ to a Gaussian distribution by moment matching, \ie
\begin{align*}
\hat h_t^{t,\sigma_t}&=\frac{1}{n^{\sigma_t}_t}\bigg\{\delta_{\begin{subarray}{l} c_t<d_{\max}\\ s_{t-1}=s_t \\ c_{t-1}=c_t+1\end{subarray}}\hskip-0.3cm+\delta_{\begin{subarray}{l} c_t\geq d_{\min} \\ c_{t-1}=1 \end{subarray}}
\rho_{\sigma_t}\hskip-0.0cm\sum\limits_{s_{t-1}}\hskip-0.0cm\pi_{s_ts_{t-1}}\bigg\}
e^{\sigma_{t-1},s_t}_t\alpha^{\sigma_{t-1}}_{t-1}\hat h_t^{t,\sigma_{t-1},s_t},\\
P_t^{t,\sigma_t}&=\frac{1}{n^{\sigma_t}_t}\bigg\{\delta_{\begin{subarray}{l} c_t<d_{\max}\\ s_{t-1}=s_t \\ c_{t-1}=c_t+1\end{subarray}}\hskip-0.3cm+\delta_{\begin{subarray}{l} c_t\geq d_{\min} \\ c_{t-1}=1 \end{subarray}}\rho_{\sigma_t}\hskip-0.0cm\sum\limits_{s_{t-1}}\hskip-0.0cm\pi_{s_ts_{t-1}}\bigg\}e^{\sigma_{t-1},s_t}_t\alpha^{\sigma_{t-1}}_{t-1}\\
&\times \big(P_t^{t,\sigma_{t-1},s_t}+\hat h_t^{t,\sigma_{t-1},s_t}(\hat h_t^{t,\sigma_{t-1},s_t})\trans\big)-\hat h_t^{t,\sigma_t}(\hat h_t^{t,\sigma_t})\trans,
\end{align*}
reduces the cost to ${\cal O}(TS^2d_{\max})$.

In similar models in which $h_{1:T}$ are discrete, the recursion for $\hat\alpha^{\sigma_t}_t$ has cost ${\cal O}(TS^2d_{\max})$.

\paragraph{Across-segment independence.}
If across-segment independence is enforced, the recursion for $\alpha^{\sigma_t}_t$ becomes
\begin{align*}
\hskip-0.3cm\alpha^{\sigma_t}_t&\propto\delta_{\begin{subarray}{l} c_t<d_{\max}\end{subarray}}
e^{s_t,c_t+1,s_t}_t\alpha^{s_t,c_t+1}_{t-1}\nonumber\\
\hskip-0.3cm &+\delta_{\begin{subarray}{l} c_t\geq d_{\min} \end{subarray}}
\rho_{\sigma_t}\sum_{s_{t-1}}p(v_t|{\myredd \cancel{s_{t-1}},c_{t-1}\!=\!1},s_t,{\myredd\cancel{v_{1:t-1}}})\pi_{s_ts_{t-1}}\alpha^{s_{t-1},1}_{t-1},
\end{align*}
with $p(v_t|c_{t-1}=1,s_t)={\cal N}(v_t;B^{s_t}\mu^{s_t},B^{s_t}\Sigma^{s_t}(B^{s_t})\trans+\Sigma^{s_t}_V)$. This recursion has cost ${\cal O}(TS^2d_{\max})$.

The recursion for $\hat\alpha^{\sigma_t}_t$ becomes
\begin{align}
\hskip-0.2cm \hat\alpha^{\sigma_t}_t&=\delta_{\begin{subarray}{l} c_t<d_{\max} \\ s_{t-1}=s_t \\ c_{t-1}=c_t+1 \end{subarray}}
p(h_t|\sigma_{t-1},s_t,v_{1:t})p(\sigma_{t-1}|\sigma_t,v_{1:t})\nonumber\\
\hskip-0.2cm &\hskip-0.0cm +\delta_{\begin{subarray}{l} c_t\geq d_{\min} \\ c_{t-1}=1 \end{subarray}}
\cancel{\sum_{s_{t-1}}}p(h_t|{\myredd\cancel{s_{t-1}},c_{t-1}},s_t,{\myredd\cancel{v_{1:t-1}}},v_t)p(\cancel{s_{t-1}},c_{t-1}|\sigma_t,v_{1:t}),\hskip-0.1cm
\label{eq:filtcat}
\end{align}
where $p(h_t|c_{t-1}=1,s_t,v_t)$ is Gaussian with mean and covariance as in \eqref{eq:predcorri}.
Notice that $p(s_{t-1},c_{t-1}=1|\sigma_t,v_{1:t})\neq p(s_{t-1},c_{t-1}=1|\sigma_t,v_{1:t-1})$ as the path $c_{t-1},h_t,v_t$ in \figref{fig:SLGSSMInc}(b) is not blocked.

As $\hat\alpha^{s_t,d_{\max}}_t$ is Gaussian, we deduce that $\hat\alpha^{s_t,d_{\max}-1}_t$ is a Gaussian mixture with 2 components and, more generally,
that $\hat\alpha^{\sigma_t}_t$ is a Gaussian mixture with $d_{\max}-c_t+1$ components, where each component corresponds to a different possible start of the segment.
Therefore, the recursion for $\hat\alpha^{\sigma_t}_t$ has cost ${\cal O}(TSd_{\max}^2)$. Gaussian collapsing is not necessarily required, but can be used to reduce the cost to ${\cal O}(TSd_{\max})$.

In similar models in which $h_{1:T}$ are discrete, the recursion for $\hat\alpha^{\sigma_t}_t$ has cost ${\cal O}(TSd_{\max})$.

\subsubsection*{Smoothing}
As for filtering, we compute the smoothed distribution $p(h_t,\sigma_t|v_{1:T})$ with separate recursions for $\gamma^{\sigma_t}_t=p(\sigma_t|v_{1:T})$ and $\hat\gamma^{\sigma_t}_t=p(h_t|\sigma_t,v_{1:T})$.

\noindent The recursion for $\gamma^{\sigma_t}_t$ is given by
\begin{align*}
\gamma^{\sigma_t}_t&=\sum_{\sigma_{t+1}}p(\sigma_t|\sigma_{t+1},v_{1:T})p(\sigma_{t+1}|v_{1:T})\nonumber\\
&=\sum_{\sigma_{t+1}}\gamma^{\sigma_{t+1}}_{t+1}\int_{h_{t+1}}p(\sigma_t|h_{t+1},\sigma_{t+1},v_{1:t},\cancel{v_{t+1:T}})\hat\gamma^{\sigma_{t+1}}_{t+1},
\end{align*}
where the integral over $h_{t+1}$ cannot be estimated in closed form. If we assume $\hat\gamma^{\sigma_{t+1}}_{t+1}$ to be Gaussian with mean $\hat h_{t+1}^{T,\sigma_{t+1}}$ and covariance $P_{t+1}^{T,\sigma_{t+1}}$,
on the line of EC \citep{barber06ec}, we can approximate $p(\sigma_t|\sigma_{t+1},v_{1:T})$ as
\begin{align*}
\hskip-0.0cm p(\sigma_t|&\sigma_{t+1},v_{1:T})\approx p(\sigma_t|h_{t+1}=\hat h_{t+1}^{T,\sigma_{t+1}},\sigma_{t+1},v_{1:t})\\
\hskip-0.0cm&=\frac{p(h_{t+1}\!=\!\hat h_{t+1}^{T,\sigma_{t+1}}|\sigma_t,s_{t+1},\cancel{c_{t+1}},v_{1:t})p(\sigma_{t+1}|\sigma_t,\cancel{v_{1:t}})p(\sigma_t|v_{1:t})}{\sum_{\tilde\sigma_t}p(h_{t+1}\!=\!\hat h_{t+1}^{T,\sigma_{t+1}}|\tilde\sigma_t,s_{t+1},\cancel{c_{t+1}}|v_{1:t})p(\sigma_{t+1}|\tilde\sigma_t,\cancel{v_{1:t}})p(\tilde\sigma_t|v_{1:t})}\nonumber\\
\hskip-0.0cm&\propto\bigg\{\delta_{\begin{subarray}{l} c_t>1 \\ s_{t+1}=s_t \\ c_{t+1}=c_t-1\end{subarray}}
\hskip-0.5cm+\delta_{c_t=1}\hskip-0.0cm
\rho_{\sigma_{t+1}}\pi_{s_{t+1}s_t}\bigg\}\alpha^{\sigma_t}_tp(h_{t+1}\!=\!\hat h_{t+1}^{T,\sigma_{t+1}}|\sigma_t,s_{t+1},v_{1:t}).\nonumber
\end{align*}
Therefore, the recursion for $\gamma^{\sigma_t}_t$ has cost ${\cal O}(TS^2d_{\max})$.

The recursion for $\hat\gamma^{\sigma_t}_t$ is given by
\begin{align}
\label{eq:smoothM1}
\hat\gamma^{\sigma_t}_t&=\sum_{\sigma_{t+1}}p(h_t|\sigma_{t:t+1},v_{1:T})p(\sigma_{t+1}|\sigma_t,v_{1:T})\\
&=\bigg\{\delta_{\begin{subarray}{l} c_t>1 \\ s_{t+1}=s_t \\ c_{t+1}=c_t-1 \end{subarray}}
\hskip-0.6cm+\delta_{\begin{subarray}{l} c_t=1 \end{subarray}}
\sum_{\sigma_{t+1}}\frac{p(\sigma_t|\sigma_{t+1},v_{1:T})\gamma^{\sigma_{t+1}}_{t+1}}{\sum\limits_{\tilde \sigma_{t+1}}p(\sigma_t|\tilde \sigma_{t+1},v_{1:T})\gamma^{\tilde\sigma_{t+1}}_{t+1}}\bigg\}p(h_t|\sigma_{t:t+1},v_{1:T}),\nonumber
\end{align}
where we have used $p(\sigma_{t+1}=(s_t,c_t-1)|s_t,c_t>1)=1$.
Notice that $h_t \cancel{\ci} c_{t+1}\,|\, \{\sigma_t,s_{t+1},v_{1:T}\}$ as the path $c_{t+1},s_{t+2},v_{t+2},h_{t+2},h_{t+1},h_t$ in \figref{fig:SLGSSMDec}(a) is not blocked (similarly, $h_t \cancel{\ci} s_{t+1}\,|\, \{\sigma_t,c_{t+1},v_{1:T}\}$).
On the line of EC \citep{barber06ec}, $p(h_t|\sigma_{t:t+1},v_{1:T})$ is approximated as
\begin{align}
\hskip-0.0cm p(h_t|\sigma_{t:t+1},v_{1:T})&\hskip-0.0cm=\hskip-0.1cm\int_{h_{t+1}}\hskip-0.0cm p(h_t|h_{t+1},\sigma_t,s_{t+1},\cancel{c_{t+1}},v_{1:t},\cancel{v_{t+1:T}})\nonumber\\
&\times p(h_{t+1}|\sigma_{t:t+1},v_{1:T})\nonumber\\
&\approx \int_{h_{t+1}}\hskip-0.2cm p(h_t|h_{t+1},\sigma_t,s_{t+1},v_{1:t})\hat\gamma^{\sigma_{t+1}}_{t+1}.
\label{eq:approx1}
\end{align}
Notice that $h_t \cancel{\ci} s_{t+1}\,|\, \{h_{t+1},\sigma_t,c_{t+1},v_{1:T}\}$, as the path $s_{t+1},h_{t+1},h_t$ in \figref{fig:SLGSSMDec}(a) is not blocked.

Assuming Gaussian collapsing of $\hat\alpha^{\sigma_t}_t$, from \eqref{eq:lgssmh} we deduce
that $p(h_{t:t+1}|\sigma_t,s_{t+1},v_{1:t})$ has covariance
\begin{displaymath}
\left[ \begin{array}{cc}
P_t^{t,\sigma_t} &  P_t^{t,\sigma_t}(A^{s_{t+1}})\trans \\
A^{s_{t+1}}P_t^{t,\sigma_t} & A^{s_{t+1}}P_t^{t,\sigma_t}(A^{s_{t+1}})\trans+\Sigma^{s_{t+1}}_H\\
\end{array} \right].
\end{displaymath}
By using the formula of Gaussian conditioning we deduce the $p(h_t|h_{t+1},\sigma_t,s_{t+1},v_{1:t})$ is Gaussian with mean and covariance given by
\begin{align*}
\hat h_t^{t,\sigma_t}+\hat A^{\sigma_t,s_{t+1}}_t(h_{t+1}-A^{s_{t+1}}\hat h_t^{t,\sigma_t}), \hskip0.3cm
P_t^{t,\sigma_t}-\hat A^{\sigma_t,s_{t+1}}_tA^{s_{t+1}}P_t^{t,\sigma_t},
\end{align*}
where $\hat A^{\sigma_t,s_{t+1}}_t=P_t^{t,\sigma_t}(A^{s_{t+1}})\trans(A^{s_{t+1}}P_t^{t,\sigma_t}(A^{s_{t+1}})\trans+\Sigma^{s_{t+1}}_H)^{-1}$.
This can be equivalently expressed by the linear system of reverse dynamics
\begin{align*}
h_t=\hat A^{\sigma_t,s_{t+1}}_th_{t+1}+\hat m^{\sigma_t,s_{t+1}}_t+\hat \eta_t,
\end{align*}
where
$m^{\sigma_t,s_{t+1}}_t\!=\!\hat h^{t,\sigma_t}_t\!-\!\hat A^{\sigma_t,s_{t+1}}_tA^{s_{t+1}}\hat h^{t,\sigma_t}_t$ and
$p(\hat\eta_t|\sigma_t,s_{t+1},v_{1:t})\!=\!{\cal N}(0,P_t^{t,\sigma_t}\!-\!\hat A^{\sigma_t,s_{t+1}}_tA^{s_{t+1}}P_t^{t,\sigma_t})$.

As $p(h_{t+1},\hat\eta_t|\sigma_{t:t+1},v_{1:T})=p(\hat\eta_t|\sigma_t,s_{t+1},v_{1:t})p(h_{t+1}|\sigma_{t:t+1},v_{1:T})$, we deduce that $p(h_t|\sigma_{t:t+1},v_{1:T})$ is Gaussian with mean and covariance
\begin{align}
\hat h_t^{T,\sigma_{t:t+1}}&=\hat A^{\sigma_t,s_{t+1}}_t\hat h_{t+1}^{T,\sigma_{t+1}}+\hat m^{\sigma_t,s_{t+1}}_t\nonumber\\
&=\hat h^{t,\sigma_t}_t+\hat A^{\sigma_t,s_{t+1}}_t(\hat h_{t+1}^{T,\sigma_{t+1}}-A^{s_{t+1}}\hat h^{t,\sigma_t}_t),\nonumber\\
P_t^{T,\sigma_{t:t+1}}&=\hat A^{\sigma_t,s_{t+1}}_tP_{t+1}^{T,\sigma_{t+1}}(\hat A^{\sigma_t,s_{t+1}}_t)\trans+P_t^{t,\sigma_t}\!-\!\hat A^{\sigma_t,s_{t+1}}_tA^{s_{t+1}}P_t^{t,\sigma_t}\nonumber\\
&=P_t^{t,\sigma_t}\!+\!\hat A^{\sigma_t,s_{t+1}}_t(P_{t+1}^{T,\sigma_{t+1}}\!-\!P_{t+1}^{t,\sigma_{t},s_{t+1}})(\hat A^{\sigma_t,s_{t+1}}_t)\trans.
\label{eq:RTS}
\end{align}
Notice that, if we remove dependence on $\sigma_{t:t+1}$, \eqref{eq:RTS} becomes the Rauch-Tung-Striebel routines of the \LGSSM~\citep{rauch65maximum,chiappa06phd}.

Since $\hat\gamma^{\sigma_T}_T=\hat\alpha^{\sigma_T}_T$ is Gaussian, $\hat\gamma^{s_{T-1},c_{T-1}>1}_{T-1}$ is Gaussian, whilst $\hat\gamma^{s_{T-1},1}_{T-1}$ is a Gaussian mixture with $Sd_{\max}$ components.
More generally, $\hat\gamma^{\sigma_t}_t$ has a complex number of components dominated by $S^{T-t}d_{\max}$.
Gaussian collapsing of $\hat\gamma^{s_t,1}_t$
\begin{align*}
\hat h_t^{T,s_t,1}\!=\!&
\hskip-0.1cm\sum_{\sigma_{t+1}}\hat h_t^{T,\sigma_{t:t+1}}p(\sigma_{t+1}|\sigma_t,v_{1:T}),\\
P_t^{T,s_t,1}\!=\!&
\hskip-0.1cm\sum_{\sigma_{t+1}}\hskip-0.1cm\big(P_t^{T,\sigma_{t:t+1}}\!\!+\!\hat h_t^{T,\sigma_{t:t+1}}
(\hat h_t^{T,\sigma_{t:t+1}})\!\trans\big)p(\sigma_{t+1}|\sigma_t,\!v_{1:T})
\!-\!\hat h_t^{T,\sigma_t}(\hat h_t^{T,\sigma_t})\!\trans\!\!,
\end{align*}
reduces the cost of the $\hat\gamma^{\sigma_t}_t$ recursion to ${\cal O}(TS^2d_{\max})$.

In similar models in which $h_{1:T}$ are discrete, the recursion for $\hat\gamma^{\sigma_t}_t$ has cost ${\cal O}(TS^2d_{\max})$.

\paragraph{Across-segments independence.}
The recursion for $\gamma^{\sigma_t}_t$ becomes
\begin{align*}
&\gamma^{\sigma_t}_t=\bigg\{\delta_{\begin{subarray}{l} c_t>1 \\ s_{t+1}=s_t \\ c_{t+1}=c_t-1\end{subarray}}\hskip-0.3cm p(\sigma_t|\sigma_{t+1},v_{1:T})
+\delta_{c_t=1}\sum_{\sigma_{t+1}}p(\sigma_t|\sigma_{t+1},v_{1:t},{\myredd\cancel{v_{t+1:T}}})\bigg\}\gamma^{\sigma_{t+1}}_{t+1}\\
&\!=\bigg\{\delta_{\begin{subarray}{l} c_t>1 \\ s_{t+1}=s_t \\ c_{t+1}=c_t-1\end{subarray}}\hskip-0.4cm (1\!-\!p(\tilde c_t\!=\!1,s_t|\sigma_{t+1},v_{1:t}))
+\delta_{c_t=1}\sum_{\sigma_{t+1}}p(\sigma_t|\sigma_{t+1},v_{1:t})\bigg\}\gamma^{\sigma_{t+1}}_{t+1},
\end{align*}
and therefore the EC approximation $p(\sigma_t|\sigma_{t+1},v_{1:T})\approx p(\sigma_t|h_{t+1}=\hat h_{t+1}^{T,\sigma_{t+1}},\sigma_{t+1},v_{1:t})$ is not required. This recursion has cost ${\cal O}(TS^2d_{\max})$.

The recursion for $\hat\gamma^{\sigma_t}_t$ becomes
\begin{align}
\hat\gamma^{\sigma_t}_t
&=\delta_{\begin{subarray}{l} c_t>1 \\ s_{t+1}=s_t \\ c_{t+1}=c_t-1 \end{subarray}}
p(h_t|\sigma_{t:t+1},v_{1:T})\nonumber\\
&+\delta_{\begin{subarray}{l} c_t=1 \end{subarray}}
\cancel{\sum_{\sigma_{t+1}}}p(h_t|\sigma_t,{\myredd\cancel{\sigma_{t+1}}},v_{1:t},{\myredd\cancel{v_{t+1:T}}})\cancel{p(\sigma_{t+1}|\sigma_t,v_{1:T})}\,.
\label{eq:smoothM1cat}
\end{align}
Notice that the simplification with respect to recursion (\ref{eq:smoothM1}) arises from the combination of across-segment independence and the fact that $c_t$ encodes information about the end of the segment, and therefore about $c_{t+1}$ for $c_t>1$.

If $\hat\alpha^{\sigma_t}_t$ is not collapsed, we can group the components of $\hat\alpha^{\sigma_t}_t$ into 2 groups corresponding to $c_{t-1}=1$ and $c_{t-1}=c_t+1$. For example, $\hat\alpha^{s_t,d_{\max}-2}_t$ is a mixture of 3
components in which two components correspond to $c_{t-1}=d_{\max}-1$ (specifically to $c_{t-1}=d_{\max}-1, c_{t-2}=d_{\max}$ and $c_{t-1}=d_{\max}-1, c_{t-2}=1$), and one component corresponds to $c_{t-1}=1$. 
Consider recursion (\ref{eq:smoothM1cat}) for $c_t>1$. At time-step $T-1$ the EC approximation $p(h_T|\sigma_{T-1:T},v_{1:T})\approx p(h_T|\sigma_{T},v_{1:T})$ in \eqref{eq:approx1} is not needed. 
The derivations following \eqref{eq:approx1} produce a Gaussian mixture $p(h_{T-1}|\sigma_{T-1},v_{1:T})$ with $d_{\max}-c_{T-1}+1$ components (due to $\hat\alpha^{\sigma_{T-1}}_{T-1}$), 
which can be grouped into 2 groups corresponding to $c_{T-2}=1$ and $c_{T-2}=c_{T-1}+1$. At time-step $T-2$, the EC approximation $p(h_{T-1}|\sigma_{T-2:T-1},v_{1:T})\approx p(h_{T-1}|\sigma_{T-1},v_{1:T})$
in \eqref{eq:approx1} is not needed, as we can use the components of $\gamma^{\sigma_{T-1}}_{T-1}$ corresponding to $c_{T-2}=c_{T-1}+1$. More generally, the EC approximation $p(h_{t+1}|\sigma_{t:t+1},v_{1:T})\approx p(h_{t+1}|\sigma_{t+1},v_{1:T})$ 
is not needed and $\hat\gamma^{\sigma_t}_t$ is a Gaussian mixture with $d_{\max}-c_t+1$ components. Therefore, the recursion for $\hat\gamma^{\sigma_t}_t$ has cost ${\cal O}(TSd^2_{\max})$.

With Gaussian collapsing of $\hat\alpha^{\sigma_t}_t$, $\hat\gamma^{\sigma_t}_t$ is Gaussian and the cost reduces to ${\cal O}(TSd_{\max})$. However the EC approximation is required in this case.

In similar models in which $h_{1:T}$ are discrete, the recursion for $\hat\gamma^{\sigma_t}_t$ has cost ${\cal O}(TSd_{\max})$ and the EC approximation is required 
(a grouping approach as the one described above could alternatively be employed, but this would increase the cost to ${\cal O}(TSd^2_{\max})$ in both filtering and smoothing).

\subsection{Increasing Count Variables \label{sec:SLGSSMInc}}
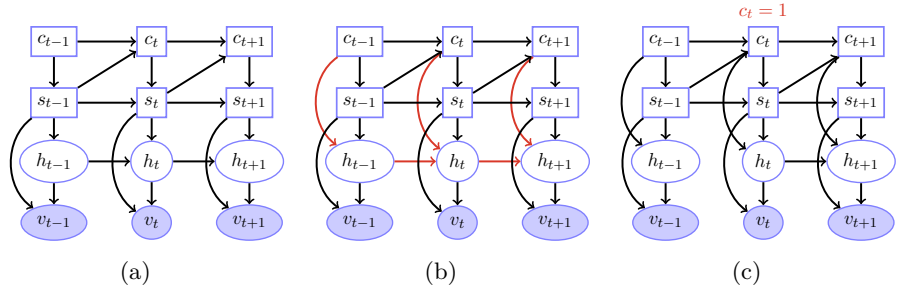
\begin{figure}[t]
\hskip-0.3cm
\subfigure[]{\scalebox{0.65}{
\begin{tikzpicture}[dgraph]
\node[disc] (sigmatm) at (2,0.2) {$c_{t-1}$};
\node[disc] (sigmat) at (4,0.2) {$c_t$};
\node[disc] (sigmatp) at (6,0.2) {$c_{t+1}$};
\node[disc] (stm) at (2,-1.05) {$s_{t-1}$};
\node[disc] (st) at (4,-1.05) {$s_t$};
\node[disc] (stp) at (6,-1.05) {$s_{t+1}$};
\node[ocont] (htm) at (2,-2.25) {$h_{t-1}$};
\node[ocont] (ht) at (4,-2.25) {$h_t$};
\node[ocont] (htp) at (6,-2.25) {$h_{t+1}$};
\node[ocont,obs] (vtm) at (2,-3.5) {$v_{t-1}$};
\node[ocont,obs] (vt) at (4,-3.5) {$v_t$};
\node[ocont,obs] (vtp) at (6,-3.5) {$v_{t+1}$};
\draw[line width=1.15pt](sigmatm)--(sigmat);\draw[line width=1.15pt](sigmat)--(sigmatp);
\draw[line width=1.15pt](stm)--(st);\draw[line width=1.15pt](st)--(stp);
\draw[line width=1.15pt](stm)--(htm);\draw[line width=1.15pt](st)--(ht);\draw[line width=1.15pt](stp)--(htp);
\draw[line width=1.15pt](sigmatm)--(stm);\draw[line width=1.15pt](sigmat)--(st);\draw[line width=1.15pt](sigmatp)--(stp);
\draw[line width=1.15pt](htm)--(vtm);\draw[line width=1.15pt](ht)--(vt);\draw[line width=1.15pt](htp)--(vtp);
\draw[line width=1.15pt](htm)--(ht);\draw[line width=1.15pt](ht)--(htp);
\draw[line width=1.15pt](stm)to [bend right=55](vtm);\draw[line width=1.15pt](st)to [bend right=55](vt);\draw[line width=1.15pt](stp)to [bend right=55](vtp);
\draw[line width=1.15pt](stm)--(sigmat);\draw[line width=1.15pt](st)--(sigmatp);
\end{tikzpicture}}}
\hskip0.1cm
\subfigure[]{\scalebox{0.65}{
\begin{tikzpicture}[dgraph]
\node[disc] (sigmatm) at (2,0.2) {$c_{t-1}$};
\node[disc] (sigmat) at (4,0.2) {$c_t$};
\node[disc] (sigmatp) at (6,0.2) {$c_{t+1}$};
\node[disc] (stm) at (2,-1.05) {$s_{t-1}$};
\node[disc] (st) at (4,-1.05) {$s_t$};
\node[disc] (stp) at (6,-1.05) {$s_{t+1}$};
\node[ocont] (htm) at (2,-2.25) {$h_{t-1}$};
\node[ocont] (ht) at (4,-2.25) {$h_t$};
\node[ocont] (htp) at (6,-2.25) {$h_{t+1}$};
\node[ocont,obs] (vtm) at (2,-3.5) {$v_{t-1}$};
\node[ocont,obs] (vt) at (4,-3.5) {$v_t$};
\node[ocont,obs] (vtp) at (6,-3.5) {$v_{t+1}$};
\draw[color=myred,line width=1.15pt](sigmatm)to [bend right=55](htm);\draw[color=myred,line width=1.15pt](sigmat)to [bend right=55](ht);\draw[color=myred,line width=1.15pt](sigmatp)to [bend right=55](htp);
\draw[line width=1.15pt](sigmatm)--(sigmat);\draw[line width=1.15pt](sigmat)--(sigmatp);
\draw[line width=1.15pt](stm)--(st);\draw[line width=1.15pt](st)--(stp);
\draw[line width=1.15pt](stm)--(htm);\draw[line width=1.15pt](st)--(ht);\draw[line width=1.15pt](stp)--(htp);
\draw[line width=1.15pt](sigmatm)--(stm);\draw[line width=1.15pt](sigmat)--(st);\draw[line width=1.15pt](sigmatp)--(stp);
\draw[line width=1.15pt](htm)--(vtm);\draw[line width=1.15pt](ht)--(vt);\draw[line width=1.15pt](htp)--(vtp);
\draw[color=myred,line width=1.15pt](htm)--(ht);\draw[color=myred,line width=1.15pt](ht)--(htp);
\draw[line width=1.15pt](stm)to [bend right=55](vtm);\draw[line width=1.15pt](st)to [bend right=55](vt);\draw[line width=1.15pt](stp)to [bend right=55](vtp);
\draw[line width=1.15pt](stm)--(sigmat);\draw[line width=1.15pt](st)--(sigmatp);
\end{tikzpicture}}}
\hskip0.1cm
\subfigure[]{\scalebox{0.65}{
\begin{tikzpicture}[dgraph]
\node[disc] (sigmatm) at (2,0.2) {$c_{t-1}$};
\node[disc] (sigmat) at (4,0.2) {$c_t$};
\node[disc] (sigmatp) at (6,0.2) {$c_{t+1}$};
\node[disc] (stm) at (2,-1.05) {$s_{t-1}$};
\node[disc] (st) at (4,-1.05) {$s_t$};
\node[disc] (stp) at (6,-1.05) {$s_{t+1}$};
\node[ocont] (htm) at (2,-2.25) {$h_{t-1}$};
\node[ocont] (ht) at (4,-2.25) {$h_t$};
\node[ocont] (htp) at (6,-2.25) {$h_{t+1}$};
\node[ocont,obs] (vtm) at (2,-3.5) {$v_{t-1}$};
\node[ocont,obs] (vt) at (4,-3.5) {$v_t$};
\node[ocont,obs] (vtp) at (6,-3.5) {$v_{t+1}$};
\node [color=myred,above] at (sigmat.north) {$c_{t}=1$};
\draw[line width=1.15pt](sigmatm)to [bend right=55](htm);\draw[line width=1.15pt](sigmat)to [bend right=55](ht);\draw[line width=1.15pt](sigmatp)to [bend right=55](htp);
\draw[line width=1.15pt](sigmatm)--(sigmat);\draw[line width=1.15pt](sigmat)--(sigmatp);
\draw[line width=1.15pt](stm)--(st);\draw[line width=1.15pt](st)--(stp);
\draw[line width=1.15pt](stm)--(htm);\draw[line width=1.15pt](st)--(ht);\draw[line width=1.15pt](stp)--(htp);
\draw[line width=1.15pt](sigmatm)--(stm);\draw[line width=1.15pt](sigmat)--(st);\draw[line width=1.15pt](sigmatp)--(stp);
\draw[line width=1.15pt](htm)--(vtm);\draw[line width=1.15pt](ht)--(vt);\draw[line width=1.15pt](htp)--(vtp);
\draw[line width=1.15pt](ht)--(htp);
\draw[line width=1.15pt](stm)to [bend right=55](vtm);\draw[line width=1.15pt](st)to [bend right=55](vt);\draw[line width=1.15pt](stp)to [bend right=55](vtp);
\draw[line width=1.15pt](stm)--(sigmat);\draw[line width=1.15pt](st)--(sigmatp);
\end{tikzpicture}}}
\caption{(a): Explicit-duration \SLGSSM~using increasing count variables.
(b): Across-segment independence is enforced with a link from $c_t$ to $h_t$,
as explicitly represented in (c).}
\label{fig:SLGSSMInc}
\end{figure}
The explicit-duration \SLGSSM~using increasing count variables has belief network representation given in \figref{fig:SLGSSMInc}(a).
Across-segment independence can be enforced by adding a link from $c_t$ to $h_t$ as in \figref{fig:SLGSSMInc}(b),
which has the effect of removing the link from $h_{t-1}$ to $h_t$ if $c_t=1$, as explicitly represented in \figref{fig:SLGSSMDec}(c).
More specifically, dependence cut is defined as
\begin{align*}
p(h_t|h_{t-1},\sigma_t)&=\begin{cases}
    p(h_t|\sigma_t)={\cal N}(h_t;\mu^{s_t},\Sigma^{s_t}) & \hspace{-0.0cm} \textrm{if } c_t\!=\!1\\
	{\cal N}(h_t;A^{s_t}h_{t-1},\Sigma^{s_t}_H) 	& \hspace{-0.0cm} \textrm{if } c_t\!>\!1.
\end{cases}
\end{align*}

\subsubsection*{Filtering}
The recursion for $\alpha^{\sigma_t}_t=p(\sigma_t|v_{1:t})$ is given by\footnote{Notice the similarity with recursion (\ref{eq:alphaM1alt}).}
\begin{align*}
\hskip-0.1cm\alpha^{\sigma_t}_t&
=\frac{\sum_{\sigma_{t-1}}p(\sigma_{t-1:t},v_t|v_{1:t-1})}{\sum_{\tilde\sigma_{t-1:t}}p(\tilde\sigma_{t-1:t},v_t|v_{1:t-1})}\nonumber\\
&\propto\sum_{\sigma_{t-1}} p(v_t|\sigma_{t-1},s_t,\cancel{c_t},v_{1:t-1})p(\sigma_t|\sigma_{t-1},\cancel{v_{1:t-1}})p(\sigma_{t-1}|v_{1:t-1})\nonumber\\
&=\bigg\{\delta_{\begin{subarray}{l} c_t>1 \\ s_{t-1}\\ c_{t-1}=c_t-1 \end{subarray}}
\hskip-0.3cm\lambda_{\sigma_{t-1}}\hskip-0.1cm
+\delta_{\begin{subarray}{l} c_t=1 \end{subarray}}\hskip-0.0cm
\sum_{s_{t-1}}\hskip-0.0cm\pi_{s_ts_{t-1}}\hskip-0.1cm\sum_{c_{t-1}}\hskip-0.01cm(1\!-\!\lambda_{\sigma_{t-1}})\bigg\}e^{\sigma_{t-1},s_t}_t\alpha^{\sigma_{t-1}}_{t-1},
\end{align*}
where $e^{\sigma_{t-1},s_t}_t=p(v_t|\sigma_{t-1},s_t,v_{1:t-1})$. This recursion has computational cost ${\cal O}(TS^2d_{\max})$.

The recursion for $\hat\alpha^{\sigma_t}_t=p(h_t|\sigma_t,v_{1:t})$ is given by
\begin{align}
\label{eq:filtM1alt}
&\hat\alpha^{\sigma_t}_t=\sum_{\sigma_{t-1}}p(h_t|\sigma_{t-1},s_t,\cancel{c_t},v_{1:t})p(\sigma_{t-1}|\sigma_t,v_{1:t})\\
&=\bigg\{\delta_{\begin{subarray}{l} c_t>1\\ s_{t-1}=s_t\\c_{t-1}=c_t-1  \end{subarray}}\hskip-0.4cm+
\delta_{\begin{subarray}{l} c_t=1\end{subarray}}\hskip-0.0cm\sum_{\sigma_{t-1}}\hskip-0.0cm\frac{p(\sigma_{t-1:t},v_t|v_{1:t-1})}{\sum_{\tilde\sigma_{t-1}} p(\tilde\sigma_{t-1},\sigma_t,v_t|v_{1:t-1})}\bigg\}p(h_t|\sigma_{t-1},s_t,v_{1:t})\nonumber,
\end{align}
where we have used $p(\sigma_{t-1}=(s_t,c_t-1)|s_t,c_t>1,v_{1:t})=1$.
Therefore, since $\hat\alpha^{\sigma_1}_1$ is Gaussian, $\hat\alpha^{s_2,c_2>1}_2$ is Gaussian and $\hat\alpha^{s_2,1}_2$ is a Gaussian mixture with $Sd_{\max}$ components. In general, $\hat\alpha^{\sigma_t}_t$ has a complex number of components dominated by $S^{t-1}d_{\max}$.
Gaussian collapsing of $\hat\alpha^{s_t,1}_t$
\begin{align*}
\hat h_t^{t,s_t,1}
&\!\!=\!\hskip-0.1cm\sum_{\sigma_{t-1}}\hskip-0.0cm \!p(\sigma_{t-1}|\sigma_t,v_{1:t}) \hat h_t^{t,\sigma_{t-1},s_t},\\
P_t^{t,s_t,1}&\!\!=\!\hskip-0.1cm\sum_{\sigma_{t-1}}\hskip-0.0cm\! p(\sigma_{t-1}|\sigma_t,v_{1:t})(P_t^{t,\sigma_{t-1},s_t}\!\!+\!\hat h_t^{t,\sigma_{t-1},s_t}(\hat h_t^{t,\sigma_{t-1},s_t})\!\trans)
\!-\!\hat h_t^{t,\sigma_t}(\hat h_t^{t,\sigma_t})\!\trans\!\!,
\end{align*}
reduces the cost of the recursion for $\hat\alpha^{\sigma_t}_t$ to ${\cal O}(TS^2d_{\max})$.

In similar models in which $h_{1:T}$ are discrete, the recursion for $\hat\alpha^{\sigma_t}_t$ has cost ${\cal O}(TS^2d_{\max})$.

\paragraph{Across-segment independence.}
The recursion for $\alpha^{\sigma_t}_t$ becomes
\begin{align*}
\alpha^{\sigma_t}_t
&\propto\sum_{\sigma_{t-1}} p(v_t|\sigma_{t-1},s_t,{\myredd c_t},v_{1:t-1})p(\sigma_t|\sigma_{t-1},\cancel{v_{1:t-1}})p(\sigma_{t-1}|v_{1:t-1})\\
&=\delta_{\begin{subarray}{l} c_t>1 \\ s_{t-1}=s_t \\ c_{t-1}=c_t-1 \end{subarray}}
\lambda_{\sigma_{t-1}}p(v_t|\sigma_{t-1:t},v_{1:t-1})\alpha^{\sigma_{t-1}}_{t-1}\\
&+\delta_{c_t=1}\sum_{s_{t-1}}\pi_{s_ts_{t-1}}\sum_{c_{t-1}}(1\!-\!\lambda_{\sigma_{t-1}})
p(v_t|{\myredd\cancel{\sigma_{t-1}}},\sigma_t,{\myredd\cancel{v_{1:t-1}}})\alpha^{\sigma_{t-1}}_{t-1},
\end{align*}
with $p(v_t|s_t,c_t=1)={\cal N}(v_t;B^{s_t}\mu^{s_t},B^{s_t}\Sigma^{s_t}(B^{s_t})\trans+\Sigma^{s_t}_V)$, and with $p(v_t|\sigma_{t-1}\!=\!(s_t,c_t-1),s_t,c_t>1,v_{1:t-1})$ estimated as in the case of across-segment dependence. This recursion has cost ${\cal O}(TS^2d_{\max})$.

The recursion for $\hat\alpha^{\sigma_t}_t$ becomes
\begin{align}
\hat\alpha^{\sigma_t}_t&=\sum_{\sigma_{t-1}}p(h_t|\sigma_{t-1},s_t,{\myredd c_t},v_{1:t})p(\sigma_{t-1}|\sigma_t,v_{1:t})\nonumber\\
&=\delta_{\begin{subarray}{l} c_t>1 \\ s_{t-1}=s_t \\ c_{t-1}=c_t-1 \end{subarray}}
p(h_t|\sigma_{t-1:t},v_{1:t})\nonumber\\
&+\delta_{\begin{subarray}{l} c_t=1 \end{subarray}}
\cancel{\sum_{\sigma_{t-1}}}p(h_t|{\myredd\cancel{\sigma_{t-1}}},\sigma_t,{\myredd\cancel{v_{1:t-1}}},v_t)\cancel{p(\sigma_{t-1}|\sigma_t,v_{1:t})},
\label{eq:filtM1altcat}
\end{align}
where $p(h_t|s_t,c_t=1,v_t)$ is Gaussian with mean and covariance as in \eqref{eq:predcorri}, and where $p(h_t|\sigma_{t-1}=(s_t,c_t-1),s_t,c_t>1,v_{1:t})$ can estimated using the recursions (\ref{eq:pred}) and (\ref{eq:corr}) with different indexes.
Therefore $\hat\alpha^{\sigma_t}_t$ is Gaussian and the recursion has cost ${\cal O}(TSd_{\max})$.
The recursion essentially performs filtering in a \LGSSM~on $v_{t:t+d_{\max}-1}$ for all $t$ and $s_t$.

Notice that the simplification with respect to recursion (\ref{eq:filtM1alt}) arises from the combination of across-segment independence and the fact that $c_t$ encodes information about the start of the segment, and therefore about $c_{t-1}$ for $c_t>1$.

In similar models in which $h_{1:T}$ are discrete, the recursion for $\hat\alpha^{\sigma_t}_t$ has cost ${\cal O}(TSd_{\max})$.

\subsubsection*{Smoothing\label{sec:SLGSSMIncSmooth}}
The recursion for $\gamma^{\sigma_t}_t$ is given by
\begin{align*}
&\gamma^{\sigma_t}_t\!=\!\sum_{\sigma_{t+1}}p(\sigma_t|\sigma_{t+1},v_{1:T})p(\sigma_{t+1}|v_{1:T})\nonumber\\
&\!=\!\delta_{\begin{subarray}{l} c_t< d_{\max} \end{subarray}}\gamma^{s_t,c_t+1}_{t+1}
+\delta_{\begin{subarray}{l} c_t\geq d_{\min} \\ c_{t+1}=1 \end{subarray}}
\sum_{s_{t+1}}\underbrace{p(\sigma_t|\sigma_{t+1},v_{1:T})}_{\approx p(\sigma_t|h_{t+1}=\hat h_{t+1}^{T,\sigma_{t+1}},\sigma_{t+1},v_{1:t})}\gamma^{\sigma_{t+1}}_{t+1}\\
&\!=\!\delta_{\begin{subarray}{l} c_t<d_{\max} \end{subarray}}\gamma^{s_t,c_t+1}_{t+1}\nonumber\\
&\!+\!\delta_{\begin{subarray}{l}  c_t\geq d_{\min} \\ c_{t+1}=1 \end{subarray}}\hskip-0.0cm\!\sum\limits_{s_{t+1}}\!\frac{\pi_{s_{t+1}s_t}(1\!-\!\lambda_{\sigma_t})\alpha^{\sigma_t}_t p(h_{t+1}\!=\!\hat h_{t+1}^{T,\sigma_{t+1}}|\sigma_t,s_{t+1},v_{1:t})}
{\sum_{\tilde\sigma_t}\pi_{s_{t+1}\tilde s_t}(1\!-\!\lambda_{\tilde \sigma_t})\alpha^{\tilde\sigma_t}_tp(h_{t+1}\!=\!\hat h_{t+1}^{T,\sigma_{t+1}}|\tilde\sigma_t,s_{t+1},v_{1:t})}\gamma^{\sigma_{t+1}}_{t+1}\hskip-0.02cm,
\end{align*}
where we have used $p(\sigma_t=(s_{t+1},c_{t+1}-1)|s_{t+1},c_{t+1}>1,v_{1:T})=1$.
This recursion has cost ${\cal O}(TS^2d_{\max})$.

The recursion for $\hat\gamma^{\sigma_t}_t=p(h_t|\sigma_t,v_{1:T})$ is given by
\begin{align}
\hat\gamma^{\sigma_t}_t&=\sum_{\sigma_{t+1}}p(h_t|\sigma_{t:t+1},v_{1:T})p(\sigma_{t+1}|\sigma_t,v_{1:T}),
\label{eq:smoothM1Alt}
\end{align}
where
\begin{align*}
p(\sigma_{t+1}|\sigma_t,v_{1:T})&=\frac{p(\sigma_t|\sigma_{t+1},v_{1:T})p(\sigma_{t+1}|v_{1:T})}
{\sum_{\tilde\sigma_{t+1}}p(\sigma_t|\tilde\sigma_{t+1},v_{1:T})p(\tilde\sigma_{t+1}|v_{1:T})}\nonumber\\
&\propto\bigg\{\delta_{\begin{subarray}{l} c_t< d_{\max} \\ s_{t+1}=s_t \\ c_{t+1}=c_t+1 \end{subarray}}
+\delta_{\begin{subarray}{l} c_t\geq d_{\min}  \\ c_{t+1}=1 \end{subarray}}
p(\sigma_t|\sigma_{t+1},v_{1:T})\bigg\}\gamma^{\sigma_{t+1}}_{t+1},
\end{align*}
and where $p(h_t|\sigma_{t:t+1},v_{1:T})$ is computed as in \eqref{eq:RTS}. However notice that, for $c_{t+1}>1$,
$p(h_{t+1}|\sigma_t,\sigma_{t+1}=(s_t,c_t+1),v_{1:T})=p(h_{t+1}|\sigma_{t+1}=(s_t,c_t+1),v_{1:T})$, and therefore the EC approximation
$p(h_{t+1}|\sigma_{t:t+1},v_{1:T})\approx \hat\gamma^{\sigma_{t+1}}_{t+1}$ in \eqref{eq:approx1} becomes exact. Indeed for $c_{t+1}>1$
\begin{align*}
p(h_{t+1}|\sigma_{t+1},v_{1:T})
&=\sum_{\sigma_t}p(h_{t+1}|\sigma_{t:t+1},v_{1:T})p(\sigma_t|\sigma_{t+1},v_{1:T})\\
&=p(h_{t+1}|\sigma_t\!=\!(s_{t+1},\!c_{t+1}\!-\!1),\sigma_{t+1},v_{1:T}).
\end{align*}
With Gaussian collapsing of $\hat\alpha^{\sigma_t}_t$, $\hat\gamma^{\sigma_t}_t$ is a Gaussian mixture with $S^{T-t+1}$ components.
Gaussian collapsing reduces the cost to ${\cal O}(TS^2d_{\max})$.

In similar models in which $h_{1:T}$ are discrete, the recursion for $\hat\gamma^{\sigma_t}_t$ has cost ${\cal O}(TS^2d_{\max})$.

\paragraph{Across-segments independence.}
The recursion for $\gamma^{\sigma_t}_t$ becomes
\begin{align*}
\gamma^{\sigma_t}_t&=\delta_{\begin{subarray}{l} c_t< d_{\max} \end{subarray}}\gamma^{s_t,c_t+1}_{t+1}
+\delta_{\begin{subarray}{l} c_t\geq d_{\min} \\ c_{t+1}=1 \end{subarray}}
\sum_{s_{t+1}}p(\sigma_t|\sigma_{t+1},v_{1:t},{\myredd\cancel{v_{t+1:T}}})\gamma^{\sigma_{t+1}}_{t+1},
\end{align*}
and therefore the EC approximation $p(\sigma_t|\sigma_{t+1},v_{1:T})\approx p(\sigma_t|h_{t+1}=\hat h_{t+1}^{T,\sigma_{t+1}},\sigma_{t+1},v_{1:t})$ is not required. This recursion has cost ${\cal O}(TS^2d_{\max})$.

The recursion for $\hat\gamma^{\sigma_t}_t$ becomes
\begin{align}
\hat\gamma^{\sigma_t}_t&=
\delta_{\begin{subarray}{l} c_t< d_{\max} \\ s_{t+1}=s_t \\ c_{t+1}=c_t+1 \end{subarray}}
p(h_t|\sigma_{t:t+1},v_{1:T})p(\sigma_{t+1}|\sigma_t,v_{1:T})\nonumber\\
&+\delta_{\begin{subarray}{l} c_t\geq d_{\min} \\ c_{t+1}=1 \end{subarray}}\cancel{\sum_{s_{t+1}}}
p(h_t|\sigma_t,{\myredd\cancel{\sigma_{t+1}}},v_{1:t},{\myredd\cancel{v_{t+1:T}}})p(\cancel{s_{t+1}},c_{t+1}|\sigma_t,v_{1:T}),
\label{eq:smoothM1altcat}
\end{align}
and therefore the EC approximation $p(h_{t+1}|\sigma_{t:t+1},v_{1:T})\approx \hat\gamma^{\sigma_{t+1}}_{t+1}$ in \eqref{eq:approx1} is not required.
Since $p(h_t|\sigma_t,v_{1:t})$ is Gaussian,
$p(h_t|\sigma_t,v_{1:T})$ is a Gaussian mixture with $d_{\max}-c_t+1$ components and therefore the recursion has cost ${\cal O}(TSd^2_{\max})$.
Gaussian collapsing is not necessarily required, but can be used to reduce the cost to ${\cal O}(TSd_{\max})$.

Notice that is the combination of across-segment independence and the fact that $c_t$ encodes
information about the start of the segment, and therefore about $c_t-1$ for $c_t>1$, that eliminates the need of the EC approximations.

In similar models in which $h_{1:T}$ are discrete, the recursion for $\hat\gamma^{\sigma_t}_t$ has cost ${\cal O}(TSd_{\max})$.

\subsection{Count-Duration Variables\label{sec:SLGSSMCD}}
\begin{figure}[t]
\hskip-0.3cm
\subfigure[]{
\scalebox{0.63}{
\begin{tikzpicture}
\node[disc] (sigmatm) at (2,1.25) {$c_{t-1}$};
\node[disc] (sigmat) at (4,1.25) {$c_t$};
\node[disc] (sigmatp) at (6,1.25) {$c_{t+1}$};
\node[disc] (dtm) at (2,0) {$d_{t-1}$};
\node[disc] (dt) at (4,0) {$d_t$};
\node[disc] (dtp) at (6,0) {$d_{t+1}$};
\node[disc] (stm) at (2,-1.25) {$s_{t-1}$};
\node[disc] (st) at (4,-1.25) {$s_t$};
\node[disc] (stp) at (6,-1.25) {$s_{t+1}$};
\node[ocont] (htm) at (2,-2.5) {$h_{t-1}$};
\node[ocont] (ht) at (4,-2.5) {$h_t$};
\node[ocont] (htp) at (6,-2.5) {$h_{t+1}$};
\node[ocont,obs] (vtm) at (2,-3.75) {$v_{t-1}$};
\node[ocont,obs] (vt) at (4,-3.75) {$v_t$};
\node[ocont,obs] (vtp) at (6,-3.75) {$v_{t+1}$};
\draw[->, line width=1.15pt](stm)--(st);\draw[->, line width=1.15pt](st)--(stp);
\draw[->, line width=1.15pt](stm)--(htm);\draw[->, line width=1.15pt](st)--(ht);\draw[->, line width=1.15pt](stp)--(htp);
\draw[->, line width=1.15pt](dtm)--(dt);\draw[->, line width=1.15pt](dt)--(dtp);
\draw[->, line width=1.15pt](sigmatm)--(dt);\draw[->, line width=1.15pt](sigmat)--(dtp);
\draw[->, line width=1.15pt](sigmatm)--(st);\draw[->, line width=1.15pt](sigmat)--(stp);
\draw[->, line width=1.15pt](dtm)--(sigmatm);\draw[->, line width=1.15pt](dt)--(sigmat);\draw[->, line width=1.15pt](dtp)--(sigmatp);
\draw[->, line width=1.15pt](sigmatm)--(sigmat);\draw[->, line width=1.15pt](sigmat)--(sigmatp);
\draw[->, line width=1.15pt](stm)to [bend right=55](vtm);\draw[->, line width=1.15pt](st)to [bend right=55](vt);\draw[->, line width=1.15pt](stp)to [bend right=55](vtp);
\draw[->, line width=1.15pt](htm)--(vtm);\draw[->, line width=1.15pt](ht)--(vt);\draw[->, line width=1.15pt](htp)--(vtp);
\draw[->, line width=1.15pt](stm)--(dtm);\draw[->, line width=1.15pt](st)--(dt);\draw[->, line width=1.15pt](stp)--(dtp);
\draw[->, line width=1.15pt](htm)--(ht);\draw[->, line width=1.15pt](ht)--(htp);
\end{tikzpicture}}}
\subfigure[]{
\scalebox{0.63}{
\begin{tikzpicture}
\node[disc] (sigmatm) at (2,1.25) {$c_{t-1}$};
\node[disc] (sigmat) at (4,1.25) {$c_t$};
\node[disc] (sigmatp) at (6,1.25) {$c_{t+1}$};
\node[disc] (dtm) at (2,0) {$d_{t-1}$};
\node[disc] (dt) at (4,0) {$d_t$};
\node[disc] (dtp) at (6,0) {$d_{t+1}$};
\node[disc] (stm) at (2,-1.25) {$s_{t-1}$};
\node[disc] (st) at (4,-1.25) {$s_t$};
\node[disc] (stp) at (6,-1.25) {$s_{t+1}$};
\node[ocont] (htm) at (2,-2.5) {$h_{t-1}$};
\node[ocont] (ht) at (4,-2.5) {$h_t$};
\node[ocont] (htp) at (6,-2.5) {$h_{t+1}$};
\node[ocont,obs] (vtm) at (2,-3.75) {$v_{t-1}$};
\node[ocont,obs] (vt) at (4,-3.75) {$v_t$};
\node[ocont,obs] (vtp) at (6,-3.75) {$v_{t+1}$};
\draw[->, line width=1.15pt](stm)--(st);\draw[->, line width=1.15pt](st)--(stp);
\draw[->, line width=1.15pt](stm)--(htm);\draw[->, line width=1.15pt](st)--(ht);\draw[->, line width=1.15pt](stp)--(htp);
\draw[->, line width=1.15pt](dtm)--(dt);\draw[->, line width=1.15pt](dt)--(dtp);
\draw[->, line width=1.15pt](sigmatm)--(dt);\draw[->, line width=1.15pt](sigmat)--(dtp);
\draw[->, line width=1.15pt](sigmatm)--(st);\draw[->, line width=1.15pt](sigmat)--(stp);
\draw[->, line width=1.15pt](dtm)--(sigmatm);\draw[->, line width=1.15pt](dt)--(sigmat);\draw[->, line width=1.15pt](dtp)--(sigmatp);
\draw[->, line width=1.15pt](sigmatm)--(sigmat);\draw[->, line width=1.15pt](sigmat)--(sigmatp);
\draw[->, line width=1.15pt](stm)to [bend right=55](vtm);\draw[->, line width=1.15pt](st)to [bend right=55](vt);\draw[->, line width=1.15pt](stp)to [bend right=55](vtp);
\draw[->, line width=1.15pt](htm)--(vtm);\draw[->, line width=1.15pt](ht)--(vt);\draw[->, line width=1.15pt](htp)--(vtp);
\draw[->, line width=1.15pt](stm)--(dtm);\draw[->, line width=1.15pt](st)--(dt);\draw[->, line width=1.15pt](stp)--(dtp);
\draw[->, color=myred,line width=1.15pt](htm)--(ht);\draw[->, color=myred,line width=1.15pt](ht)--(htp);
\draw[->, color=myred,line width=1.15pt](sigmatm)to [bend left=35](htm);\draw[->, color=myred,line width=1.15pt](sigmat)to [bend left=35](ht);\draw[->, color=myred,line width=1.15pt](sigmatp)to [bend left=35](htp);
\end{tikzpicture}}}
\subfigure[]{
\scalebox{0.63}{
\begin{tikzpicture}
\node[disc] (sigmatm) at (2,1.25) {$c_{t-1}$};
\node[disc] (sigmat) at (4,1.25) {$c_t$};
\node[disc] (sigmatp) at (6,1.25) {$c_{t+1}$};
\node[disc] (dtm) at (2,0) {$d_{t-1}$};
\node[disc] (dt) at (4,0) {$d_t$};
\node[disc] (dtp) at (6,0) {$d_{t+1}$};
\node[disc] (stm) at (2,-1.25) {$s_{t-1}$};
\node[disc] (st) at (4,-1.25) {$s_t$};
\node[disc] (stp) at (6,-1.25) {$s_{t+1}$};
\node[ocont] (htm) at (2,-2.5) {$h_{t-1}$};
\node[ocont] (ht) at (4,-2.5) {$h_t$};
\node[ocont] (htp) at (6,-2.5) {$h_{t+1}$};
\node[ocont,obs] (vtm) at (2,-3.75) {$v_{t-1}$};
\node[ocont,obs] (vt) at (4,-3.75) {$v_t$};
\node[ocont,obs] (vtp) at (6,-3.75) {$v_{t+1}$};
\node [color=myred,above] at (sigmat.north) {$c_t=1$};
\draw[->, line width=1.15pt](stm)--(st);\draw[->, line width=1.15pt](st)--(stp);
\draw[->, line width=1.15pt](stm)--(htm);\draw[->, line width=1.15pt](st)--(ht);\draw[->, line width=1.15pt](stp)--(htp);
\draw[->, line width=1.15pt](dtm)--(dt);\draw[->, line width=1.15pt](dt)--(dtp);
\draw[->, line width=1.15pt](sigmatm)--(dt);\draw[->, line width=1.15pt](sigmat)--(dtp);
\draw[->, line width=1.15pt](sigmatm)--(st);\draw[->, line width=1.15pt](sigmat)--(stp);
\draw[->, line width=1.15pt](dtm)--(sigmatm);\draw[->, line width=1.15pt](dt)--(sigmat);\draw[->, line width=1.15pt](dtp)--(sigmatp);
\draw[->, line width=1.15pt](sigmatm)--(sigmat);\draw[->, line width=1.15pt](sigmat)--(sigmatp);
\draw[->, line width=1.15pt](stm)to [bend right=55](vtm);\draw[->, line width=1.15pt](st)to [bend right=55](vt);\draw[->, line width=1.15pt](stp)to [bend right=55](vtp);
\draw[->, line width=1.15pt](htm)--(vtm);\draw[->, line width=1.15pt](ht)--(vt);\draw[->, line width=1.15pt](htp)--(vtp);
\draw[->, line width=1.15pt](stm)--(dtm);\draw[->, line width=1.15pt](st)--(dt);\draw[->, line width=1.15pt](stp)--(dtp);
\draw[->, line width=1.15pt](htm)--(ht);
\draw[->, line width=1.15pt](sigmatm)to [bend left=35](htm);\draw[->, line width=1.15pt](sigmat)to [bend left=35](ht);\draw[->, line width=1.15pt](sigmatp)to [bend left=35](htp);
\end{tikzpicture}}}
\caption{(a): Explicit-duration \SLGSSM~using count-duration variables.
(b): Across-segment independence is enforced with a link from $c_t$ to $h_t$,
as explicitly represented in (c).}
\label{fig:SLGSSMCD}
\end{figure}
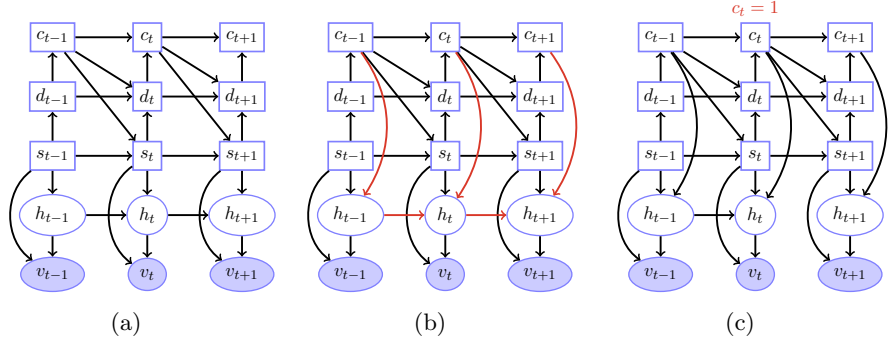
The explicit-duration \SLGSSM~using count-duration variables has belief network representation given in \figref{fig:SLGSSMCD}(a).
Across-segment independence can be enforced with a link from $c_t$ to $h_t$ as in \figref{fig:SLGSSMCD}(b),
which has the effect of removing the link from $h_t$ to $h_{t+1}$ if $c_t=1$, as explicitly represented in \figref{fig:SLGSSMCD}(c). More specifically, dependence cut is defined as
\begin{align*}
p(h_t|h_{t-1},c_{t-1},s_t)&=\begin{cases}
    p(h_t|c_{t-1},s_t)={\cal N}(h_t;\mu^{s_t},\Sigma^{s_t}) & \hspace{-0.0cm} \textrm{if } c_{t-1}\!=\!1\\
	{\cal N}(h_t;A^{s_t}h_{t-1},\Sigma^{s_t}_H) 	& \hspace{-0.0cm} \textrm{if } c_{t-1}\!>\!1.
\end{cases}
\end{align*}
In this section, we only discuss the across-segment-independence case and leave the description of the across-segment-dependence case to Appendix~\ref{app:SS}.

If only segmentation is of interest, we can employ the segmental inference approach described in \secref{sec:M2Inf}
with segment-emission distribution
$e^{s_t,d_t}_t=p(v_{t-d_t+1:t}|\sigma^{1}_t)$ estimated as the likelihood of a \LGSSM.
Naive estimation would require to perform filtering in a \LGSSM~with cost ${\cal O}(d_t)$ for each $t$, $s_t$ and $d_t$, and therefore with
total cost ${\cal O}(TS^2d_{\max}^2)$. However, the cost can be reduced to ${\cal O}(TS^2d_{\max})$ by recursive computation of $e^{s_t,d_t}_t$ as,
dropping conditioning on the regime and count-duration variables,
\begin{align*}
e^{s_t,d_t}_t
\!=\!p(v_t|v_{t-d_t+1:t-1})\hskip-0.4cm\prod_{\tau=t-d_t+1}^{t-1}\hskip-0.4cm p(v_{\tau}|v_{t-d_t+1:\tau-1})=p(v_t|v_{t-d_t+1:t-1})e^{s_t,d_t-1}_{t-1}\!\!,
\end{align*}
with $p(v_t|v_{t-d_t+1:t-1})\!=\!{\cal N}(B\hat h_t^{t-1},BP_t^{t-1}B\trans+\Sigma_V)$, where $\hat h_t^{t-1}$ and $P_t^{t-1}$ are the mean and covariance of $p(h_t|v_{1:t-1})$.

If also estimation of the smoothed distribution $p(h_t|v_{1:T})$ is of interest, $\gamma^{\sigma_t}_t$ can be obtained from the equivalence $\gamma^{\sigma_t}_t=\gamma^{s_t,d_t,1}_{t+c_t-1}$, 
where $\gamma^{s_t,d_t,1}_{t+c_t-1}$ can be computed with segment-recursive routines.

If also estimation of the filtered distribution $p(h_t|v_{1:t})$ is of interest, a time-recursive routine for $\alpha^{\sigma_t}_t$ with cost ${\cal O}(TS^2d_{\max})$ is required (recursion (\ref{eq:alphaM2t})).

The distributions $\hat\alpha^{\sigma_t}_t=p(h_t|\sigma_t,v_{1:t})$ and $\hat\gamma^{\sigma_t}_t=p(h_t|\sigma_t,v_{1:T})$ can be obtained with cost ${\cal O}(TSd_{\max})$ and ${\cal O}(TSd^2_{\max})$ respectively.

Indeed, the computation of $\hat\alpha^{\sigma_t}_t=p(h_t|\sigma_t,v_{t-d_t+c_t:t})$ would seem to
require filtering in a \LGSSM~with cost ${\cal O}(d_t)$ for each $t$ and $\sigma_t$, and therefore with
total cost ${\cal O}(TSd_{\max}^3)$.
However, we can observe that $\hat\alpha^{\sigma_t}_t$ is equivalent to all $\hat\alpha^{\sigma_t'}_t$ for which $s'_t=s_t$ and for which $d_t'-c_t'=d_t-c_t$ (\ie~for which the segment starts at time-step $t-d_t+c_t$).
Therefore, only filtering in a \LGSSM~with cost ${\cal O}(d_{\max})$ on segment $v_{t:t+d_{\max}-1}$ for each $t$ and $s_t$ is required.
The same observation can be made from the time-recursive routine (\ref{eq:alphahatM2t}).

The computation of $\hat\gamma^{\sigma_t}_t=p(h_t|\sigma_t,v_{t-d_t+c_t:t+c_t-1})$ requires smoothing in the same \LGSSM~as the computation of
$\hat\gamma^{s_t,d_t,d_t}_{t-d_t+c_t}, \ldots, \hat\gamma^{s_t,d_t,c_t+1}_{t-1}$, $\hat\gamma^{s_t,d_t,c_t-1}_{t+1},\ldots,\gamma^{s_t,d_t,1}_{t+c_t-1}$,
as the same segment $v_{t-d_t+c_t:t+c_t-1}$ is involved.
Therefore, smoothing in a \LGSSM~with cost ${\cal O}(d_t)$
on segment $v_{t:t+d_t-1}$ for each $t$, $s_t$ and $d_t$ is required.
The same observation can be made from the time-recursive routine (\ref{eq:gammahatM2t}).

Notice that the use of uncollapsed count variables, rather than collapsed ones as done in the standard approach to explicit-duration modelling \citep{ferguson80variable,rabiner89tutorial,ostendorf96from,murphy2002hsm,yu10hidden}, simplifies the derivation of $p(h_t|v_{1:t})$ and $p(h_t|v_{1:T})$.

\subsubsection{Movement segmentation example\label{sec:uslgssm}}
\begin{figure}[t] 
\includegraphics[]{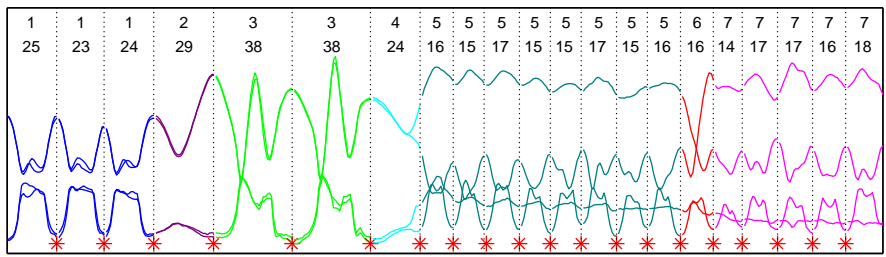}
\caption{\Timeseries~corresponding to the recording of the leg positions of an individual performing repetitions of the actions low jumping up and down, high
jumping up and down, hopping on the left foot, and hopping on the right foot.
The dotted vertical lines give a manual segmentation into 7 basic movements and their repetitions. The numbers in the first row indicate the movement types,
whilst the numbers in the second row indicate the durations.
The stars indicate the segmentation obtained with the explicit-duration \SLGSSM~(the movement types were all correctly inferred).}
\label{fig:uslgssm}
\end{figure}
In this section we show that the explicit-duration \SLGSSM~with across-segment independence and the constraint $\pi_{ii}\neq 0$ can be used to solve the
segmentation task discussed in \chapref{chap:intro}, namely to segment
the \timeseries~displayed in \figref{fig:uslgssm} -- corresponding to the recording of the leg positions of an individual performing repetitions of the actions low jumping up and down, high
jumping up and down, hopping on the left foot, and hopping on the right foot -- into the underlying actions and their repetitions.

The \timeseries~was manually segmented with the help of an associated video, assuming 7 basic movement types. The manual segmentation is shown in \figref{fig:uslgssm}, where  
the dotted vertical lines indicate the movement starts, the numbers in the first row indicate the movement types, and the numbers in the second row indicate the durations.

We used the manual segmentation and 7 \LGSSMs~to learn the parameters representing each movement type. We then performed extended Viterbi with 
an explicit-duration \SLGSSM~using the learned parameters and employing a uniform segment-duration distribution,
with minimum and maximum durations 15 and 50 for the first 4 types of movement respectively, and 10 and 25 for the second 3 types of movement respectively. 

The model correctly inferred all movement types and accurately detected the movement starts, as indicated by the stars in \figref{fig:uslgssm}.

\section{Approximations\label{sec:Approx}}
Whilst empowering standard \HMSMs~with stronger modelling capabilities, explicit-duration \HMSMs~can have
high computational cost.
For simplicity, consider the case of one regime only. The computation of $\alpha^{\sigma_t}_t=p(\sigma_t|v_{1:t})$ and $\gamma^{\sigma_t}_t=p(\sigma_t|v_{1:T})$ has cost ${\cal O}(Td_{\max})$.
In models with unobserved variables $h_{1:T}$ related by first-order Markovian dependence, the computation of $\hat\alpha^{\sigma_t}_t=p(h_t|\sigma_t,v_{1:t})$
and $\hat\gamma^{\sigma_t}_t=p(h_t|\sigma_t,v_{1:T})$ has also at best cost ${\cal O}(Td_{\max})$.
If $d_{\max}$ is large the cost becomes prohibitive. If $d_{\max}=\infty$, the cost becomes at best ${\cal O}(T^2)$.
Several approximation techniques have been proposed in the literature to address this issue.

A review of approximation methods introduced for extended Viterbi in the explicit-duration \HMM~is given in \cite{ostendorf96from}.
The basic idea is to reduce the space of possible segmentations by constraining the maximization. For example, in the segmental extended
Viterbi of \secref{sec:M2Inf}, maximization over $d_t$ can be constrained to a subset ${\cal D}_t$ of the original set $\{d_{\min},\ldots,d_{\max}\}$.
The subset ${\cal D}_t$ can be chosen in advance with a simpler model or during the decoding.

Pruning methods were also introduced in the changepoint/reset model literature. In 
changepoint/reset models, dependence from the past is cut at the occurrence of a changepoint.
Older approaches employ one regime only, and therefore the occurrence of a changepoint corresponds to the reset of the current dynamics to its initial condition.
More recent approaches can employ more than one regime, and therefore the occurrence of a changepoint can also correspond to the reset and change of the current dynamics.
Whilst the goal of changepoint models is only to detect abrupt changes in the \timeseries, reset models are also often used as approximations of complex models. 
Commonly, changepoint/reset models do not impose constraints on the segment duration.

Older approaches to changepoint models fix the number of changepoints a priori. More recent Bayesian approaches
define a distribution on the number and positions of changepoints through a segment-duration distribution \citep{fearnhead06exact,fearnhead07online,adams07bayesian,fearnhead09bayesian,eckley11analysis}.
This is achieved by using either increasing count variables or variables that indicate the time-step
of the most recent changepoint prior to time-step $t$ (\ie~$c_t=c_{t-1}$ or $c_t=t-1$),
which provide the same information as increasing count variables. 

In reset models, dependence cut is commonly obtained with a variable $c_t$ taking value 1 when a changepoint occurs and 2 otherwise
\citep{cemgil06generative,barber10graphical} (\eg~with $p(c_t=2)=\lambda$ and $p(c_t=1)=1-\lambda$, which gives a geometric segment-duration distribution -- this approach can be seen as a special case of the increasing-count-variable approach).
As discussed in \secref{sec:SLGSSM}, \cite{bracegirdle11switch} and \cite{bracegirdle13inference} recently suggested the use of increasing count variables and increasing-decreasing count variables in reset models in order 
to achieve sequential filtering-smoothing and to approximate inference. 

To understand the basic idea of the pruning methods suggested, consider the increasing-count-variable approach with $d_{\max}=\infty$ and the constraint $p(c_t > t)=0$ (obtained, \eg, by imposing $\tilde{\lambda}_1=1$, see \secref{sec:M1alt}). From recursion (\ref{eq:alphaM1alt}),
we can deduce that $\alpha^{\sigma_t}_t=0$ implies $\alpha^{s_t,c_t+1}_{t+1}=\ldots=\alpha^{s_t,c_t+T-t}_T=0$,
\ie~if according to $v_{1:t}$ a segment starting at time-step $t-c_t+1$ and generated by $s_t$ cannot have duration $\geq c_t$, that segment cannot have duration $\geq c_t+1$ after incorporating observations $v_{t+1}$, etc.
If only ${\cal D}$ elements of $\alpha^{\sigma_t}_t$ are non-zero, then only ${\cal D}+1$ elements of $\alpha^{s_t,c_{t+1}}_{t+1}$ are non-zero, and so on.
Therefore, we can retain only ${\cal D}$ elements of the count variable by eliminating one element at each time-step, reducing the computational cost from ${\cal O}(T^2)$ to ${\cal O}(T{\cal D})$.
As $\alpha^{\sigma_t}_t=0$ implies $\gamma^{\sigma_t}_t=\gamma^{s_t,c_t+1}_{t+1}=\ldots=\gamma^{s_t,c_t+T-t}_T=0$ (recursion (\ref{eq:gammaM1alt})), pruning of $\alpha^{\sigma_t}_t$ automatically reduces the cost of computing $\gamma^{\sigma_t}_t$ to ${\cal O}(T{\cal D})$.

A similar reasoning can be made in the count-duration-variable approach with $d_{\max}=\infty$ and the constraint $p(d_t>t, c_t=1)=0$ (obtained, \eg, by imposing $\tilde{\tilde{\rho}}_{d_1d_1}=1$, see \secref{sec:M2}), by looking at time-recursive routines (\ref{eq:alphaM2t}) and (\ref{eq:gammaM2t}).
From routine (\ref{eq:alphaM2t}), we deduce that $\alpha^{\sigma_t}_t=0$ implies $\alpha^{s_t,d_t,c_t-1}_{t+1}=\ldots=\alpha^{s_t,d_t,1}_{t+c_t-1}=0$
and $\alpha^{s_t,d_t',c_t'}_t=0$ if $d_t-c_t=d_t'-c_t'$, \ie~if according to $v_{1:t}$ a segment starting at time-step $t-d_t+c_t$ and generated by $s_t$ cannot have duration $\geq d_t-c_t+1$, that segment cannot have duration $\geq d_t-c_t+2$
after incorporating observations $v_{t+1}$, etc.
From routine (\ref{eq:gammaM2t}), we deduce that $\alpha^{\sigma_t}_t=0$ implies $\gamma^{\sigma_t}_t=\gamma^{s_t,d_t,c_t-1}_{t+1}=\ldots=\gamma^{s_t,d_t,1}_{t+c_t-1}=0$.

For the explicit-duration \SLGSSM~with across-segment independence, this pruning procedure implies that, instead of \LGSSM~filtering on $v_{t:T}$ for each $t$ and $s_t$ with cost ${\cal O}(T^2S)$ and \LGSSM~smoothing
on $v_{t:t+d_t-1}$ for each $t$, $d_t\in\{1,\ldots,T-t+1\}$ and $s_t$ with cost ${\cal O}(T^3S)$, only \LGSSM~filtering on $v_{t:t+{\cal D}-1}$ for each $t$ and $s_t$ with cost ${\cal O}(TS{\cal D})$
and \LGSSM~smoothing on $v_{t:t+d_t-1}$ for each $t$ and $d_t\in\{1,\ldots,{\cal D}\}$ with cost ${\cal O}(T{\cal D}^2)$ is required.

Pruning is performed using a resampling idea from \cite{fearnhead07online,liu98rejection} in \cite{fearnhead09bayesian},
and by dropping the element of $\alpha^{\sigma_t}_t$ with lowest value in \cite{bracegirdle11switch} and \cite{bracegirdle13inference}.

Other approximation techniques based on combining regime variables with and without corresponding explicit-duration variables, on binning the duration distribution and on beam-sampling were proposed in \cite{stanke93gene}, \cite{jiang10hidden} and \cite{dewar12inference} respectively.

\chapter{Discussion\label{chap:Discuss}}
Explicit-duration Markov switching models (\HMSMs)~enrich the modelling capabilities of standard \HMSMs~with the possibility to define segment-duration distributions of any form,
to impose complex dependence between the observations, and to reset the dynamics to initial conditions.

From a generative viewpoint, they differ from standard \HMSMs~as the regime variable $s_t$ is either sampled from the transition distribution or
set to $s_{t-1}$, depending on the values taken by the explicit-duration variables. This mechanism is achieved through
a first-order Markov chain on the combined regime and explicit-duration variables that
partitions the \timeseries~into segments, with boundaries at those time-steps in which sampling occurs.

The first-order Markov chain can be defined using three fundamentally different encodings for the explicit-duration variables,
namely distance to current-segment end with decreasing count variables, distance to current-segment beginning
with increasing count variables, or distance to both current-segment beginning and current-segment end with count-duration variables.
%
%
%

Different encoding leads to different possible structures for the conditional distribution 
of the observations. Information about both segment beginning and segment end allows the most complex structure, namely any conditional distribution within a segment.
In this complex case, inference can only be achieved with recursions that operate at a segment level rather than at a single time-step level.

In models that have complex unobserved structure,
different encoding gives rise to different computational cost and approximation requirements for inference.
As we have seen in \secref{sec:SLGSSMCD}, in models containing additional unobserved variables related by first-order Markovian 
dependence, increasing count variables are overall preferable.

In the literature, explicit-duration \HMSMs~are most commonly called hidden semi-Markov models or segment models
and are informally described as extensions of standard \HMSMs~in which, rather than single observations, segments of observations are generated
from a sampled regime \citep{ostendorf96from,yu10hidden}.
They originate from the idea to extend the hidden Markov model by defining a semi-Markov process on the regime variables.
The original approach, introduced in \cite{ferguson80variable} and later re-explained in \cite{rabiner89tutorial}, achieves that
by introducing duration variables, and by deriving inference recursions that operate at a segment level. 
This approach is currently the most common approach to explicit-duration modelling.

Although count-duration variables are mentioned in \cite{yu10hidden}, their use to simplify derivations with respect to the standard approach
first appeared in \cite{chiappa10movement}.
As we have seen in \secref{sec:M2Inf} and Appendix~\ref{app:SL}, computing posterior distributions at time-steps that do not correspond to segment ends with count-duration variables 
is more immediate than with the standard approach. The benefit is particularly evident when
full inference in models that have complex unobserved structure is required, as discussed in \secref{sec:SLGSSMCD}.

The decreasing-count-variable approach with independence among observations was introduced in \cite{yu03efficient} to enable the derivation of
computationally less expensive inference routines than the segmental routines.
However, as explained in \secref{sec:M2Inf} and already observed in \cite{mitchell95complexity} and \cite{murphy2002hsm},
the same improvement can also be reached with recursive computation of the segment-emission distribution in the segmental routines.

Work in the direction of increasing count variables first appeared in \cite{djuric02mcmc}, but explicit introduction was given in \cite{huang06variable} and \cite{oh08learning}.


\begin{acknowledgements}
\addcontentsline{toc}{chapter}{Acknowledgements}
This work has partly been funded by the Marie Curie Intra European Fellowship IEF-237555, by Microsoft Research Cambridge, and by Microsoft Research Connections.
The author would like to thank David Barber and Shakir Mohamed for helping with the proof-reading of the manuscript.
\end{acknowledgements}

\appendix
\chapter{Miscellaneous}

\section{EM in the Switching Autoregressive Model\label{app:SARM}}
Consider the switching autoregressive model (\ref{eq:sarm})
with $\tilde v_t=[v_{t-k} \ldots v_{t-1}]\trans$, where the symbol $\trans$ indicates the transpose operator, and $a^{s_t}=[a^{s_t}_1 \ldots a^{s_t}_k]$.
The expectation of the complete data log-likelihood is given by (omitting the first $k$ observations)
\begin{align*}
&\hskip-0.05cm\sum_{t=k+1}^T\!\!\av{\log p(v_t|s_t,v_{t-k:t-1})}_{\gamma^{s_t}_t}
\!+\!\av{\log p(s_1)}_{\gamma^{s_1}_1}\!+\!\sum_{t=2}^T\av{\log p(s_t|s_{t-1})}_{\tilde\gamma^{s_{t-1:t}}_t}\!=\!\\
\hskip-0.05cm&\!-\!\frac{1}{2}\sum_{t}\av{\log (\sigma^{s_t})^2\!+\!\frac{(v_t\!-\!a^{s_t}\tilde v_t)^2}{(\sigma^{s_t})^2}}_{\gamma^{s_t}_t}
\!+\!\av{\log \tilde\pi_{s_1}}_{\gamma^{s_1}_1}\!+\!\sum_{t}\av{\log \pi_{s_ts_{t-1}}}_{\tilde\gamma^{s_{t-1:t}}_t},
\end{align*}
where (see \eqref{eq:gamma})
\begin{align*}
\tilde\gamma^{s_{t-1:t}}_t&=p(s_{t-1:t}|v_{1:T})
=\frac{\pi_{s_ts_{t-1}}\alpha^{s_{t-1}}_{t-1}}{\sum_{\tilde s_{t-1}}\pi_{s_t\tilde s_{t-1}}\alpha^{\tilde s_{t-1}}_{t-1}}\gamma^{s_t}_t,
\end{align*}
giving updates
\begin{align*}
&a^{s_t}=\sum_{t}\gamma^{s_t}_tv_t\tilde v_t\trans\big(\sum_{t}\gamma^{s_t}_t\tilde v_t\tilde v_t\trans\big)^{-1},
\hskip0.3cm(\sigma^{s_t})^2=\frac{\sum_{t}\gamma^{s_t}_t(v_t-a^{s_t}\tilde v_t)^2}{\sum_{t}\gamma^{s_t}_t},\\
&\tilde \pi_{s_1}=\gamma^{s_1}_1,
\hskip0.3cm\pi_{s_ts_{t-1}}=\frac{\sum_{t=2}^T\tilde\gamma^{s_{t-1:t}}_t}{\sum_{t=2}^T\sum_{\tilde s_t}\tilde\gamma^{s_{t-1},\tilde s_t}_t}.
\end{align*}

\section{\HMM~as a Decreasing-Count-Variable \HMSM\label{app:HMM}}
The \HMM~with initial-regime distribution $\tilde{\hat{\pi}}$ and transition distribution $\hat\pi$ has the same joint distribution $p(s_{1:T},v_{1:T})$ of a decreasing-count-variable \HMSM~with $d_{\min}=1, d_{\max}=\infty$, and with
\begin{align*}
&\tilde{\pi}_{s_1}=\tilde{\hat{\pi}}_{s_1}\,,\hskip0.3cm
\pi_{s_{t+1}s_t}=
\begin{cases}
\frac{\hat\pi_{s_{t+1}s_t}}{1-\hat\pi_{s_ts_t}} & \textrm{if } s_{t+1}\!\neq\!s_t\\
0 & \textrm{if } s_{t+1}\!=\!s_t\,,
\end{cases}\\
&\tilde{\rho}_{\sigma_1}=\hat\pi^{c_1-1}_{s_1s_1}(1-\hat\pi_{s_1s_1})\,,\hskip0.3cm
\rho_{\sigma_t}=\hat\pi^{c_t-1}_{s_ts_t}(1-\hat\pi_{s_ts_t}).
\end{align*}
This can be demonstrated by showing that
\begin{align}
\sum_{c_{1:T}}p(s_1)p(c_1|s_1)\prod_{t=2}^T p(s_t|\sigma_{t-1})p(c_t|s_t,c_{t-1})=\tilde{\hat{\pi}}_{s_1}\prod_{t=2}^T\hat \pi_{s_ts_{t-1}}.
\label{eq:hmmspec2}
\end{align}
Since $\pi_{s_ts_t}=0$, $s_{1:T}$ determine the values of the count variables at all time-steps with exception of the last segment.
Let's consider the case of two or more regime changes (the other cases can be demonstrated similarly).
Suppose that two consecutive regime changes occur at time-steps $\tau+1>1$ and $\tau+d+1\leq T$, \ie~$s_{\tau}\neq s_{\tau+1}=\cdots=s_{\tau+d}\neq s_{\tau+d+1}$. Then
$c_{\tau}=1,c_{\tau+1}=d,\ldots,c_{\tau+d-1}=2,c_{\tau+d}=1$, and therefore
\begin{align}
\hskip-0.1cm\prod_{t=\tau+1}^{\tau+d}\!\!\!p(s_t|\sigma_{t-1})p(c_t|s_t,c_{t-1})&=p(s_{\tau+1}|s_{\tau},c_{\tau}\!=\!1)p(c_{\tau+1}|s_{\tau+1},c_{\tau}\!=\!1)\nonumber\\
\hskip-0.1cm&=(1\!-\!\hat\pi_{s_{\tau+1}s_{\tau+1}})\pi_{s_{\tau+1}s_{\tau}}\prod_{t=\tau+2}^{\tau+d}\!\!\!\hat\pi_{s_ts_{t-1}}\nonumber\\
\hskip-0.1cm&=\frac{1\!-\!\hat\pi_{s_{\tau+1}s_{\tau+1}}\!=\!n_{\tau+1}}{1\!-\!\hat\pi_{s_{\tau}s_{\tau}}\!=\!n_{\tau}}\prod_{t=\tau+1}^{\tau+d} \hat\pi_{s_ts_{t-1}}.
\label{eq:hmmspec}
\end{align}
There are two possible scenarios for the change of regime after time-step $\tau+d+1$, namely it occurs
\begin{itemize}
\item At time-step $T$ or before, \ie~$s_{\tau+d+1}=\cdots=s_{\tau+d'}\neq s_{\tau+d'+1}$ with $\tau+d'<T$, giving
\begin{align*}
\prod_{t=\tau+d+1}^{\tau+d'}\!\!\!p(s_t|\sigma_{t-1})p(c_t|s_t,c_{t-1})&
\!=\!\frac{1\!-\!\hat\pi_{s_{\tau+d+1}s_{\tau+d+1}}}{1\!-\!\hat\pi_{s_{\tau+d}s_{\tau+d}}\!=\!n_{\tau+1}}
\prod_{t=\tau+d+1}^{\tau+d'}\!\!\!\!\hat\pi_{s_ts_{t-1}}.
\end{align*}
\item After time-step $T$, \ie~$s_{\tau+d+1}=\cdots=s_T$, giving
\begin{align*}
\sum_{c_{\tau+d+1:T}}\prod_{t=\tau+d+1}^T &p(s_t|\sigma_{t-1})p(c_t|s_t,c_{t-1})=\pi_{s_{\tau+d+1}s_{\tau+d}}\\
&\times\underbrace{(1\!-\!\hat\pi_{s_{\tau+d+1}s_{\tau+d+1}})\hskip-0.2cm\sum_{c_{\tau+d+1}=T-\tau-d}^{\infty}\hskip-0.2cm\hat\pi_{s_{\tau+d+1}s_{\tau+d+1}}^{c_{\tau+d+1}-1}}_{\hat\pi_{s_{\tau+d+1}s_{\tau+d+1}}^{T-\tau-d-1}}\\
&=\frac{1}{1\!-\!\hat\pi_{s_{\tau+d}s_{\tau+d}}\!=\!n_{\tau+1}}\prod_{t=\tau+d+1}^T\hskip-0.1cm\hat\pi_{s_ts_{t-1}}.
\end{align*}
\end{itemize}
Analogously, there are two possible scenarios for the change of regime before time-step $\tau+1$, namely it occurs
\begin{itemize}
\item After time-step 1, \ie~$s_{\tau-d'}\neq s_{\tau-d'+1}=\cdots=s_{\tau}$ with $\tau-d'+1>1$, giving
\begin{align*}
\prod_{t=\tau-d'+1}^{\tau}\hskip-0.4cm p(s_t|\sigma_{t-1})p(c_t|s_t,c_{t-1})&\!=\!\frac{1\!-\!\hat\pi_{s_{\tau-d'+1}s_{\tau-d'+1}}\!\!=\!n_{\tau}}{1\!-\!\hat\pi_{s_{\tau-d'}s_{\tau-d'}}}
\hskip-0.45cm\prod_{t=\tau-d'+1}^{\tau}\hskip-0.4cm \hat\pi_{s_ts_{t-1}}.
\end{align*}
\item At time-step 1 or before, giving
\begin{align*}
\prod_{t=1}^{\tau} p(s_t|\sigma_{t-1})p(c_t|s_t,c_{t-1})&=(1\!-\!\hat\pi_{s_{1}s_{1}}
\!=\!n_{\tau})\tilde{\hat{\pi}}_{s_1}\prod_{t=2}^{\tau} \hat\pi_{s_ts_{t-1}}.
\end{align*}
\end{itemize}
Therefore, in \eqref{eq:hmmspec2}, $n_{\tau+1}$ of \eqref{eq:hmmspec} cancels with $n_{\tau+1}$ in the following regime, whilst $n_{\tau}$ cancels with $n_{\tau}$ in the preceding regime.

Notice that, to use the model, conditioning on the event $c_T=1$ would be required and
the equivalence would not longer hold.

\noindent The recursion for $p(s_t,v_{1:t})$ using recursion (\ref{eq:alphaM1st}) reduces to the \HMM~recursion for $p(s_t,v_{1:t})$ (see recursion (\ref{eq:alpha})).
Indeed
\begin{align*}
p(s_t,&v_{1:t})=\sum_{c_t=1}^{\infty}p(\sigma_t,v_{1:t})=\sum_{c_t=1}^{\infty}\hat\alpha^{\sigma_t}_t\\
&=p(v_t|s_t)\sum_{c_t=1}^{\infty}\bigg\{\hat\alpha^{s_t,c_t+1}_{t-1}+\rho_{\sigma_t}
\sum_{s_{t-1}\neq s_t}\pi_{s_ts_{t-1}}\hat\alpha^{s_{t-1},1}_{t-1}\bigg\}\\
&=p(v_t|s_t)\bigg\{\sum_{c_t=1}^{\infty}\hat\pi_{s_ts_t}\hat\alpha^{\sigma_t}_{t-1}\!+\!\hskip-0.2cm\sum_{s_{t-1}\neq s_t}\hskip-0.25cm
\pi_{s_ts_{t-1}}\hskip-0.55cm\underbrace{(1\!-\!\hat\pi_{s_ts_t})\sum_{c_t=1}^{\infty}\hat\pi^{c_t-1}_{s_ts_t}}_{(1\!-\!\hat\pi_{s_{t-1}s_{t-1}})\sum_{c_t=1}^{\infty}\hat\pi^{c_t-1}_{s_{t-1}s_{t-1}}}\hskip-0.55cm\hat\alpha^{s_{t-1},1}_{t-1}\bigg\}\\
&=p(v_t|s_t)\sum_{s_{t-1}}\hat\pi_{s_ts_{t-1}}\sum_{c_{t-1}=1}^{\infty}\hat\alpha^{\sigma_{t-1}}_{t-1},
\end{align*}
where $\hat\alpha^{s_t,c_t}_t=\hat\pi_{s_ts_t}\hat\alpha^{s_t,c_t-1}_t$ for $c_t>1$, and therefore $\hat\alpha^{\sigma_t}_t=\hat\pi_{s_ts_t}^{c_t-1}\hat\alpha^{s_t,1}_t$,
can be proven by induction. The proof is trivial for $t=1$. Suppose that the result holds for $t-1$, then
it holds for $t$ as
\begin{align*}
\hat\alpha^{\sigma_t}_t
&=p(v_t|s_t)\bigg\{\hat\alpha^{s_t,c_t+1}_{t-1}+\rho_{\sigma_t}\sum_{s_{t-1}\neq s_t}\pi_{s_ts_{t-1}}\hat\alpha^{s_{t-1},1}_{t-1}\bigg\}\\
&=p(v_t|s_t)\bigg\{\hat\pi_{s_ts_t}\hat\alpha^{s_t,c_t}_{t-1}
+\hat\pi_{s_ts_t}\hat\pi_{s_ts_t}^{c_t-1-1}(1-\hat\pi_{s_ts_t})\hskip-0.1cm\sum_{s_{t-1}\neq s_t}\hskip-0.1cm\pi_{s_ts_{t-1}}
\hat\alpha^{s_{t-1},1}_{t-1}\bigg\}\\
&=\hat\pi_{s_ts_t}\hat\alpha^{s_t,c_t-1}_t.
\end{align*}
The \HMM~recursion for $p(v_{t+1:T}|s_t,v_{t-k+1:t})$ (see recursion (\ref{eq:beta})) cannot be obtained.

\section{Relation between EM in \secref{sec:M2L} and in \cite{rabiner89tutorial}\label{app:SL}}
In the explicit-duration \HMM~of \cite{rabiner89tutorial}, $\alpha_t(s_t)$ (Equation (65)) corresponds to the sum over $d_t$ of $\bar\alpha^{\sigma^1_t}=p(s_t,d_t,c_t=1,v_{1:t})$, 
whilst $\beta_t(s_t)$ (Equation (72)) is equivalent to $\beta^{s_t,1}_t=p(v_{t+1:T}|s_t,c_t=1)$.
\cite{rabiner89tutorial} additionally defines
the joint probability of observations up to time $t$ and change to regime $s_{t+1}$ at time-step $t+1$, $\alpha^*_t(s_{t+1})$ (Equation (71)),
and the probability of observations from time $t+1$ given change to regime $s_{t+1}$, $\beta^*_t(s_{t+1})$ (Equation (73)).

\noindent The update for the segment-duration distribution is given by (Equation (81))
\begin{align*}
\bar{\rho}_{s_t}(d_t)=\frac{\sum_{t=1}^T \alpha^*_t(s_t)\rho_{s_t}(d_t)\beta_{t+d_t}(s_t)\prod_{s=t+1}^{t+d_t}b_{s_t}(O_s)}{\sum_{d_t=1}^D\sum_{t=1}^T \alpha^*_t(s_t)\rho_{s_t}(d_t)\beta_{t+d_t}(s_t)\prod_{s=t+1}^{t+d_t}b_{s_t}(O_s)}.
\end{align*}
From the relation between $\alpha_t(s_t)$ and $\alpha^*_t(s_t)$ (Equation (75))
\begin{align*}
\alpha_t(s_t)=\sum_{d_t} \alpha^*_{t-d_t}(s_t)\rho_{s_t}(d_t)\prod_{s=t-d_t+1}^{t}b_{s_t}(O_s),
\end{align*}
we obtain $\alpha^*_t(s_t)\rho_{s_t}(d_t)\prod_{s=t+1}^{t+d_t}b_{s_t}(O_s)=\bar\alpha^{s_t,d_t,1}_{t+d_t}$, which gives equivalence with update (\ref{eq:updatepho}).

The update for the transition distribution is given by (Equation (79))
\begin{align*}
\bar{a}_{s_{t-1}s_t}=\frac{\sum_{t=2}^{T}\alpha_{t-1}(s_{t-1})a_{s_{t-1}s_t}\beta^*_{t-1}(s_t)}{\sum_{j=1}^N\sum_{t=2}^{T}\alpha_{t-1}(s_{t-1})a_{s_{t-1}s_t}\beta^*_{t-1}(s_t)}.
\end{align*}
\eqref{eq:updatepho1} can be expressed in terms of $\bar\alpha^{\sigma^1_{t-1}}_{t-1}$ and $\beta^{s_t,1}_{t+d_t-1}$ as
\begin{align*}
p(s_{t-1},c_{t-1}\!=\!1,s_t|v_{1:T})&=\frac{\pi_{s_ts_{t-1}}\sum_{d_{t-1}}\alpha^{\sigma^1_{t-1}}_{t-1}}
{\sum_{\tilde s_{t-1}}\pi_{s_t\tilde s_{t-1}}\sum_{\tilde d_{t-1}}\alpha^{\tilde\sigma^1_{t-1}}_{t-1}}\sum_{d_t}\gamma^{s_t,d_t,1}_{t+d_t-1}\\
&\propto\pi_{s_ts_{t-1}}\sum_{d_{t-1}}\bar\alpha^{\sigma^1_{t-1}}_{t-1}\sum_{d_t}\beta^{s_t,1}_{t+d_t-1}\rho_{s_td_t}p(v_{t:t+d_t-1}).
\end{align*}
From the relation between $\beta^*_t(s_t)$ and $\beta_t(s_t)$ (Equation (77)), we obtain 
\begin{align*}
\beta^*_{t-1}(s_t)=\sum_{d_t}\hat\beta_{t+d_t-1}(s_t)\rho_{s_t}(d_t)\prod_{s=t}^{t+d_t-1}b_{s_t}(O_s),
\end{align*}
and therefore $p(s_{t-1},c_{t-1}\!=\!1,s_t|v_{1:T}) \propto \pi_{s_ts_{t-1}} \alpha_{t-1}(s_{t-1}) \beta^*_{t-1}(s_t)$, which gives equivalence with update (\ref{eq:updatepi}).

The smoothed distribution $p(s_t|v_{1:T})$ is computed as $p(s_t|v_{1:T})\propto \sum_{\tau<t}^{}\alpha^{*}_{\tau}(s_t)\beta^{*}_{\tau}(s_t)-\beta_{\tau}(s_t)\alpha_{\tau}(s_t)$ (Equation (80)),
\ie~by summing over the set of segments passing through time-step $t$,
which is obtained by subtracting all segments ending before time-step $t$ from all segments starting at time-step $t$ or before.
In \eqref{eq:gpost}, we instead obtain this set as the set of segments that start at time-step $t$ or before and end at time-step $t$ or after.

\section{Robot Localization with the \SLGSSM\label{app:RL}}
In this section, we describe in detail the robot localization problem discussed \secref{chap:intro} and in \secref{sec:SLGSSM}.
Consider a two-wheeled robot moving at constant velocity in the two-dimensional plane. At each time-step, the robot undertakes one of the following three types of movement:
\begin{itemize}
\item Straight movement: Move both wheels forward by the same distance $k$ ($DR=DL=k$, where $DR$ and $DL$ indicate the distance traveled by the right and left wheel respectively).
\item Right-wheel rotation: Move the right wheel forward and keep the left wheel fixed ($DR=2k,DL=0$).
\item Left-wheel rotation: Move the left wheel forward and keep the right wheel fixed ($DR=0,DL=2k$).
\end{itemize}
Due to external forces affecting the motion, such as wheel slippage, the movements effectively performed by the robot differ slightly from the intended ones. 
The location of the robot at time-step $t$ is defined by a triplet $(x_t, y_t, \phi_t)$, where $x_t$ and $y_t$ represent the position of the midpoint of the wheel axle, 
whilst $\phi_t$ represents the orientation of the robot (angle formed by the perpendicular to the wheel axle and the horizontal axis). The dynamics of the robot is given by \citep{wang90location}
\begin{align}
\label{eq:rob}
&x_t=x_{t-1}+r\Delta D\cos(\phi_{t-1}+\Delta\phi/2)+\eta^x_t\nonumber,\\
&y_t=y_{t-1}+r\Delta D\sin(\phi_{t-1}+\Delta\phi/2)+\eta^y_t\nonumber,\\
&\phi_t=\phi_{t-1}+\Delta\phi+\eta^{\phi}_t,
\end{align}
with $\Delta D=(DR+DL)/2$, $\Delta\phi=(DR-DL)/L$ (where $L$ is the width of the mower), and with $r=1$, $r=\sin(\Delta\phi/2)/(\Delta\phi/2)$ for straight and rotation movements respectively.
In \eqref{eq:rob}, $\eta^x_t, \eta^y_t$ and $\eta^{\phi}_t$ are Gaussian noise terms that account for the external forces responsible for the deviations from the intended movements.

Suppose that, due to errors in the measurement system, only noisy measurements of the positions can be obtained. The goal is to estimate, at each time-step $t$, the actual robot position from the set of measurements up to time-step $t$ (on-line localization) and from all measurements (off-line localization). We can compactly write \eqref{eq:rob} and the observation process as
\begin{align}
\label{eq:roblgssm}
&h_1\!=\![x_1\hskip0.1cm y_1\hskip0.1cm\phi_1]\trans\!\sim\!{\cal N}(h_1;\mu,\Sigma)\nonumber,\\[5pt]
&h_t\!=\!f^{s_t}(h_{t-1})+\eta^h_t, \hskip0.1cm h_{t-1}\!=\![x_t\hskip0.1cm y_t\hskip0.1cm \phi_t]\trans, \hskip0.1cm\eta^h_t\!=\![\eta^x_t\hskip0.1cm \eta^y_t\hskip0.1cm \eta^{\phi}_t]\trans\!\sim\!{\cal N}(\eta^h_t;0,\Sigma_H)\nonumber,\\
&v_t\!=\!Bh_t+\eta^v_t, \hskip0.3cm B=
\left[ \begin{array}{ccc}
1 &  0 & 0\\
0 & 1 & 0\\
\end{array} \right],\hskip0.3cm\eta^v_t\sim{\cal N}(\eta^v_t;0,\Sigma_V),
\end{align}
where $s_t\in\{1,2,3\}$ indicates the type of movement undertaken by the robot, and $f^{s_t}$ is the corresponding nonlinear function.
We have therefore formulated the model as a \SLGSSM~with the only difference that the hidden dynamics evolves nonlinearly.
We can deal with that with an unscented approximation similar to one proposed in \cite{sarkka8unscented} for the \LGSSM,
which enables us to use similar inference routines to the linear case.

\section{Count-Duration-Variable \SLGSSM\label{app:SS}}
In this section, we provide time-recursive inference routines for the explicit-duration \SLGSSM~that uses count-duration variables.
\subsubsection*{Filtering}
The recursion for $\alpha^{\sigma_t}_t=p(\sigma_t|v_{1:t})$ is given by
\begin{align*}
\alpha^{\sigma_t}_t&=\frac{\sum_{\sigma_{t-1}}p(v_t,\sigma_{t-1:t}|v_{1:t-1})}{\sum_{\tilde\sigma_{t-1:t}}p(v_t,\tilde\sigma_{t-1:t}|v_{1:t-1})}\nonumber\\
&\propto \sum_{\sigma_{t-1}}p(v_t|\sigma_{t-1},s_t,\cancel{d_t,c_t},v_{1:t-1})p(\sigma_t|\sigma_{t-1},\cancel{v_{1:t-1}})p(\sigma_{t-1}|v_{1:t-1})\nonumber\\
&=\delta_{\begin{subarray}{l} c_t<d_t \\ s_{t-1}=s_t \\ d_{t-1}=d_t \\ c_{t-1}=c_t+1 \end{subarray}}\hskip-0.3cm e^{\sigma_{t-1},s_t}_t\alpha^{\sigma_{t-1}}_{t-1}
+\delta_{\begin{subarray}{l} c_t=d_t \\ c_{t-1}=1 \end{subarray}}\hskip-0.0cm
\rho_{s_td_t}\sum_{s_{t-1}}\pi_{s_ts_{t-1}}e^{\sigma_{t-1},s_t}_t\sum_{d_{t-1}}\alpha^{\sigma_{t-1}}_{t-1},
\end{align*}
where $e^{\sigma_{t-1},s_t}_t=p(v_t|\sigma_{t-1},s_t,v_{1:t-1})$.
With pre-summation over $d_{t-1}$, this recursion has computational cost ${\cal O}(TS^2d^2_{\max})$.

However, notice that $\alpha^{\sigma_t}_t$ and $\alpha^{\sigma_t'}_t$ for which $d_t-c_t=d'_t-c'_t$ differ only in $\rho_{s_td_t}$ and $\rho_{s'_td'_t}$.
As $d_t-c_t$ ranges from 0 to $d_{\max}-1$, we can define a variable $\tilde{c}_t\in\{1,\ldots,d_{\max}\}$ and form a recursion over
$\tilde\alpha^{\tilde{c}_t}_t$ such that $\alpha^{\sigma_t}_t=\rho_{s_td_t}\tilde\alpha^{s_t,d_t-c_t}_t$.
This approach reduces the cost to ${\cal O}(TS^2d_{\max})$ (a similar approach was introduced in \cite{jiang10hidden}).

Notice that $\alpha^{\sigma_t}_t=0$ implies $\alpha^{s_t,d_t,c_t-1}_{t+1}=\ldots=\alpha^{s_t,d_t,1}_{t+c_t-1}=0$
and $\alpha^{\sigma_t'}_t=0$ if $d_t-c_t=d_t'-c_t'$, \ie~if according to $v_{1:t}$ a segment starting at time-step $t-d_t+c_t$ and generated by $s_t$ cannot have duration $\geq d_t-c_t+1$, that segment cannot have duration $\geq d_t-c_t+2$
after incorporating observations $v_{t+1}$, etc.

The recursion for $\hat\alpha^{\sigma_t}_t=p(h_t|\sigma_t,v_{1:t})$ is given by
\begin{align*}
\hat\alpha^{\sigma_t}_t&=\sum_{\sigma_{t-1}}p(h_t|\sigma_{t-1},s_t,\cancel{d_t,c_t},v_{1:t})p(\sigma_{t-1}|\sigma_t,v_{1:t})\nonumber\\
&=\Bigg\{\delta_{\begin{subarray}{l} c_t<d_t \\ s_{t-1}=s_t \\ d_{t-1}=d_t \\ c_{t-1}=c_t+1 \end{subarray}}
\hskip-0.4cm+\delta_{\begin{subarray}{l} c_t=d_t \\ c_{t-1}=1 \end{subarray}}
\frac{\sum_{\substack{s_{t-1}\\d_{t-1}}}p(v_t,\sigma_{t-1:t}|v_{1:t-1})}{\sum_{\tilde\sigma_{t-1}}p(v_t,\tilde\sigma_{t-1:t}|v_{1:t-1})}\Bigg\}p(h_t|\sigma_{t-1},s_t,v_{1:t}),
\end{align*}
where we have used $p(\sigma_{t-1}=(s_t,d_t,c_t+1)|s_t,d_t,c_t<d_t)=1$. Therefore,
$\hat\alpha^{\sigma_t}_t$ is a Gaussian mixture with a complex number of components. 
Gaussian collapsing of $\hat\alpha^{s_t,d_t,c_t=d_t}_t$ reduces the cost to ${\cal O}(TS^2d^2_{\max})$.

\paragraph{Across-segment independence.}
The recursion for $\alpha^{\sigma_t}_t$ becomes
\begin{align}
\alpha^{\sigma_t}_t&\propto \sum_{\sigma_{t-1}}p(v_t|\sigma_{t-1},s_t,{\myredd {c_t,d_t}},v_{1:t-1})p(\sigma_t|\sigma_{t-1},\cancel{v_{1:t-1}})p(\sigma_{t-1}|v_{1:t-1})\nonumber\\
&=\delta_{\begin{subarray}{l} c_t<d_t \\ s_{t-1}=s_t \\ d_{t-1}=d_t \\ c_{t-1}=c_t+1 \end{subarray}}
p(v_t|\sigma_{t-1:t},v_{1:t-1})\alpha^{\sigma_{t-1}}_{t-1}\nonumber\\
&+\delta_{\begin{subarray}{l} c_t=d_t \\ c_{t-1}=1 \end{subarray}}
p(v_t|{\myredd\cancel{\sigma_{t-1}}},\sigma_t,{\myredd\cancel{v_{1:t-1}}})\rho_{s_td_t}\sum_{s_{t-1}}\pi_{s_ts_{t-1}}\sum_{d_{t-1}}\alpha^{\sigma_{t-1}}_{t-1},
\label{eq:alphaM2t}
\end{align}
where $p(v_t|s_t,d_t,c_t=d_t)={\cal N}(v_t;B^{s_t}\mu^{s_t},B^{s_t}\Sigma^{s_t}(B^{s_t})\trans+\Sigma^{s_t}_V)$.
As in the case of across-segment dependence, the computational cost can be reduced to ${\cal O}(TS^2d_{\max})$.

\noindent The recursion for $\hat\alpha^{\sigma_t}_t=p(h_t|\sigma_t,v_{1:t})$ becomes
\begin{align}
\hat\alpha^{\sigma_t}_t&=\sum_{\sigma_{t-1}}p(h_t|\sigma_{t-1},s_t,{\myredd d_t,c_t},v_{1:t})p(\sigma_{t-1}|\sigma_t,v_{1:t})\nonumber\\
&=\delta_{\begin{subarray}{l} c_t<d_t \\ s_{t-1}=s_t \\ d_{t-1}=d_t \\ c_{t-1}=c_t+1 \end{subarray}}
p(h_t|\sigma_{t-1:t},v_{1:t})
\nonumber\\
&+\delta_{\begin{subarray}{l} c_t=d_t \\ c_{t-1}=1 \end{subarray}}
\cancel{\sum_{\substack{s_{t-1}\\d_{t-1}}}}p(h_t|{\myredd\cancel{\sigma_{t-1}}},\sigma_t,{\myredd\cancel{v_{1:t-1}}},v_t)\cancel{p(\sigma_{t-1}|\sigma_t,v_{1:t})},
\label{eq:alphahatM2t}
\end{align}
where $p(h_t|s_t,c_t,d_t=c_t,v_t)$ is Gaussian with mean and covariance as in \eqref{eq:predcorri}.
Therefore $\hat\alpha^{\sigma_t}_t$ is Gaussian.
Naive computation of this recursion has cost ${\cal O}(TSd^2_{\max})$.
However, as $\hat\alpha^{\sigma_t}_t$ varies only with the difference $d_t-c_t$ for which the segment starts at a different time-step, rather than with single values of $d_t$ and $c_t$,
the cost can be reduced to ${\cal O}(TSd_{\max})$. This recursion essentially performs filtering in a \LGSSM~on segment $v_{t:t+d_{\max}-1}$ for each $t$ and $s_t$, in agreement with the explanation in \secref{sec:SLGSSMCD},
and as recursion (\ref{eq:filtM1altcat}).

\subsubsection*{Smoothing}
The recursion for $\gamma^{\sigma_t}_t=p(\sigma_t|v_{1:T})$ is given by
\begin{align*}
\gamma^{\sigma_t}_t&=\delta_{\begin{subarray}{l} c_t>1 \end{subarray}}\gamma^{s_t,d_t,c_t-1}_{t+1}
+\delta_{\begin{subarray}{l} c_t=1 \\ c_{t+1}=d_{t+1}\end{subarray}}\sum_{s_{t+1}}\underbrace{p(\sigma_t|\sigma_{t+1},v_{1:T})}_{\approx p(\sigma_t|h_{t+1}=\hat h_{t+1}^{T,\sigma_{t+1}},\sigma_{t+1},v_{1:t})}\gamma^{\sigma_{t+1}}_{t+1},
\end{align*}
where we have used $p(\sigma_t=(s_{t+1},d_{t+1},c_{t+1}+1)|s_{t+1},d_{t+1},c_{t+1}<d_{t+1},v_{1:T})=1$.

The recursion for $\hat\gamma^{\sigma_t}_t$ is given by
\begin{align*}
\hat\gamma^{\sigma_t}_t&=\sum_{\sigma_{t+1}}p(h_t|\sigma_{t:t+1},v_{1:T})p(\sigma_{t+1}|\sigma_t,v_{1:T})\\
&=\Bigg\{\delta_{\begin{subarray}{l} c_t>1 \\ s_{t+1}=s_t \\ d_{t+1}=d_t \\ c_{t+1}=c_t-1 \end{subarray}}
\hskip-0.6cm +\delta_{\begin{subarray}{l} c_t=1 \\ c_{t+1}=d_{t+1}\end{subarray}}
\sum_{\substack{s_{t+1}\\d_{t+1}}}p(\sigma_{t+1}|\sigma_t,v_{1:T})\Bigg\}p(h_t|\sigma_{t:t+1},v_{1:T}),
\end{align*}
where we have used $p(\sigma_{t+1}=(s_t,d_t,c_t-1)|s_t,d_t,c_t>1)=1$.

\noindent Notice that the approximation $p(h_{t+1}|\sigma_{t:t+1},v_{1:T})\approx \hat\gamma^{\sigma_{t+1}}_{t+1}$ in the computation of $p(h_t|\sigma_{t:t+1},v_{1:T})$ (see \eqref{eq:approx1}) becomes exact for $c_{t+1}<d_{t+1}$.
Indeed $p(h_{t+1}|\sigma_t,\sigma_{t+1}=(s_t,d_t,c_t-1),v_{1:T})=p(h_{t+1}|\sigma_{t+1}=(s_t,d_t,c_t-1),v_{1:T})$ follows from the fact that
$d_{t+1}=d_t\geq c_t>c_t-1=c_{t+1}$ and therefore $c_t$ must be equal to $c_{t+1}+1$.
Therefore $\hat\gamma^{\sigma_t}_t$ is a Gaussian mixture with a complex number of components. Gaussian collapsing of $\hat\gamma^{s_t,c_t=1,d_t}_t$ reduces the cost to ${\cal O}(TS^2d^2_{\max})$.

\paragraph{Across-segment independence.}
From Equations (\ref{eq:gammacountdurg1}) and (\ref{eq:gammacountdur}), we deduce that a time-recursive approach to computing $\gamma^{\sigma_t}_t=p(\sigma_t|v_{1:T})$ is given by
\begin{align}
\label{eq:gammaM2t}
\hskip-0.2cm&\gamma^{\sigma_t}_t=\delta_{\begin{subarray}{l} c_t>1 \end{subarray}}\gamma^{s_t,d_t,c_t-1}_{t+1}+\delta_{\begin{subarray}{l} c_t=1 \\ c_{t+1}=d_{t+1}\end{subarray}}\sum_{s_{t+1}}p(\sigma_t|\sigma_{t+1},v_{1:t},{\myredd\cancel{v_{t+1:T}}})\gamma^{\sigma_{t+1}}_{t+1}\hskip-0.1cm\\
\hskip-0.2cm&=\delta_{\begin{subarray}{l} c_t>1 \end{subarray}}\gamma^{s_t,d_t,c_t-1}_{t+1}+\delta_{\begin{subarray}{l} c_t=1 \end{subarray}}\alpha^{\sigma^{1}_t}_t\sum_{s_{t+1}}\frac{\pi_{s_{t+1}s_t}}{\sum_{\tilde s_t}\pi_{s_{t+1}\tilde s_t}\sum_{\tilde d_t}\alpha^{\tilde\sigma^{1}_t}_t}\sum_{d_{t+1}}\gamma^{s_{t+1},d_{t+1},d_{t+1}}_{t+1}.\nonumber
\end{align}
This recursion has cost ${\cal O}(TSd^2_{\max})$. 
Notice that $\alpha^{\sigma_t}_t=0$ implies $\gamma^{\sigma_t}_t=\gamma^{s_t,d_t,c_t-1}_{t+1}=\ldots=\gamma^{s_t,d_t,1}_{t+c_t-1}=0$.

The recursion for $\hat\gamma^{\sigma_t}_t$ becomes
\begin{align}
\hat\gamma^{\sigma_t}_t
&=\delta_{\begin{subarray}{l} c_t>1 \\ s_{t+1}=s_t \\ d_{t+1}=d_t \\ c_{t+1}=c_t-1 \end{subarray}}
\hskip-0.2cm p(h_t|\sigma_{t:t+1},v_{1:T})\nonumber\\
&+\delta_{\begin{subarray}{l} c_t=1 \\ c_{t+1}=d_{t+1}\end{subarray}}
\cancel{\sum_{\substack{s_{t+1}\\d_{t+1}}}}p(h_t|\sigma_t,{\myredd\cancel{\sigma_{t+1}}},v_{1:t},{\myredd\cancel{v_{t+1:T}}})\cancel{p(\sigma_{t+1}|\sigma_t,v_{1:T})}.
\label{eq:gammahatM2t}
\end{align}
This recursion has cost ${\cal O}(TSd^2_{\max})$.
It essentially performs smoothing in a \LGSSM~with cost ${\cal O}(d_t)$
on segment $v_{t:t+d_t-1}$ for each $t$, $s_t$ and $d_t$, in agreement with the explanation in \secref{sec:SLGSSMCD}.

\backmatter

\bibliographystyle{plainnat}
\bibliography{UHMSM}
\end{document}